\documentclass{article}

\usepackage[preprint]{neurips_2026_editted}

\usepackage[utf8]{inputenc} 
\usepackage[T1]{fontenc}    
\usepackage{url}            
\usepackage{booktabs}       
\usepackage{graphicx}
\usepackage{caption}
\usepackage{subcaption}
\usepackage{wrapfig}
\usepackage{float}
\usepackage{amsmath,amssymb,amsfonts}
\usepackage{nicefrac}       
\usepackage{microtype}      
\usepackage{xcolor}         
\usepackage[ruled,vlined,linesnumbered]{algorithm2e}
\usepackage{algpseudocode}

\usepackage{amsmath,amsfonts,bm}

\def\eqref#1{equation~\ref{#1}}

\def\1{\bm{1}}

\def\vu{{\bm{u}}}
\def\vv{{\bm{v}}}

\def\mA{{\bm{A}}}

\DeclareMathAlphabet{\mathsfit}{\encodingdefault}{\sfdefault}{m}{sl}
\SetMathAlphabet{\mathsfit}{bold}{\encodingdefault}{\sfdefault}{bx}{n}

\usepackage{hyperref}       

\hypersetup{
  colorlinks,
  citecolor=green,
  linkcolor=violet,
  urlcolor=blue
}

\definecolor{yhOrange}{RGB}{230,120,20}

\definecolor{hozyBlue}{RGB}{3,71,161}

\title{A Mechanistic Study of Algorithmic Generalization \\ in Diffusion Models on Modular Arithmetic}

\author{%
  Joon Hyeok Kim${}^\ast$ Yonghyun Park\thanks{Equal contribution} \; Mattis Dalsætra Østby\; Jiatao Gu \\
  Department of Computer and Information Science\\
  University of Pennsylvania\\
  Philadelphia, PA 19104, USA \\
  \texttt{\{hozy, park19, mdals, jgu32\}@seas.upenn.edu} \\
}

\begin{document}

\maketitle

\begin{abstract}
Despite their empirical success, how diffusion models generalize remains poorly understood from a mechanistic perspective. In this work, we provide a controlled mechanistic account of algorithmic learning in diffusion models. By utilizing the grokking phenomenon on an image-based modular addition task as a tractable testbed, we dissect the model's internal dynamics across two distinct visual regimes. In a single-image regime, mechanistic dissection reveals that the model implements modular addition by encoding operands as periodic representations and composing them at the self-attention--feedforward interface in a way that recovers the trigonometric addition identities. In a diverse-image regime with high intraclass variability, we find that the iterative sampling process partitions the task into an arithmetic computation phase and a subsequent visual refinement phase, with their transition precisely captured by an internal frequency-entropy signal. Together, our analysis clarifies the mechanics of algorithmic generalization of diffusion models in a controlled setting, while leaving open how broadly these mechanisms transfer beyond modular addition. Our code is publicly available at \href{https://github.com/JoonHyeok-hozy-Kim/mechanistic_diffusion}{https://github.com/JoonHyeok-hozy-Kim/mechanistic\_diffusion}.

\end{abstract}

\section{Introduction} 
Diffusion models \citep{sohl2015deep, ho2020denoising, song2020score} have achieved state-of-the-art performance across images \citep{rombach2022high, saharia2022photorealistic, ramesh2022hierarchical}, audio \citep{zhang2023survey}, video \citep{wan2025wan, wiedemer2025video, wu2025hunyuanvideo}, and scientific applications \citep{abramson2024accurate, price2025probabilistic}. Their success increasingly depends not only on generating novel samples, but also on satisfying structured constraints, e.g., capturing the chemistry of valid protein structures \citep{abramson2024accurate} or physical regularities for world simulation \citep{bruce2024genie}. We refer to this constraint-following capability as \emph{algorithmic generalization}.

Despite this empirical success, our understanding of their generalization capability remains limited. Recent studies have begun to address this gap, examining why these models avoid memorization and how their inductive biases enable novel sample generation \citep{pham2025memorization, kamb2025an, song2025selective, buchanan2025edge}. While valuable, these analyses do not yet explain \emph{how} diffusion models implement the computational rules that make their outputs not merely novel but systematically correct. This gap is particularly pressing as diffusion models advance toward tasks requiring increasingly sophisticated rule understanding, which must capture syntactic logic, physical laws, and beyond.

In contrast, research on large language models has shown that carefully controlled tasks can make internal mechanisms of algorithmic generalization more tractable to study \citep{elhage2021mathematical, olsson2022context, von2023transformers}. A notable example is the grokking phenomenon---delayed generalization after overfitting---first observed on modular addition \citep{power2022grokking, nanda2023progress, zhong2023clock}. This setup showed that transformers can learn ring-structured representations and implement explicit algorithmic circuits \citep{zhong2023clock}. Beyond merely reproducing algorithmic generalization behavior, such findings offer mechanistic accounts for how it emerges. For diffusion models, however, analogous grokking behavior and circuit-level accounts are much less established.

In this paper, we address this gap by conducting a mechanistic investigation into how diffusion models---specifically those trained with a flow-matching objective \citep{lipman2022flow, liu2022flow, li2026basicsletdenoisinggenerative}---achieve algorithmic generalization. We utilize image-based modular addition as our primary testbed because its well-characterized algorithmic solutions \citep{nanda2023progress, zhong2023clock} provide a clear reference against which we can validate our mechanistic discoveries. To understand how diffusion models combine arithmetic structure with visual generation, we study two complementary data regimes: single-image and diverse-image.

\begin{figure}[!htbp]
\centering
    \includegraphics[width=1.0\linewidth]{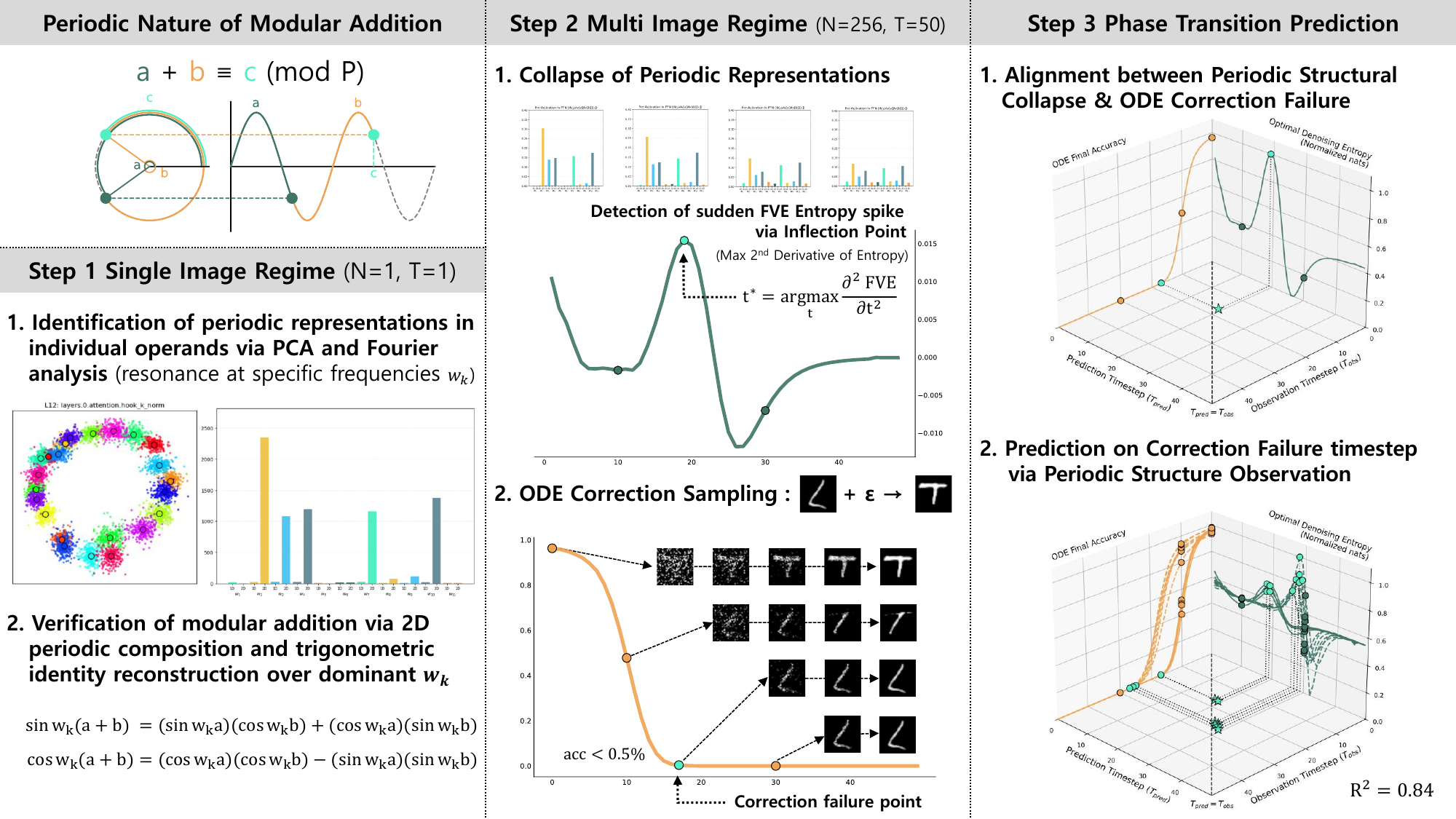}
    \caption{
        \textbf{Overview of the Analysis Pipeline.} \textbf{(Left)} To mechanistically explain the observed grokking behavior, we demonstrate that the model generalizes the periodic algorithmic structure of modular addition. Based on the periodic representations of individual operands revealed by principal component analysis (PCA) and Fourier analysis, we further verify that their 2D composition reconstructs the trigonometric addition identities. \textbf{(Middle)} Extending our analysis to the diverse-image regime, we investigate the temporal dynamics of Fractional Variance Explained (FVE) entropy and ODE sampling accuracy. \textbf{(Right)} A sharp increase in FVE entropy---reflecting the collapse of frequency concentration underpinning the periodic structure---coincides with the timestep where final ODE sampling accuracy drops near zero ($< 0.5\%$). This critical timestep, $t^\ast$, marks the transition from an arithmetic-dominated phase to a visual-refinement-dominated phase. We discover that $t^*$ corresponds to the timestep with the maximum second derivative of the FVE entropy, marking the point of the most rapid structural collapse. We further show that this transition consistently holds across various moduli $P$ and dataset variants (EMNIST/KMNIST).
    }
    \label{fig:teaser}
\end{figure}

In the single-image regime, where each class has a fixed visual representative, Fourier analysis suggests that the model relies on periodic operand representations whose interactions recover the trigonometric addition identities (Figure~\ref{fig:teaser}, left). Building on this mechanistic evidence, we then turn to the diverse-image regime and analyze the iterative denoising process. We find that the sampling trajectory exhibits two empirically distinguishable phases: an arithmetic-dominated phase with concentrated frequency structure, and a visual-refinement-dominated phase where this concentration is substantially attenuated (Figure~\ref{fig:teaser}, middle). We probe this transition by injecting noise at varying timesteps into an incorrect result image and measuring whether the model can still rectify it. The timestep beyond which rectification fails is strongly predicted by the transition estimated from internal entropy signals ($R^2=\text{0.84}$, Figure~\ref{fig:teaser}, right).



Our contributions are threefold:  
\begin{enumerate}
    \item \textbf{Grokking in Generative Diffusion Model}: We show that diffusion models can exhibit grokking on image-based modular addition, establishing a controlled setting in which algorithmic generalization can be studied mechanistically in generative diffusion models.
    \item \textbf{Mechanistic Circuit Analysis in Diffusion Model}: We reveal that the diffusion model implements modular addition through periodic representations that compose via trigonometric identities, yielding a mechanistic account of arithmetic computation in a generative model.
    \item \textbf{Mechanistic Characterization of Mode Transitions}: We identify a dynamic functional shift during the iterative denoising process. The model transitions from arithmetic computation to visual refinement near a critical noise threshold in our interventions, providing a temporal map of the measured generative dynamics. Our entropy-based analysis predicts this transition timestep, aligning closely with empirical rectification failures.
\end{enumerate}

\section{Related Work}

\paragraph{Mechanistic Interpretability}
This field aims to uncover the internal computations of neural networks by identifying the representations and circuits responsible for task-solving behavior \citep{cammarata2020thread, elhage2021mathematical}. Early studies demonstrated that large, nonlinear models often exhibit a modular structure with clean computational decompositions---such as copying, sorting, or induction---whose components can be causally isolated and manipulated \citep{wang2022interpretability, conmy2023towards, cunningham2023sparse}. These advances have established mechanistic interpretability as a powerful framework for explaining how and why generalization emerges in modern architectures.

\paragraph{Grokking and Modular Addition}
First observed in modular arithmetic tasks \citep{power2022grokking}, grokking describes delayed generalization after overfitting. Networks rapidly achieve perfect training accuracy while validation performance remains near zero, then abruptly transition to perfect generalization. This phenomenon has become a key testbed for studying how networks learn discrete rules \citep{kumar2023grokking, davies2023unifying, liu2022omnigrok, varma2023explaining, thilak2022slingshot}. Prior analyses revealed that transformers eventually discover explicit computational solutions implementing modular arithmetic, forming periodically structured representations consistent with the task's algebraic structure \citep{nanda2023progress}. Different architectures converge to distinct solutions, such as the ``clock'' or ``pizza'' strategies \citep{zhong2023clock}, providing ground-truth baselines for interpretability research. However, grokking has been studied almost exclusively in discrete transformer models, leaving diffusion-based generative models largely unexplored.



\paragraph{Generalization in Diffusion Models} Recent theoretical work examines why diffusion models avoid memorization and how their inductive biases support novel sample generation \citep{pham2025memorization, buchanan2025edge, kadkhodaie2023generalization, bonnaire2025diffusion}. A complementary line of work investigates when and why generalization fails, identifying failure modes such as mode interpolation and local generation bias that give rise to hallucinations \citep{aithal2024understanding, lu2025towards}. However, these works focus on the boundaries of generalization---either why memorization is avoided or why generalization fails---rather than explaining the underlying mechanisms of successful algorithmic generalization.

\paragraph{Mechanistic Interpretability in Diffusion Models}
Prior work has examined the internal representations of diffusion models through the lens of geometry and sparsity, identifying semantic latent spaces in U-Net bottlenecks \citep{kwon2023diffusionmodelssemanticlatent, park2023understandinglatentspacediffusion} and decomposing model activations into interpretable features via sparse autoencoders \citep{tian2025sparseautoencoderzeroshotclassifier, surkov2025onestepenoughsparseautoencoders}. The most closely related line of work studies compositional generalization in diffusion models, revealing how models combine learned concepts to produce novel outputs \citep{okawa2025compositionalabilitiesemergemultiplicatively, park2024emergencehiddencapabilitiesexploring, wiedemer2023compositionalgeneralizationprinciples, deschenaux2024goingcompositionsddpmsproduce}. However, none of these works address \emph{algorithmic} generalization, nor do they provide circuit-level mechanistic accounts of how diffusion models implement discrete computational rules. Our work fills this gap.

\section{Experimental Setup}
\label{sec:setup}

We introduce our experimental setup for mechanistic analysis of the diffusion model's algorithmic generalization during image generation. We first describe the image-based modular addition task and the DiT backbone, followed by the Fourier lens used to connect the model's self-attention computation to modular addition.

\subsection{Dataset: Image-based Modular Addition}
\label{subsec:image_dataset}
To investigate grokking in generative diffusion models, we adapt the modular addition task to the image domain. Unlike standard token-based models that predict discrete logits from one-hot vectors \citep{power2022grokking, nanda2023progress, zhong2023clock}, our framework requires the model to generate the result image $c$ directly in a high-dimensional continuous pixel space (i.e., a direct $x_0$-prediction objective; \citet{li2026basicsletdenoisinggenerative}). For the baseline $P=23$ setting, we use uppercase letters A--W from EMNIST \citep{cohen2017emnist} as visual representatives for modular classes $\{0, \dots, P-1\}$; throughout the paper, ``class'' refers to the underlying modular value and ``image'' refers to its EMNIST visual representative. As reported in Appendices~\ref{appendix:ablations_p} and~\ref{appendix:ablations_kuzushiji}, we observe the same qualitative grokking behavior, selective Fourier structure, and entropy-based transition between arithmetic computation and visual refinement across various moduli $P$ and the KMNIST dataset\citep{DBLP:journals/corr/abs-1812-01718}.

Following \citet{power2022grokking}, we partition the $P^2$ possible operand pairs into training and validation sets with a training ratio of $0.9$. To reduce commutativity-based shortcuts, we treat each unordered pair $\{a, b\}$ as a single unit; if $(a, b)$ is not included in the training set, neither is $(b, a)$. This constraint ensures that the validation set strictly contains unseen operand pairs, requiring generalization of the arithmetic rule rather than memorization of the observed pairs to achieve high validation accuracy.

Within this setup, we study two data regimes with different levels of visual diversity. We first examine the case of a single image per class ($N=1$), which provides a simplified, token-like setting for studying grokking and identifying the arithmetic circuit. We then increase the diversity to $N=256$ images per class to investigate whether the model forms class-separable representations from highly varied visual inputs. This high-diversity setting also allows us to use the denoising trajectory for a timestep-wise analysis of when the arithmetic circuit is active during sampling. For more details on the dataset construction, please refer to Appendix~\ref{appendix:multiple_image_dataset_construction}.

\subsection{Model Architecture: Single-Layer DiT}
\label{subsec:model_architecture}
We employ a single-layer Diffusion Transformer (DiT) \citep{peebles2023scalable} to facilitate a mechanistic analysis of the attention mechanism (Figure~\ref{fig:appendix_architecture}). The model is trained with a flow-matching objective \citep{lipman2022flow, liu2022flow}. Given a clean image $x_0$ and noise $x_1 \sim \mathcal{N}(0, I)$, we define the interpolated state $x_t = (1-t)x_0 + tx_1$, so that $t=0$ corresponds to the clean image and $t=1$ to pure noise. The training objective minimizes $\mathcal{L}_\text{CFM}(x_t;\theta) = \| x_1 - x_0 - v_\theta(x_t) \|_2^2$. In this work, we adopt an $x_0$-parameterization \citep{li2026basicsletdenoisinggenerative}, where the network directly predicts $x_0$, yielding the equivalent velocity $v_\theta(x_t, t) = (x_t - x_0) / t$. For further implementation details, see Appendix~\ref{appendix:impl_detail}. Appendix~\ref{appendix:ablations_multi_layer} demonstrates grokking in a standard depth-2 architecture, indicating that the phenomenon is not limited to the single-layer model.

\paragraph{Initial Noise Input for Generation} Unlike previous modular addition benchmarks that employ a special placeholder token (e.g., ``$=$'') to trigger logit-based result generation \citep{power2022grokking, nanda2023progress}, our diffusion-based framework employs a $32 \times 32$ Gaussian noise map as the initial state (Figure~\ref{fig:appendix_architecture}). The model iteratively denoises this map, conditioned on the operand images, to generate a final visual representation of the modular sum.

\paragraph{Evaluation Protocol} To evaluate modular addition accuracy, we map generated images back to their modular classes using a ResNet18 classifier \citep{he2015deepresiduallearningimage} trained on EMNIST, which achieved over 95\% accuracy on the full EMNIST test set. During dataset construction, we apply a confidence-based filtering scheme and retain only high-confidence samples for the diffusion experiments. This filtered pool is designed to make the class-to-image mapping unambiguous, so the remaining evaluation errors primarily reflect the performance of the diffusion model rather than classifier ambiguity.

\subsection{Modular Addition and Fourier Analysis}
\label{subsec:modular_addition}
We now connect the above image-generation setup to the algebraic structure of modular addition. For any classes $a,b,c \in \{0,1,\dots,P-1\}$ satisfying $a+b=c \pmod P$, our model is trained to generate the image corresponding to class $c$, given the concatenated images of $a$ and $b$. As these classes serve as either the operands or the result of the modular addition, they form a periodic structure, for which Fourier features provide a natural coordinate system: a class $x$ can be represented at frequency $w_k=\frac{2\pi k}{P}$ via the circular embedding $(\cos(w_k x),\sin(w_k x))$. In this coordinate system, adding two classes corresponds to composing their phases. Thus, an algorithmic solution to modular addition should not merely show periodicity somewhere in the network; it should synthesize one factor from $a$ and one factor from $b$ into product terms that represent $a+b$.

Prior analyses of modular addition grokking have identified this phase-composition structure in terms of the standard trigonometric addition identities \citep{nanda2023progress}. We use these identities as a compact reference for the product terms that a learned arithmetic circuit should assemble:

\begin{equation} \label{trigonometric_addition_identities}
\small
\begin{aligned}
\cos(w_k(a+b))
&= \textcolor[RGB]{128,0,32}{\cos(w_k a)}\textcolor[RGB]{0,45,110}{\cos(w_k b)}
- \textcolor[RGB]{128,0,32}{\sin(w_k a)}\textcolor[RGB]{0,45,110}{\sin(w_k b)},\\
\sin(w_k(a+b))
&= \textcolor[RGB]{128,0,32}{\sin(w_k a)}\textcolor[RGB]{0,45,110}{\cos(w_k b)}
+ \textcolor[RGB]{128,0,32}{\cos(w_k a)}\textcolor[RGB]{0,45,110}{\sin(w_k b)}.
\end{aligned}
\end{equation}

The DiT architecture provides a natural locus to investigate these product terms. In the single-image regime, the $32 \times 32$ patch size makes the concatenated input behave as three tokens: the current noisy result token $c$ and the operand tokens $a,b$. At the result position, self-attention forms attention-weighted value terms, and the FFN pre-GeLU activation is the affine transform of this attention mixture:

\begin{equation}
\label{eq:sa_ffn_circuit_schematic}
\mathrm{FFNPreGELU}_c
=
\operatorname{Affine}\!\left(
\textcolor[RGB]{128,0,32}{\vv_a}\textcolor[RGB]{0,45,110}{\mA_{c\leftarrow a}}
+
\textcolor[RGB]{128,0,32}{\vv_b}\textcolor[RGB]{0,45,110}{\mA_{c\leftarrow b}}
+ \textcolor[gray]{0.62}{\cdots}
\right).
\end{equation}

Here $\vv_i\mA_{c\leftarrow i}$ denotes the attention-weighted value contribution from token $i$ to the result position $c$; the order is schematic, matching the block view rather than the implementation's matrix multiplication order. Eq.~\ref{eq:sa_ffn_circuit_schematic} gives the DiT analogue of the product structure in Eq.~\ref{trigonometric_addition_identities}: one attention-side factor is supplied by the value vectors while the other arises from the attention weights, and their products are passed through the affine FFN map before the GeLU. Section~\ref{subsec:warm_up} tests this correspondence directly, showing that SA exposes the degree-one factors while $\mathrm{FFNPreGELU}_c$ contains the resulting degree-two Fourier terms.

\section{Results}
In this section, we first show that our diffusion model exhibits grokking in image-based modular addition, and then reveal the internal mechanisms enabling this algorithmic generalization.

\subsection{Single-image regime (Single-step Sampling)}
\label{subsec:warm_up} 

\begin{wrapfigure}[15]{r}{0.31\textwidth}
    \vspace{-35pt}
    \centering
    \includegraphics[width=0.99\linewidth]{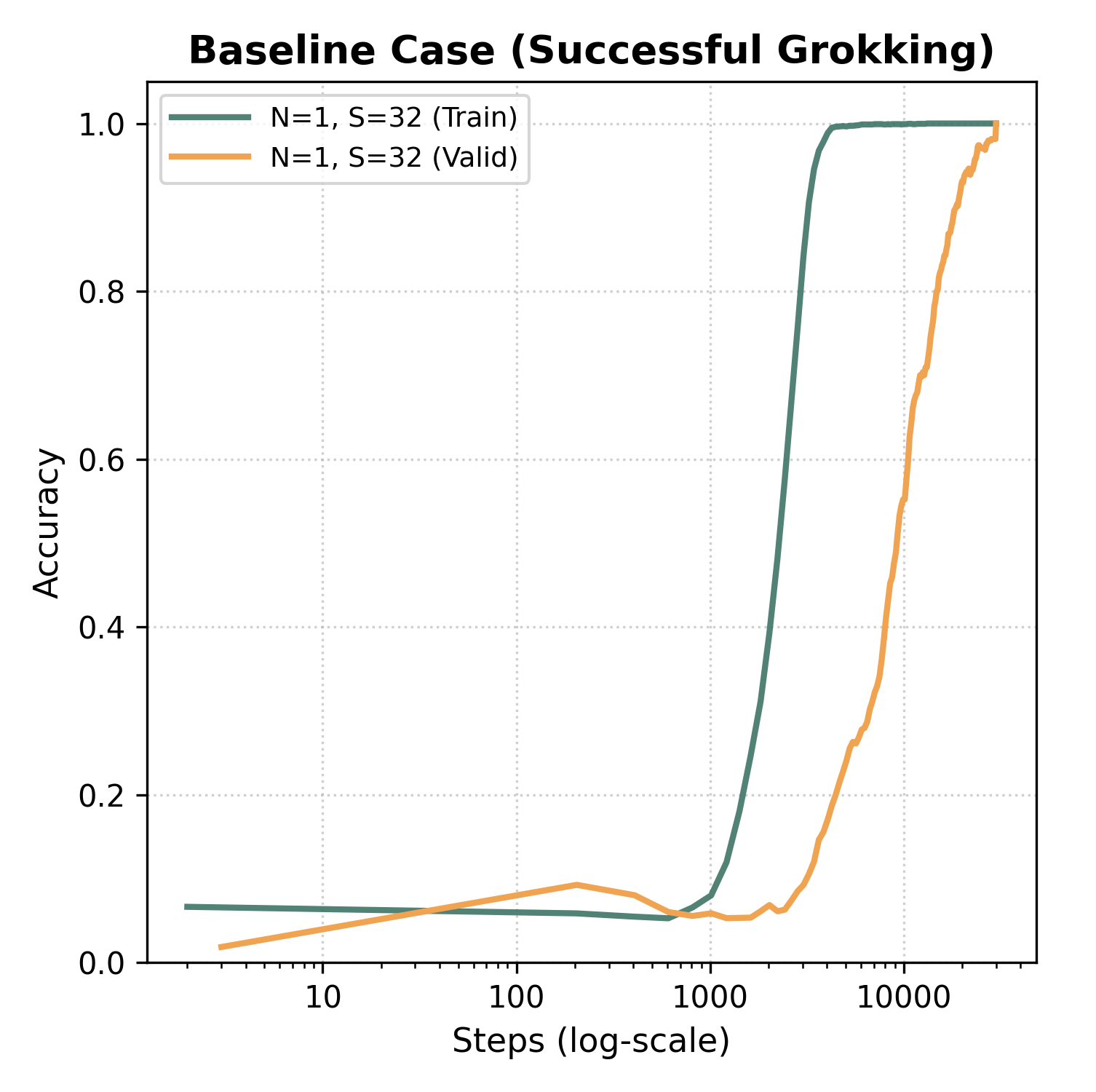}
    \caption{
        Grokking dynamics in ($N=1, S=32$): Classic grokking with a significant generalization lag post-training saturation.
    }
    \label{fig:grokking_baseline}
\end{wrapfigure}

We start with the single-image regime ($N=1$), where each image acts like a token. Beyond reproducing grokking, Fourier analysis reveals that the model encodes images as periodic features that compose via trigonometric identities to implement modular addition.

\paragraph{Grokking} Figure~\ref{fig:grokking_baseline} illustrates the grokking phenomenon reproduced in the single-image regime ($N=1$). In this setting, each modular class is represented by a fixed visual image, effectively reducing the task to a token-like prediction problem. The model first memorizes the training pairs, leading to a rapid rise in training accuracy while validation accuracy remains near chance for an extended period. After prolonged overfitting, validation accuracy undergoes a sharp transition to near-perfect generalization, indicating that the model has discovered the underlying algorithmic structure of modular addition. To investigate what internal representations drive this transition, we employ the Fourier analysis framework described in Section~\ref{subsec:modular_addition} to dissect the model's learned computations.

\paragraph{Mechanistic Evidence via Fourier Analysis} In the single-image setting, where each label maps to a unique image, the task can be analyzed as a single-step prediction from $t=1$ to $t=0$ \citep{zhang2022gddim}. We therefore focus our mechanistic analysis on the network's prediction at $t=1$.

\begin{figure}[!htbp]
\centering
\includegraphics[width=\linewidth]{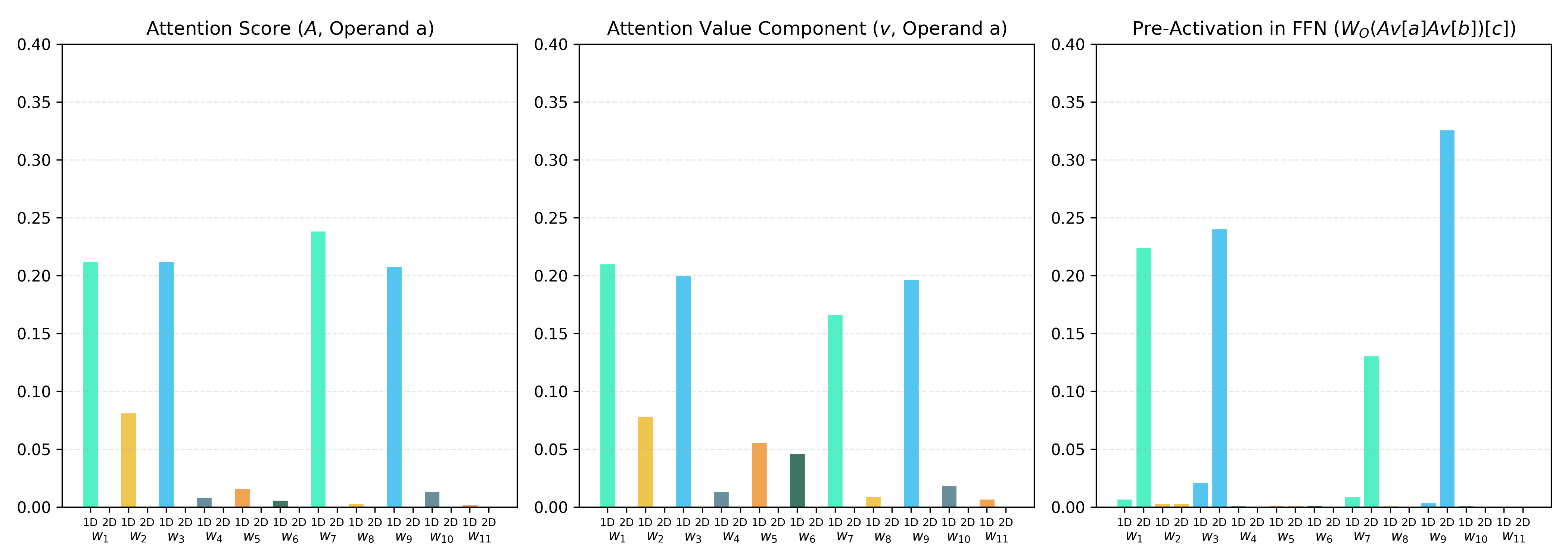}
\caption{
    \textbf{Spectral Analysis via Fractional Variance Explained (FVE).}
    FVE measures the proportion of total power attributed to each frequency's Fourier coefficients. Bars represent spectral densities for frequencies $w_k$, including degree-one (1D) and degree-two (2D) components; 2D features capture quadratic interactions such as $\cos(w_k a)\sin(w_k b)$.
    (\textbf{Left}) In Attention Score $\mA$, mediating query ($c$) and key ($a$), four selective frequencies emerge.
    (\textbf{Middle}) Value activations ($\vv[a]$) align with frequencies $w_1, w_3, w_7, w_9$, showing periodic operand encoding.
    (\textbf{Right}) At $\mathrm{FFNPreGELU}_c$, 1D features diminish while 2D components become dominant, indicating synthesis of both periodic structures.
}
\label{fig:fve_barcharts}
\end{figure}

As shown in Figure~\ref{fig:teaser} (Left), applying even a naive PCA to the self-attention (SA) Key activations at class position $a$ and timestep $T=1$ reveals a circular structure embedded within the model. Furthermore, Section~\ref{subsec:modular_addition} turns the Fourier view into two empirical tests: SA should expose degree-one operand factors, and $\mathrm{FFNPreGELU}_c$ should contain their affine readout as degree-two product terms. We first examine the attention side. As illustrated in Fig.~\ref{fig:fve_barcharts} (Left, Middle), both the attention weights ($\mA$) and value components ($\vv$) concentrate energy on a sparse set of frequencies ($w_1, w_3, w_7, w_9$), forming degree-one factors of the form $\cos(w_k a)$, $\sin(w_k a)$, $\cos(w_k b)$, and $\sin(w_k b)$.

The critical transition occurs at $\mathrm{FFNPreGELU}_c$, the affine FFN transform of the attention mixture at the result position (Fig.~\ref{fig:fve_barcharts}, Right). At this layer, degree-one components largely disappear while degree-two cross-terms, such as $\cos(w_k a)\sin(w_k b)$, become dominant. This is exactly the transition predicted by Eq.~\ref{eq:sa_ffn_circuit_schematic}: the model is no longer merely representing $a$ and $b$ independently, but is synthesizing them into interaction terms. Crucially, this confirms that the model satisfies the structural prerequisite of combining two operands---the defining characteristic of a binary operation---setting the stage to verify the exact computation in the subsequent analysis.

To prove that these synthesized 2D terms genuinely compute the exact modular addition, we follow the Fourier readout analysis of \citet{nanda2023progress}. The purpose is to test whether the activation direction associated with the correct output image encodes the modular-sum phase, rather than an arbitrary 2D periodic pattern. Since our diffusion model predicts images rather than class logits, we cannot obtain class readout directions from an unembedding matrix. We therefore estimate target-image readout directions $(\vu_k,\vv_k)$ from forward activations when the result slot is initialized with the ground-truth image.\footnote{In \citet{nanda2023progress}, the analogous directions are obtained by decomposing the neuron-logit map from output space. Here, the output is an image rather than a class logit, so we use forward activations with the ground-truth result image to estimate the corresponding readout directions; these directions are used only for analysis.} Projecting $\mathrm{FFNPreGELU}_c(a,b)$ onto such a direction turns this question into a scalar test: each dot product, such as $\vu_k^\top \mathrm{FFNPreGELU}_c(a,b)$, is a function of the operand pair $(a,b)$, and its 2D Fourier expansion should match the trigonometric addition identities if the direction indeed reads out $a+b$.

Table~\ref{tab:fourier_projection_selective} answers this question affirmatively. The projected functions are dominated by the expected sums of trigonometric products, with the signs and pairings required for $\cos(w_k(a+b))$ and $\sin(w_k(a+b))$. Thus, $\mathrm{FFNPreGELU}_c$ is the ring-structured arithmetic state produced by the SA-to-FFN circuit, not merely another layer with periodic activations.

\begin{table}[!ht]
\centering
\small
\caption{
    Projection of $\mathrm{FFNPreGELU}_c$ onto target-image readout directions yields scalar functions well explained by the trigonometric addition identities in Eq.~\ref{trigonometric_addition_identities}. For the full data, refer to Table~\ref{tab:fourier_projection_full}.
}
\label{tab:fourier_projection_selective}
\resizebox{0.9\linewidth}{!}{ 
\begin{tabular}{llc} 
\toprule
Projected channel & Fourier expansion of projected function & FVE \\
\midrule
$\cos(w_{1} (a+b))$ & $138,910 \cos(w_1 a) \cos(w_1 b) - 139,849 \sin(w_1 a) \sin(w_1 b)$ & \textbf{0.95} \\
$\sin(w_{1} (a+b))$ & $137,133 \cos(w_1 a) \sin(w_1 b) + 136,206 \sin(w_1 a) \cos(w_1 b)$ & \textbf{0.94} \\
\midrule
$\cos(w_{3} (a+b))$ & $171,404 \cos(w_3 a) \cos(w_3 b) - 181,490 \sin(w_3 a) \sin(w_3 b)$ & \textbf{0.95} \\
$\sin(w_{3} (a+b))$ & $163,861 \cos(w_3 a) \sin(w_3 b) + 160,307 \sin(w_3 a) \cos(w_3 b)$ & \textbf{0.93} \\
\midrule
$\cos(w_{7} (a+b))$ & $52,852 \cos(w_7 a) \cos(w_7 b) - 53,542 \sin(w_7 a) \sin(w_7 b)$ & \textbf{0.87} \\
$\sin(w_{7} (a+b))$ & $64,718 \cos(w_7 a) \sin(w_7 b) + 64,358 \sin(w_7 a) \cos(w_7 b)$ & \textbf{0.88} \\
\midrule
$\cos(w_{9} (a+b))$ & $176,727 \cos(w_9 a) \cos(w_9 b) - 178,583 \sin(w_9 a) \sin(w_9 b)$ & \textbf{0.95} \\
$\sin(w_{9} (a+b))$ & $192,981 \cos(w_9 a) \sin(w_9 b) + 195,516 \sin(w_9 a) \cos(w_9 b)$ & \textbf{0.96} \\
\midrule
Others ($w_{k \notin \{1,3,7,9\}}$) & Coefficients range between $\pm 10^{-3}$ and $\pm 10^3$ (Negligible) & $\le 0.01$ \\
\bottomrule
\end{tabular}
}
\end{table}

To summarize, the combination of high FVE values at position $c$ and the recovery of trigonometric identities identifies a frequency-based arithmetic circuit in this controlled single-image regime. However, analyzing a single-image case within a single-step sampling setting remains close to token-based models, as each image serves as a fixed representation for a discrete label. To move beyond this token-like setting and leverage the iterative sampling process inherent to diffusion models, we transition to a more complex regime that incorporates the temporal dimension: the sampling timestep $t$. In the following section, we investigate the multi-step sampling case, where iterative denoising is required to generalize from diverse visual inputs and progressively synthesize the arithmetic result.

\subsection{Diverse-image Regime (Multi-step Sampling)}
\label{subsec:multi_step}

Building on our observation that grokking remains feasible at $N=4$ (Appendix~\ref{appendix:grokking_variants}, Left), we scale the dataset to $N=256$ diverse images per label. While the set of arithmetic operand pairs $(a,b)$ remains identical to the single-image setting, each pair now admits multiple visual instantiations. This expansion introduces substantial intraclass variability, enabling us to study how generalization dynamics and internal arithmetic representations evolve across denoising timesteps.

\begin{wrapfigure}[12]{r}{0.31\textwidth}
    \vspace{-10pt}
    \centering
    \includegraphics[width=0.9\linewidth]{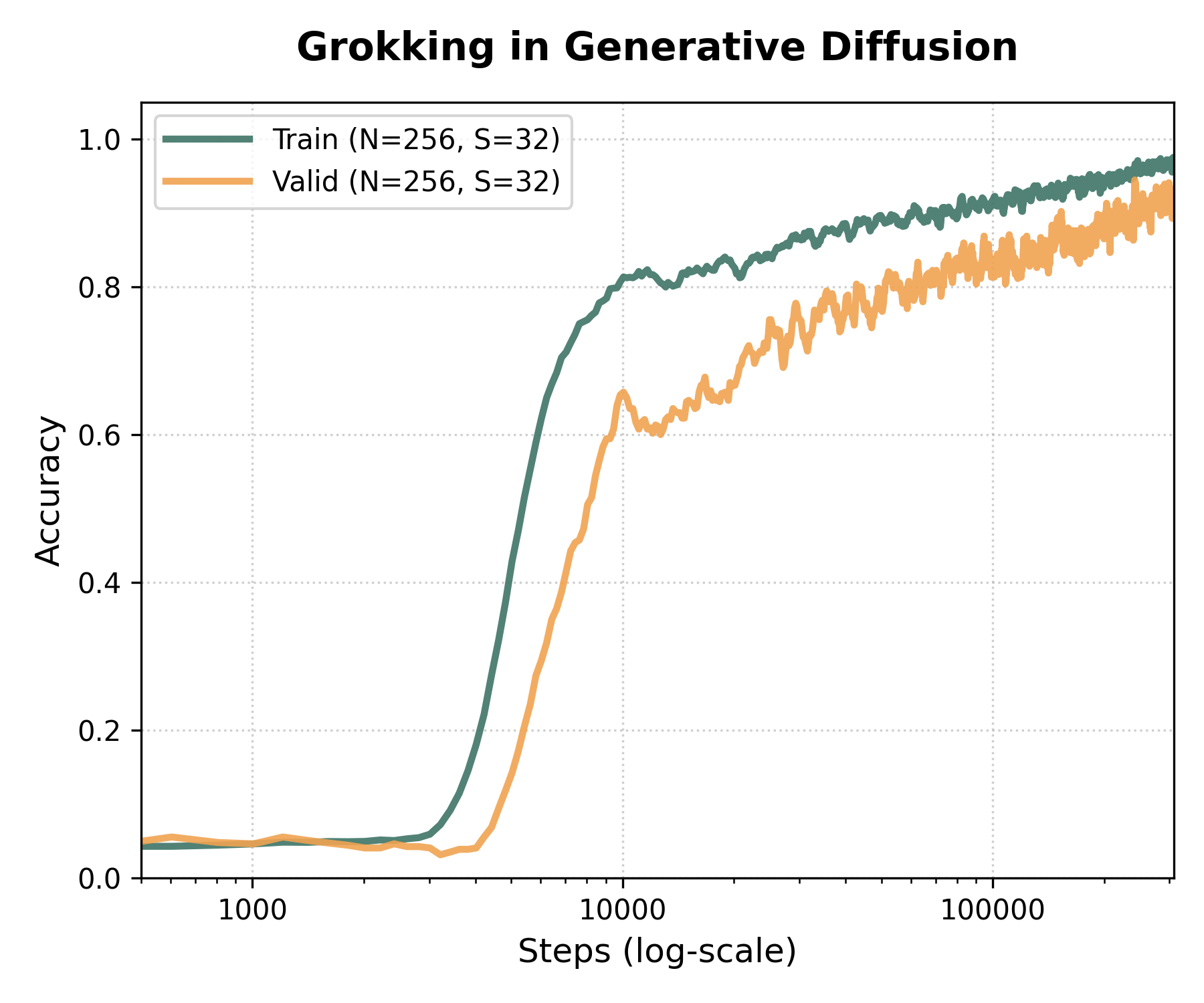}
    \caption{
        Grokking dynamics in $N=256$
    }
    \label{fig:grokking_train_result_n_256}
\end{wrapfigure}

\begin{figure}[!htbp]
    \centering
    \begin{subfigure}[b]{0.25\textwidth}
        \centering
        \includegraphics[width=\linewidth]{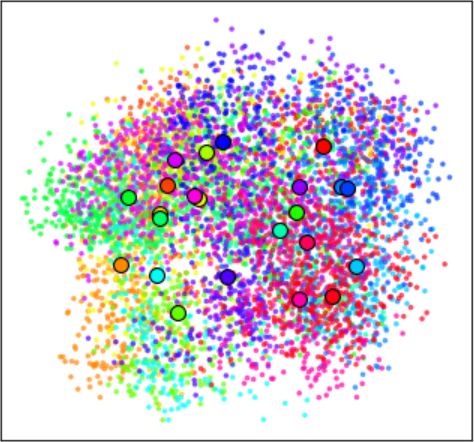}
        \caption*{Input to Pre-SA--FFN}
        \label{fig:pre-sa-ffn-pca-in-out_left}
    \end{subfigure}
    \hspace{0.02\textwidth}
    \begin{subfigure}[b]{0.25\textwidth}
        \centering
        \includegraphics[width=\linewidth]{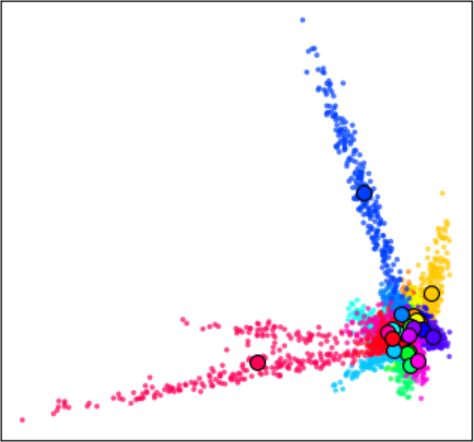}
        \caption*{Output of Pre-SA--FFN}
        \label{fig:pre-sa-ffn-pca-in-out_right}
    \end{subfigure}    
    \hspace{0.01\textwidth}
    \begin{subfigure}[b]{0.25\textwidth}
        \centering
        \includegraphics[width=\linewidth]{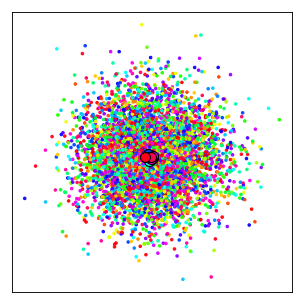}
        \caption*{Input to Standard SA--FFN}
        \label{fig:std_sa_ffn_input}
    \end{subfigure}    
    
    \caption{
    \textbf{PCA visualization of the Pre-SA--FFN layer's activations.} Each class is color-coded, with class centroids marked by circles. \textbf{(Left)} In the proposed FFN-sandwich architecture, the input activations---which correspond to the embedding layer's output---exhibit highly entangled representations, reflecting the high intra-class variance of the continuous input space. \textbf{(Middle)} Conversely, the output activations from the first FFN layer preceding the SA layer form clearer class-separable clusters, showing that the auxiliary non-linearity maps diverse visual inputs into class-aligned representations prior to self-attention. \textbf{(Right)} In contrast, the standard SA--FFN architecture's SA layer directly receives the highly entangled representations. For PCA visualizations of the subsequent layers in the FFN-sandwich architecture, refer to Figure~\ref{fig:ffn_sandwich_pca1}.
    }
    \label{fig:ffn_sandwich_comparison}
\end{figure}

\paragraph{Discrete Concept Formation via FFN} However, the standard baseline architecture (SA--FFN) failed to trigger grokking in the diverse-image regime ($N=256$). Notably, scaling up the model's width was insufficient to overcome this failure, indicating that raw capacity alone could not resolve this limitation. This suggests a structural bottleneck: the self-attention mechanism struggles to compose arithmetic operations when directly fed highly entangled visual features (Figure~\ref{fig:ffn_sandwich_comparison}, Right). To address this, we adopt an "FFN-sandwich" architecture (FFN--SA--FFN) by prepending an FFN layer. PCA visualizations (Figure~\ref{fig:ffn_sandwich_comparison}, Left and Middle) confirm that this pre-SA FFN transforms the entangled embedding layer into clearer class-wise clusters, providing the subsequent self-attention mechanism with cleaner, class-aligned inputs for arithmetic composition.\footnote{While this single-layer FFN-sandwich is a deliberate methodological choice to isolate and interpret the arithmetic mechanics, it is not a restricted edge case. We verify that the grokking phenomenon successfully reproduces in standard 2-layer diffusion architectures (SA-FFN-SA-FFN), as detailed in Appendix~\ref{appendix:ablations_multi_layer}.}

\paragraph{Grokking in the $N=256$ Regime}
As illustrated in Fig.~\ref{fig:grokking_train_result_n_256}, the $N=256$ regime exhibits a grokking pattern consistent with the $N=4$ case (Appendix~\ref{appendix:grokking_variants}, Left) despite the significantly higher visual complexity. The model achieves a terminal validation accuracy of 94\% on a rigorously constructed, non-redundant evaluation dataset (detailed in Appendix~\ref{appendix:multiple_image_dataset_construction}). For further details on the model's inference performance, please refer to Appendix~\ref{appendix:multiple_image_inference_results}.

\begin{figure}[!htbp]
\centering
\includegraphics[width=\linewidth]{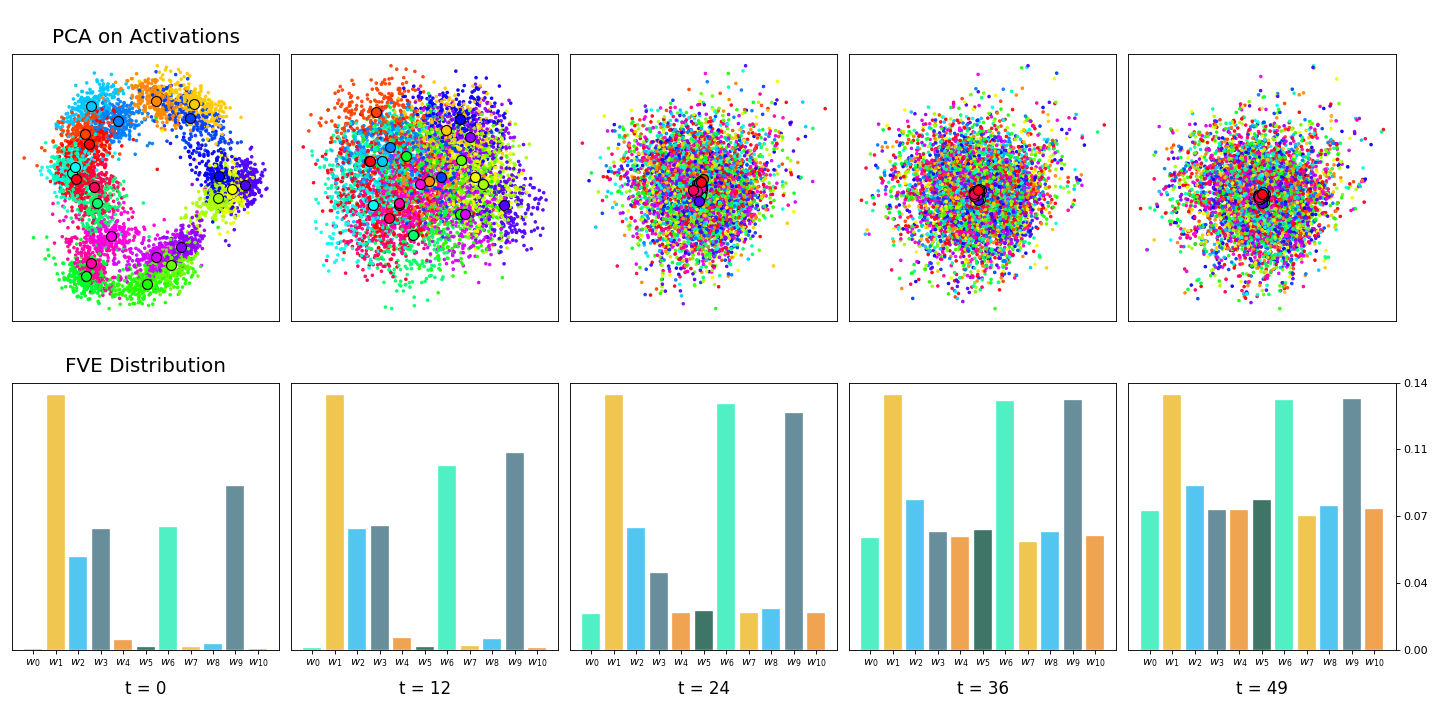}
\caption{
    \textbf{Gradual algorithmic structural collapse.}
    \textbf{(Top)} PCA on the FFN pre-GeLU activations reveals that at the onset of the sampling procedure ($t=0$), the result classes of $c$ form a vivid ring structure, reflecting the underlying periodicity. As the timestep increases, this manifold undergoes complete dissolution.
    \textbf{(Bottom)} The FVE distribution demonstrates a strong concentration on a few selective frequencies at the beginning, matching the single-image regime. This concentration steadily spreads out over time, visually confirming the breakdown of the arithmetic circuit.
}
\label{fig:timestep_fve_pca_p_23_emnist}
\end{figure}

\paragraph{Phase Transition: Arithmetic Inference and Denoising}
To probe the temporal structure ($\text{NFE}=50$), we investigate timestep-specific functional roles via input-level interventions at the result position $c$. Standard inference initializes the model with concatenated images $[x^{a}, x^{b}, x_{\text{noise}}]$, where $x^{a}, x^{b}$ denote operand images and $x_{\text{noise}} \sim \mathcal{N}(0, I)$. To test rectification capability, we intervene at an intermediate timestep $\hat{t}$ by providing $[x^{a}, x^{b}, \tilde{x}_{\hat{t}}]$. Here, $\tilde{x}_{\hat{t}} = (1-\hat{t})x^{c'} + \hat{t} x_{\text{noise}}$ represents an incorrect class $c'\;(c' \neq a+b \pmod{P})$ corrupted to noise level $\hat{t}$. Resuming ODE sampling from $\hat{t}$, we evaluate if the intervened slot is successfully rectified to the correct class $c$. This determines whether the model engages its arithmetic circuit to correct errors or merely denoises the input. To mechanistically track this activity, we monitor FFN pre-GeLU activations via PCA and FVE. Figure~\ref{fig:timestep_fve_pca_p_23_emnist} illustrates a gradual collapse of the arithmetic structure during sampling, evidenced by the dissolution of the ring-shaped manifold (PCA) and the dispersion of concentrated frequencies (FVE).

\begin{figure}[!htbp]
    \centering
    \begin{subfigure}[b]{0.54\textwidth}
        \centering
        \includegraphics[width=\linewidth]{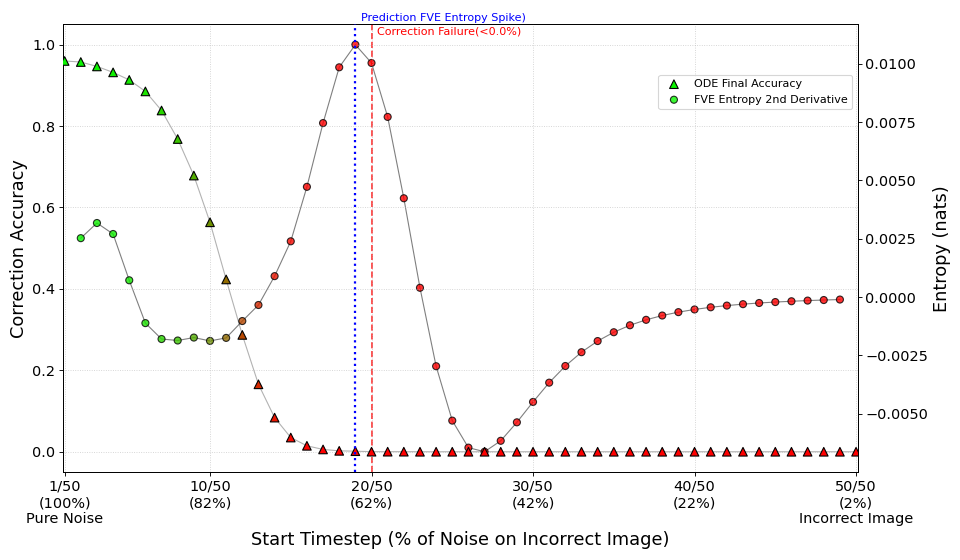}
        \label{fig:causal_intervention_trajectory_left}
    \end{subfigure}
    \hfill 
    \begin{subfigure}[b]{0.44\textwidth}
        \centering
        \includegraphics[width=\linewidth]{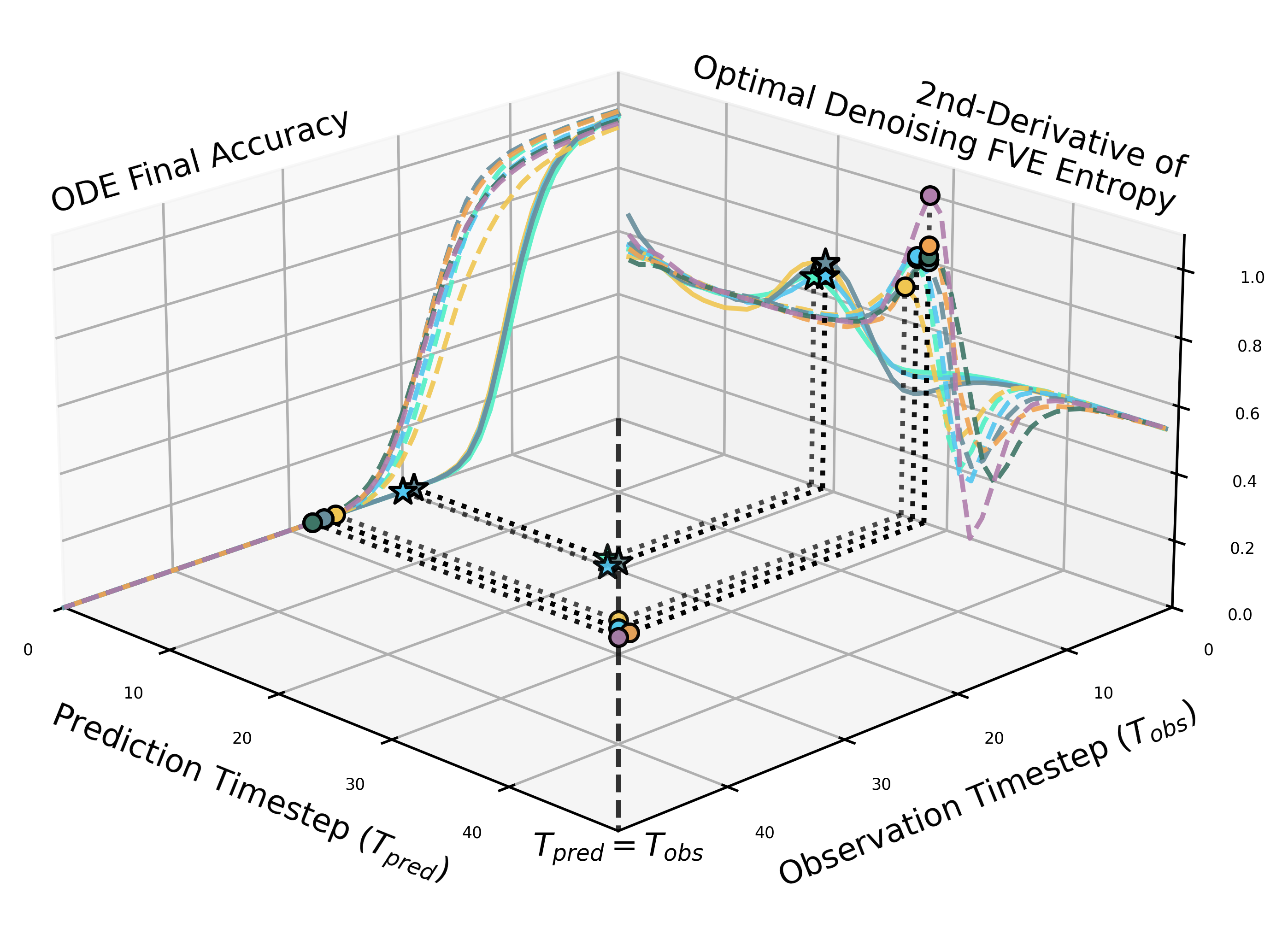}
        \label{fig:causal_intervention_trajectory_right}
    \end{subfigure}
    
    \caption{
        \textbf{Input-level intervention on result images.}
        \textbf{(Left)} The diffusion model's rectification performance via ODE sampling is indicated by triangles, while the second derivative of the FVE entropy is plotted with circles. The markers are color-coded to reflect performance, transitioning from green (successful rectification) to red (failure). Notably, the timestep of complete rectification failure (vertical red line) closely aligns with the peak of the second derivative (vertical blue line). 
        \textbf{(Right)} The rectification performance and the second derivative of the FVE entropy are plotted on the left and right vertical planes, respectively. Their critical timestep alignment is projected onto the bottom horizontal plane. The multiple trajectories represent varying moduli $P$ across heterogeneous datasets (EMNIST and KMNIST). For further details, refer to Appendices~\ref{appendix:ablations_p} and~\ref{appendix:ablations_kuzushiji}.
    }
    \label{fig:2nd_derivative_and_prediction}
\end{figure}

Mirroring the gradual collapse of the arithmetic structure within the FFN pre-GeLU activations, the ODE sampling process increasingly fails to rectify the incorrect image as $\hat{t}$ increases. Its rectification performance is depicted in Figure~\ref{fig:2nd_derivative_and_prediction} (Left, triangle markers). For timesteps $\hat{t} \ge 20$, the model fails to correct any of the perturbed samples $\tilde{x}_{\hat{t}}$ (see Figure~\ref{fig:app_mode_shift_p_23_emnist} for visual sampling examples). Coupled with the gradual collapse of the periodic structure indicative of the model's arithmetic circuit, these results strongly suggest a critical \textbf{phase transition}: in this intervention setting, the model shifts its primary functional role from arithmetic computation to purely visual refinement.

We further introduce a statistic that robustly aligns with the rectification failure timestep. Specifically, we quantify the spectral concentration---previously observed via Fourier analysis---by calculating the entropy of the FVE across all frequencies $w_k$. As the frequency concentration gradually disperses, the FVE entropy steadily increases (see Appendix~\ref{appendix:fve_entropy_trajectory}), culminating in a sudden spike near the timestep where ODE sampling completely fails. Notably, this failure timestep is closely aligned with the maximum second derivative of the FVE entropy trajectory (Figure~\ref{fig:2nd_derivative_and_prediction}, left, vertical blue line), which marks the point of most rapid structural collapse. Figure~\ref{fig:2nd_derivative_and_prediction} (right) shows a strong linear correlation ($R^2=0.84$) between the ODE sampling failure timestep and this point of structural collapse, verified across various moduli $P$ and two heterogeneous datasets (EMNIST and KMNIST \citep{DBLP:journals/corr/abs-1812-01718}).

In summary, our findings establish a definitive phase transition within the diffusion sampling trajectory. By bridging the internal collapse of periodic representations with behavioral rectification failure, we demonstrate that the model strictly partitions its generation process: an early algorithmic reasoning phase followed by a late visual refinement phase.

\section{Conclusion}
We mechanistically analyze algorithmic learning in diffusion models. Using grokking as a testbed for generalization, we identify a circuit encoding operands as periodic representations that compose via trigonometric identities, facilitated by an FFN-sandwich architecture in diverse-image regimes. Trajectory analysis reveals a phase transition at a critical noise threshold, where the model's functional role shifts from arithmetic computation to visual refinement. These findings provide a basis for interpreting diffusion models across broader datasets and reasoning tasks.

\bibliographystyle{plainnat}
\bibliography{iclr2026_delta}

\appendix
\clearpage
\section{Implementation Details}
\label{appendix:impl_detail}

We train our model using the Rectified Flow (RF) framework \citep{liu2022flow} to predict the velocity vector $v_{\theta}(x_t, t)$. Following the $x_0$-parameterization adopted in this work, the model directly predicts the clean image $x_0$, which is related to the velocity by $v_{\theta}(x_t, t) = (x_t - x_0)/t$. This formulation enables a straight-path mapping between noise and pixel space, which we find more conducive to visualizing the structural organization of representations during the denoising process \citep{li2026basicsletdenoisinggenerative}. Figure~\ref{fig:appendix_architecture} depicts the overall DiT architecture. The detailed hyperparameters used in our experiments are as follows:

\begin{figure}[H]
\centering
    \includegraphics[width=0.40\linewidth]{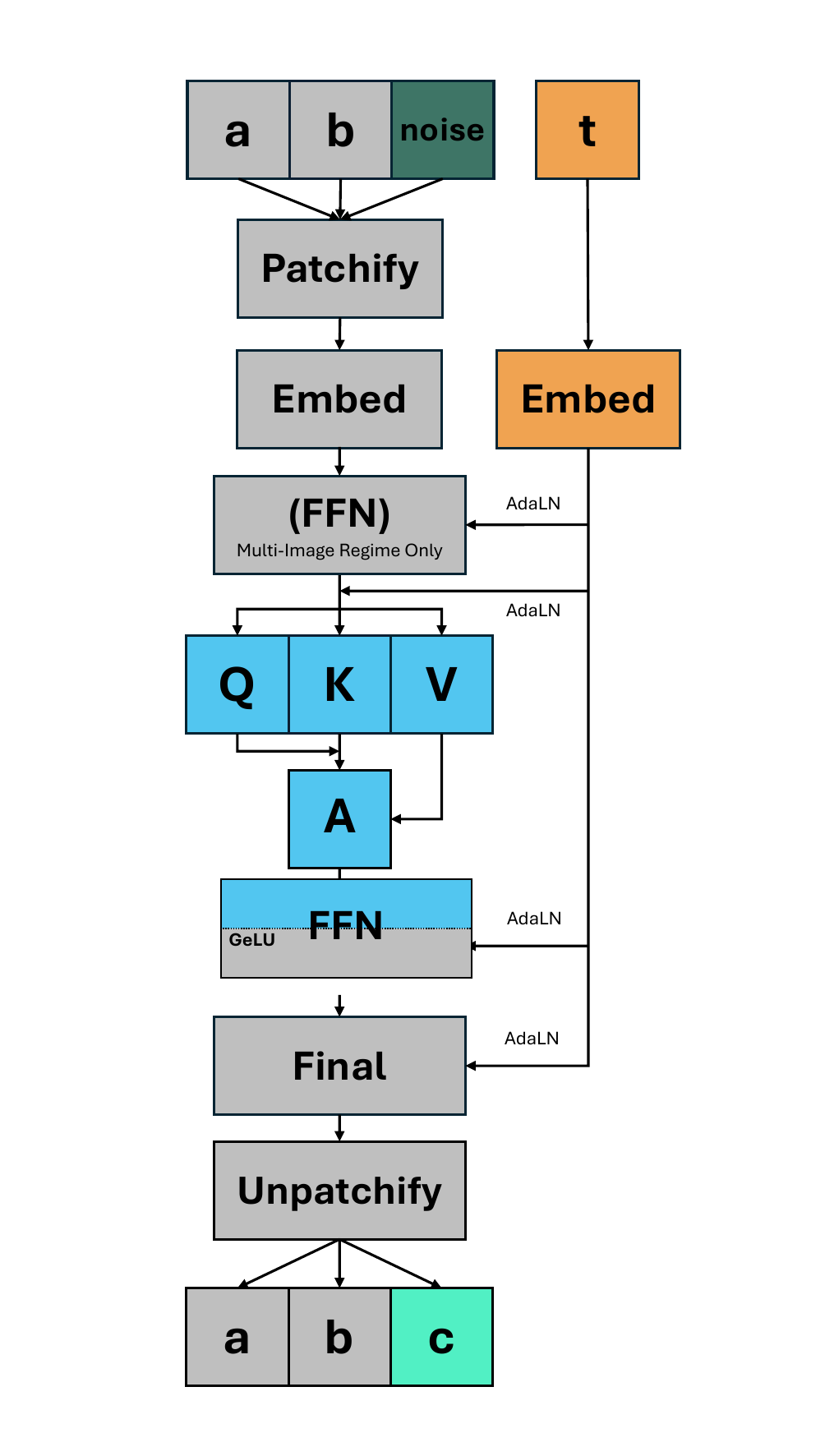}
    \caption{
    \textbf{Schematic of the Single-Layer Diffusion Transformer Architecture.} 
    The operands $a, b$ and a Gaussian noise map are concatenated, patchified, and projected into the latent space. Within the self-attention block, the model computes components $Av[a]$ and $Av[b]$, which are then fused at the attention-FFN interface. This transition facilitates a quadratic mixture of operand features, enabling the emergence of the modular addition algorithm. Temporal context from sampling time $t$ is injected via Adaptive Layer Normalization (AdaLN) \citep{peebles2023scalable}, while the Residual Gate dynamically modulates the information flow between layers. The process culminates in the direct synthesis of the target image $c$ within the high-dimensional pixel space at the original noise location.
    }
    \label{fig:appendix_architecture}
\end{figure}

\paragraph{Patchification}
Following the Diffusion Transformer (DiT) architecture \citep{peebles2023scalable}, we divide each $32 \times 32$ pixel image into non-overlapping patches of side length $S$. Our setup consists of three images---the current result slot $c$ and the operands $a,b$---which are concatenated and converted into a sequence of spatial tokens. The baseline uses $S=32$, so each image contributes one token and the sequence length is three; this is the setting used for the clean SA-to-FFN circuit analysis in Section~\ref{subsec:warm_up}. We also evaluate $S=16$, yielding four patches per image and a total sequence length of 12 tokens. This configuration successfully exhibited the grokking phenomenon (Appendix~\ref{appendix:grokking_variants}). Analyzing dependencies across patch boundaries is a natural next setting for studying how the arithmetic circuit distributes across spatial tokens.

\paragraph{Embedding and Unembedding Layers} We utilize untied linear layers for patch processing. The embedding matrix $W_E \in \mathbb{R}^{d \times d_{\text{patch}}}$ and unembedding matrix $W_U \in \mathbb{R}^{d_{\text{patch}} \times d}$ are independently parameterized. For the baseline model ($S=32$), $d_{\text{patch}} = 1024$. For the reported multi-patch configuration ($S=16$), $d_{\text{patch}} = 256$.

\paragraph{Self-attention} The self-attention block consists of 16 heads with a model dimension $d = 512$, resulting in a head dimension of $d_h = 32$. We incorporate 2D Rotary Positional Embeddings (RoPE) \citep{su2024roformer} to provide spatial context for the patchified tokens.

\paragraph{Feedforward Network} For the single-image model-size-L baseline, the transformer uses $d=512$ and FFN hidden dimension $d_{\mathrm{FFN}}=2048$. For the large FFN-sandwich model used in the diverse-image regime, width scale 4 gives $d=2048$ and $d_{\mathrm{FFN}}=8192$ for each GeLU FFN. While we explored more complex variants such as SwiGLU \citep{shazeer2020gluvariantsimprovetransformer} for improved reconstruction, the standard GeLU activation gives the clearest mechanistic hook for tracking pre-activation entropy and Fourier structure.

\begin{table}[H]
\centering
\caption{Model and dataset hyperparameters.}
\label{tab:hyperparameters}
\small
\begin{tabular}{lc}
\toprule
\textbf{Parameter} & \textbf{Value} \\
\midrule
\multicolumn{2}{l}{\textit{Dataset}} \\
Modulus ($P$) & 23 \\
Images per symbol ($N$) & 1, 4, 256 \\
Training ratio ($R$) & 0.9 \\
Image resolution & $32 \times 32$ \\
Patch size & $32 \times 32$, $16 \times 16$ \\
\midrule
\multicolumn{2}{l}{\textit{Architecture}} \\
Model dimension ($d_{\text{model}}$) & 512, 2048 \\
Number of layers & 1 \\
Number of heads & 16, 64 \\
Head dimension ($d_h$) & 32 \\
FFN dimension & 2048, 8192 \\
Positional encoding & RoPE \\
FFN activation & GeLU \\
Total parameters & 6.7M, 106.2M \\
\bottomrule
\end{tabular}
\end{table}

\paragraph{Compute Resources}
All experiments were conducted on a single NVIDIA A100 GPU (80GB). A typical training run for the $P=23$ EMNIST task takes approximately 8 hours to reach 100K steps. The total compute consumed for this project, including preliminary experiments and hyperparameter tuning, is estimated to be around 300 GPU hours.

\paragraph{Training} Models were trained using the AdamW optimizer \citep{DBLP:journals/corr/abs-1711-05101} with a Warmup-Stable-Decay (WSD) learning rate scheduler \citep{hu2024minicpmunveilingpotentialsmall}. The WSD scheduler was employed to ensure stable convergence during the initial arithmetic learning phase before decaying the learning rate for final performance refinement.

\section{Grokking Dynamics Across Visual and Structural Variations}
\label{appendix:grokking_variants}
We verify that the grokking dynamics are robustly controlled by two critical parameters: the number of images per class ($N$) and the patch size ($S$). As shown in Figure 9, even a modest increase in visual diversity ($N=4$) significantly reduces the generalization lag compared to the single-image baseline. This success at a smaller scale served as a crucial stepping stone for scaling our analysis to $N=256$ in Section~\ref{subsec:multi_step}. Furthermore, decreasing $S$ (increasing the number of tokens) is found to prolong overfitting and delay the grokking point. Nevertheless, the reproduction of grokking in the patchification case establishes a mechanistic account for more generalized settings, requiring a more nuanced analysis of the interactions between patches.

\begin{figure}[H]
\centering
    \includegraphics[width=0.80\linewidth]{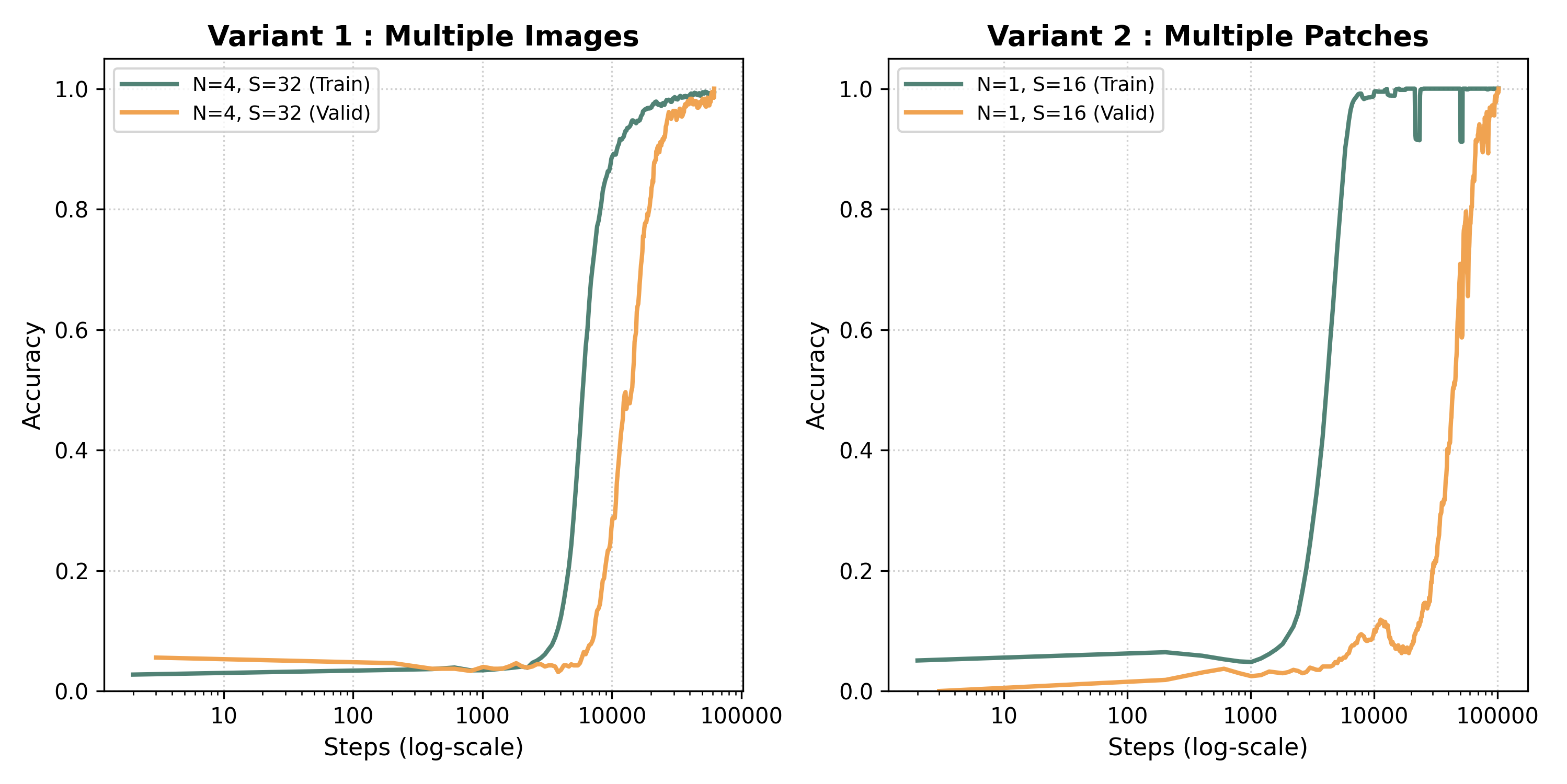}
    \caption{
    (\textbf{Left}) Variant 1 ($N=4, S=32$): Increased visual diversity reduces the lag, showing synchronized convergence where validation accuracy spikes before training saturation—motivating the $N=256$ scaling in Section~\ref{subsec:multi_step}.
    (\textbf{Right}) Variant 2 ($N=1, S=16$): Higher resolution via smaller patches prolongs overfitting and delays the grokking point. See Appendix~\ref{appendix:impl_detail} for the patchification settings.
    }
    \label{fig:appendix_grokking_demo_variant_settings}
\end{figure}

\section{Single-Image Regime Details}
\label{appendix:single_step_analysis}

\subsection{Sampling Results}
We first present the single-step inference results generated by our model. Since each label is represented by a single image, the model simply memorizes the mapping and outputs the corresponding image according to the rules of modular addition.

\begin{figure}[H]
    \centering
    \begin{subfigure}[b]{0.09\textwidth}
        \centering
        \includegraphics[width=\linewidth]{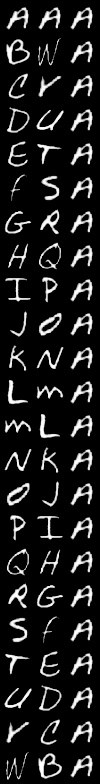}
    \end{subfigure}
    \hfill 
    \begin{subfigure}[b]{0.09\textwidth}
        \centering
        \includegraphics[width=\linewidth]{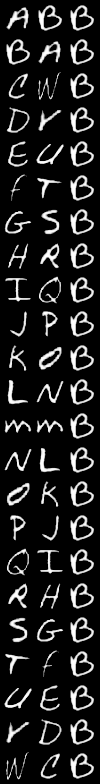}
    \end{subfigure}
    \hfill 
    \begin{subfigure}[b]{0.09\textwidth}
        \centering
        \includegraphics[width=\linewidth]{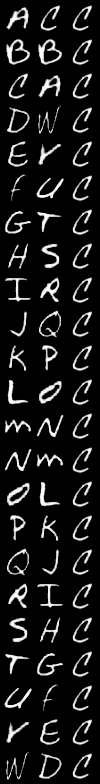}
    \end{subfigure}
    \hfill 
    \begin{subfigure}[b]{0.09\textwidth}
        \centering
        \includegraphics[width=\linewidth]{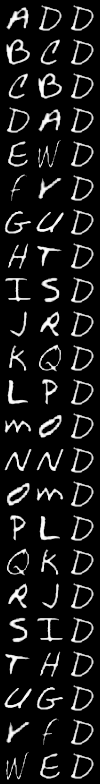}
    \end{subfigure}
    \hfill 
    \begin{subfigure}[b]{0.09\textwidth}
        \centering
        \includegraphics[width=\linewidth]{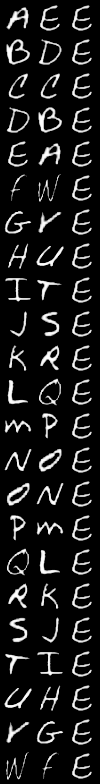}
    \end{subfigure}
    \hfill 
    \begin{subfigure}[b]{0.09\textwidth}
        \centering
        \includegraphics[width=\linewidth]{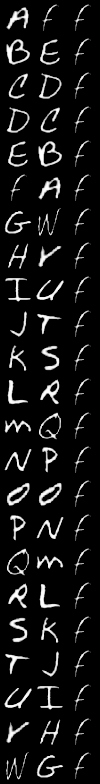}
    \end{subfigure}
    \hfill 
    \begin{subfigure}[b]{0.09\textwidth}
        \centering
        \includegraphics[width=\linewidth]{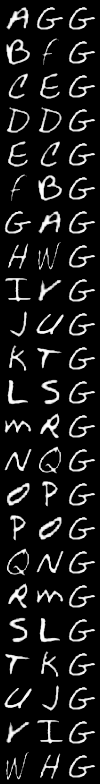}
    \end{subfigure}
    \hfill 
    \begin{subfigure}[b]{0.09\textwidth}
        \centering
        \includegraphics[width=\linewidth]{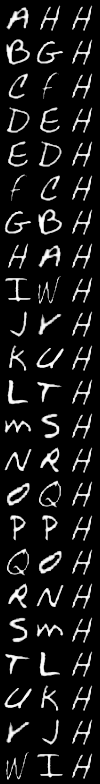}
    \end{subfigure}
    
\caption{
    \textbf{Single-step sampling results}
}
    \label{fig:appendix_single_step_sampling_result}
\end{figure}

\clearpage

\subsection{Fourier Analyses Results}
We present comprehensive Fourier analysis results across all layers, including both FVE bar charts and neuron-level activations. For each layer, the left-hand bar chart displays the FVE for frequencies $w_1, \ldots, w_{11}$. As discussed in Section~\ref{subsec:warm_up}, the early embedding layer does not yet concentrate FVE on the selective frequencies used by the arithmetic circuit.

\begin{figure}[H]
    \centering
    \begin{subfigure}[b]{0.49\textwidth}
        \centering
        \includegraphics[width=\linewidth]{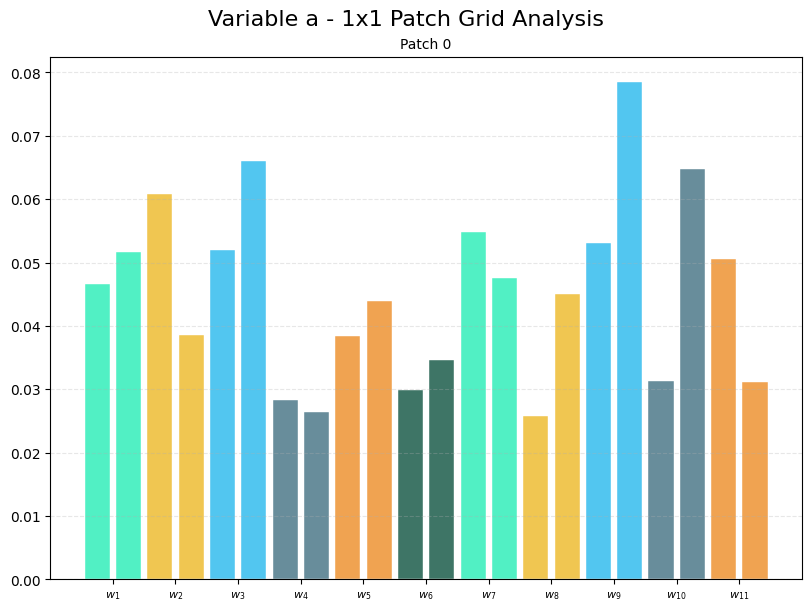}
        \caption{FVE barchart}
    \end{subfigure}
    \hfill 
    \begin{subfigure}[b]{0.49\textwidth}
        \centering
        \includegraphics[width=\linewidth]{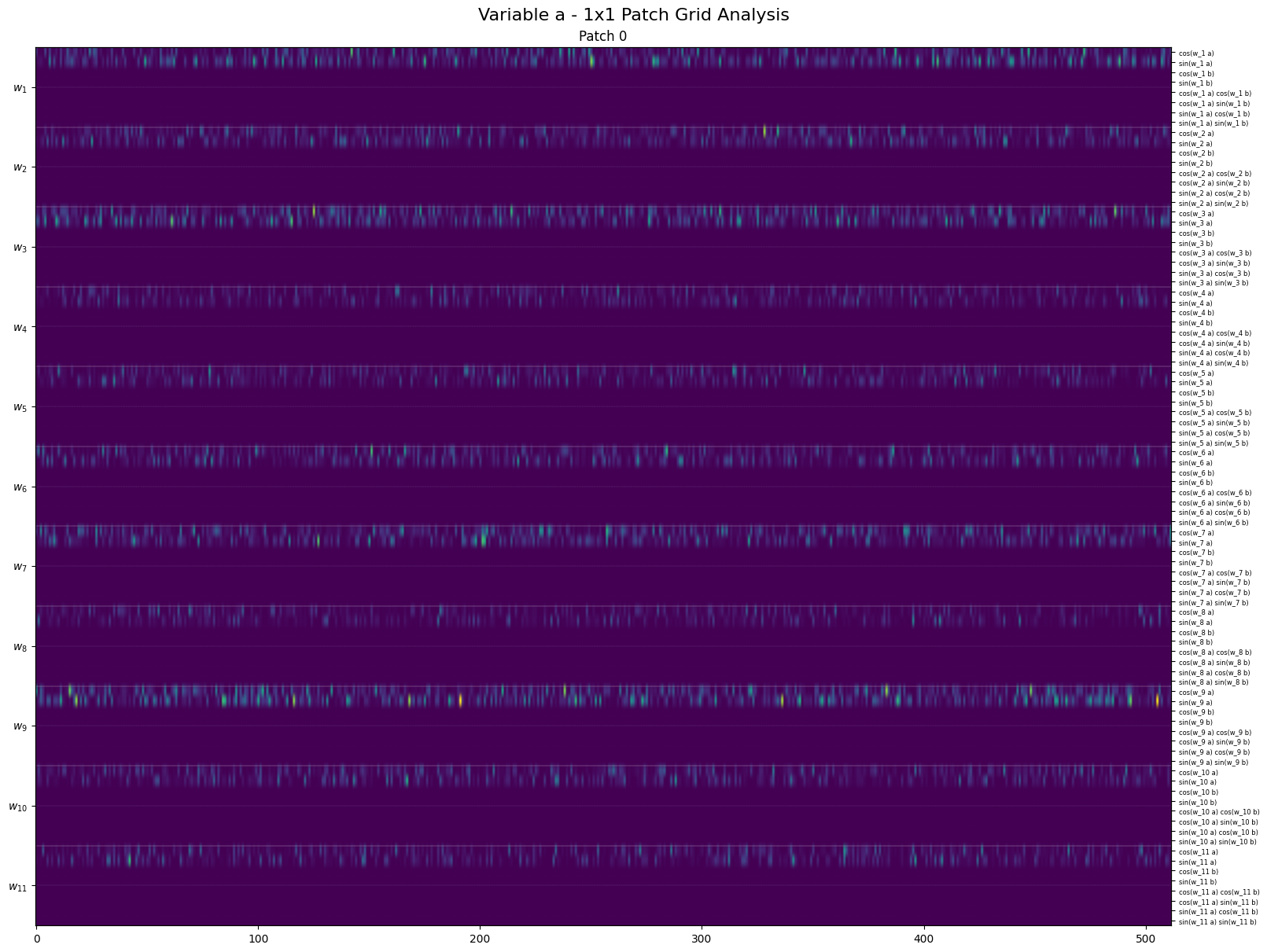}
        \caption{Heatmap of activations in neuron and head level}
    \end{subfigure}
    
\caption{
    \textbf{Embedding Activation of the operand} $a$ 
}
    \label{fig:appendix_embedding_activation}
\end{figure}

However, we observe that four specific frequencies--$w_1, w_3, w_7$, and $w_9$--account for a large fraction of the variance within the self-attention blocks, as shown below. For these layers, we further provide activations at the individual attention head level. Several heads show concentrated sensitivity to individual frequencies, revealing frequency-specific specialization.

\begin{figure}[H]
    \centering
    \begin{subfigure}[b]{0.40\textwidth}
        \centering
        \includegraphics[width=\linewidth]{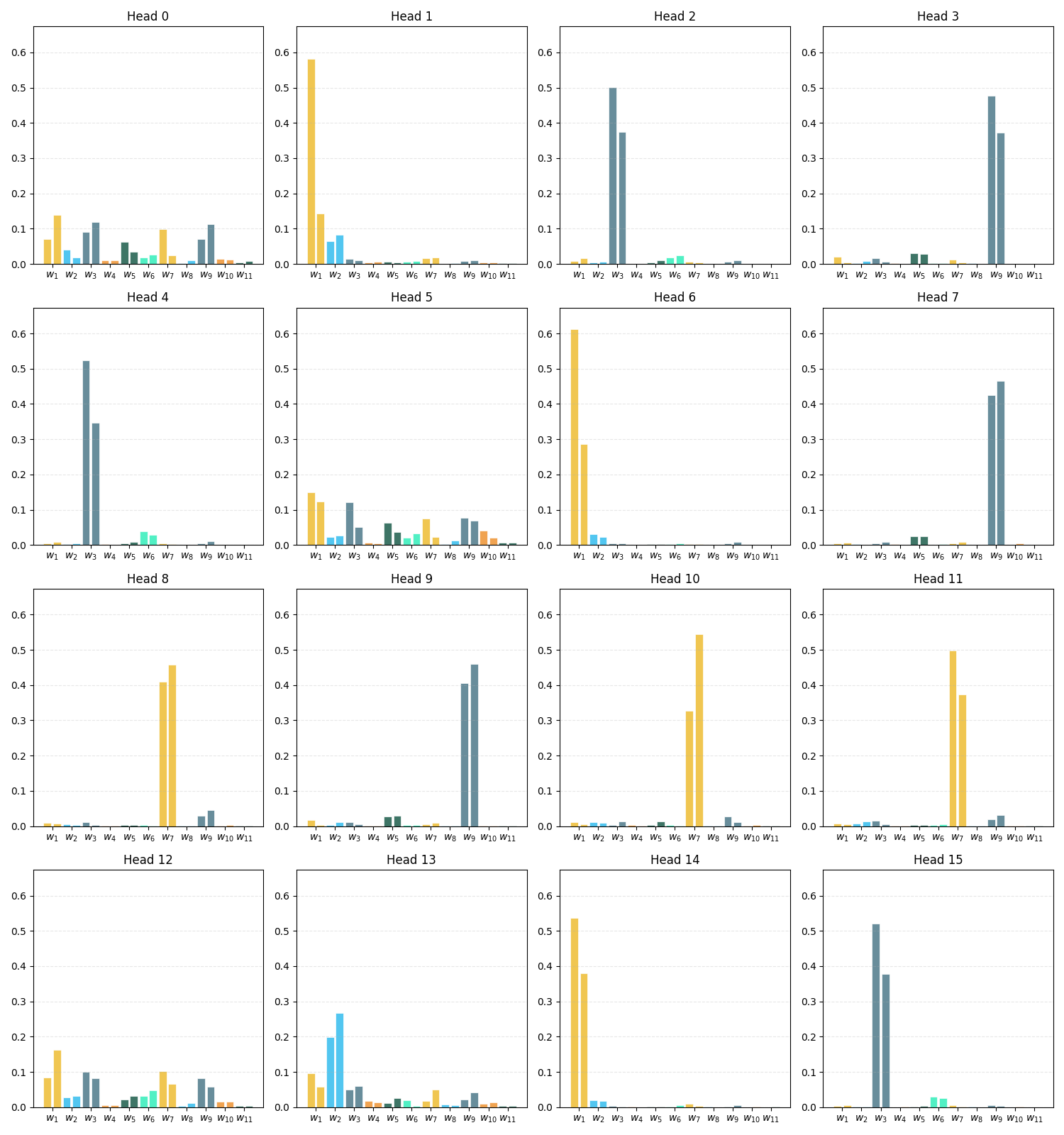}
        \caption{FVE barchart}
    \end{subfigure}
    \hfill 
    \begin{subfigure}[b]{0.59\textwidth}
        \centering
        \includegraphics[width=\linewidth]{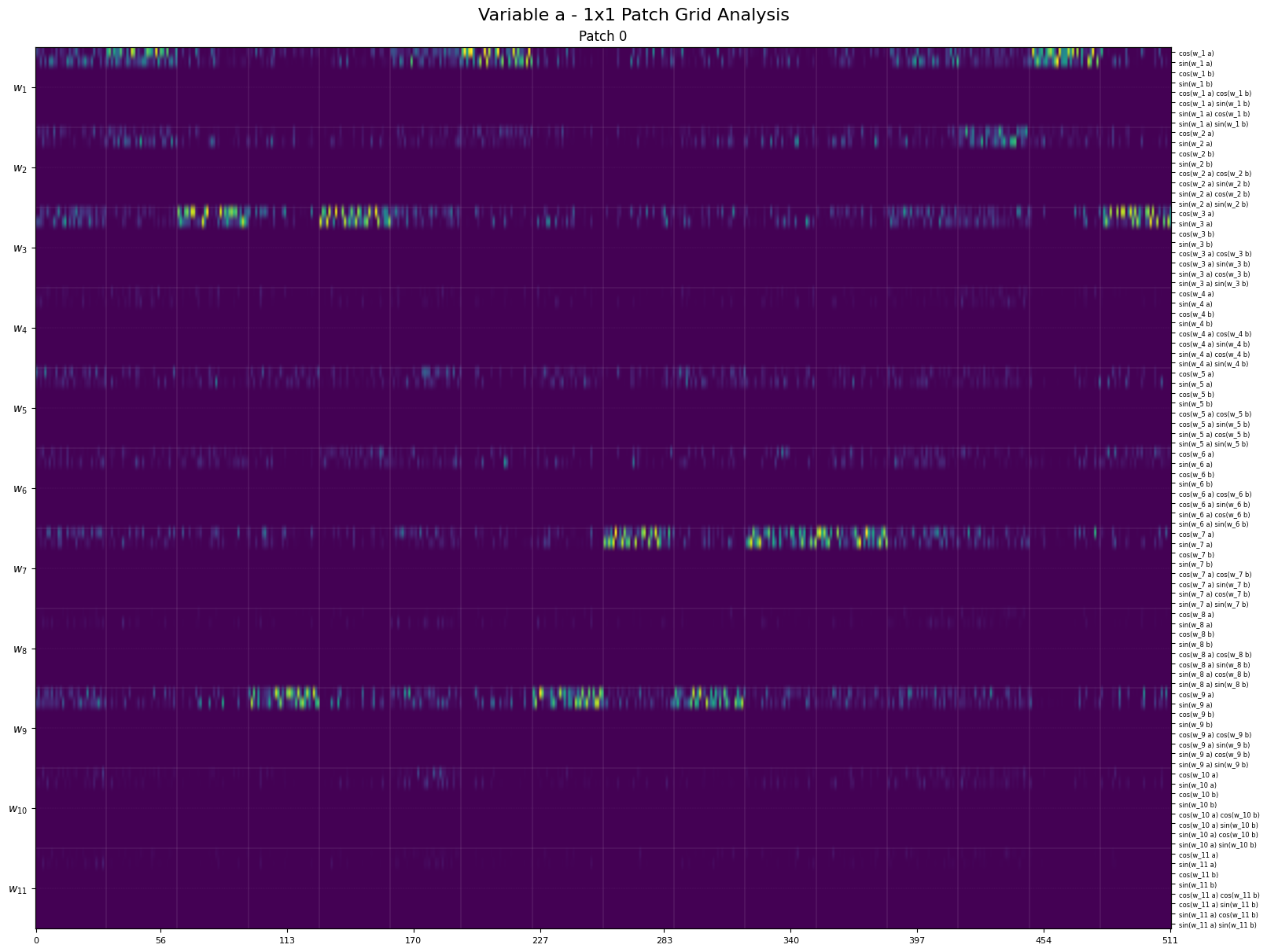}
        \caption{Heatmap of activations in neuron and head level}
    \end{subfigure}
    
\caption{
    \textbf{Attention Value Activation of the operand} $a$ 
}
    \label{fig:appendix_attention_value}
\end{figure}

\clearpage

\begin{figure}[H]
    \centering
    \begin{subfigure}[b]{0.35\textwidth}
        \centering
        \includegraphics[width=\linewidth]{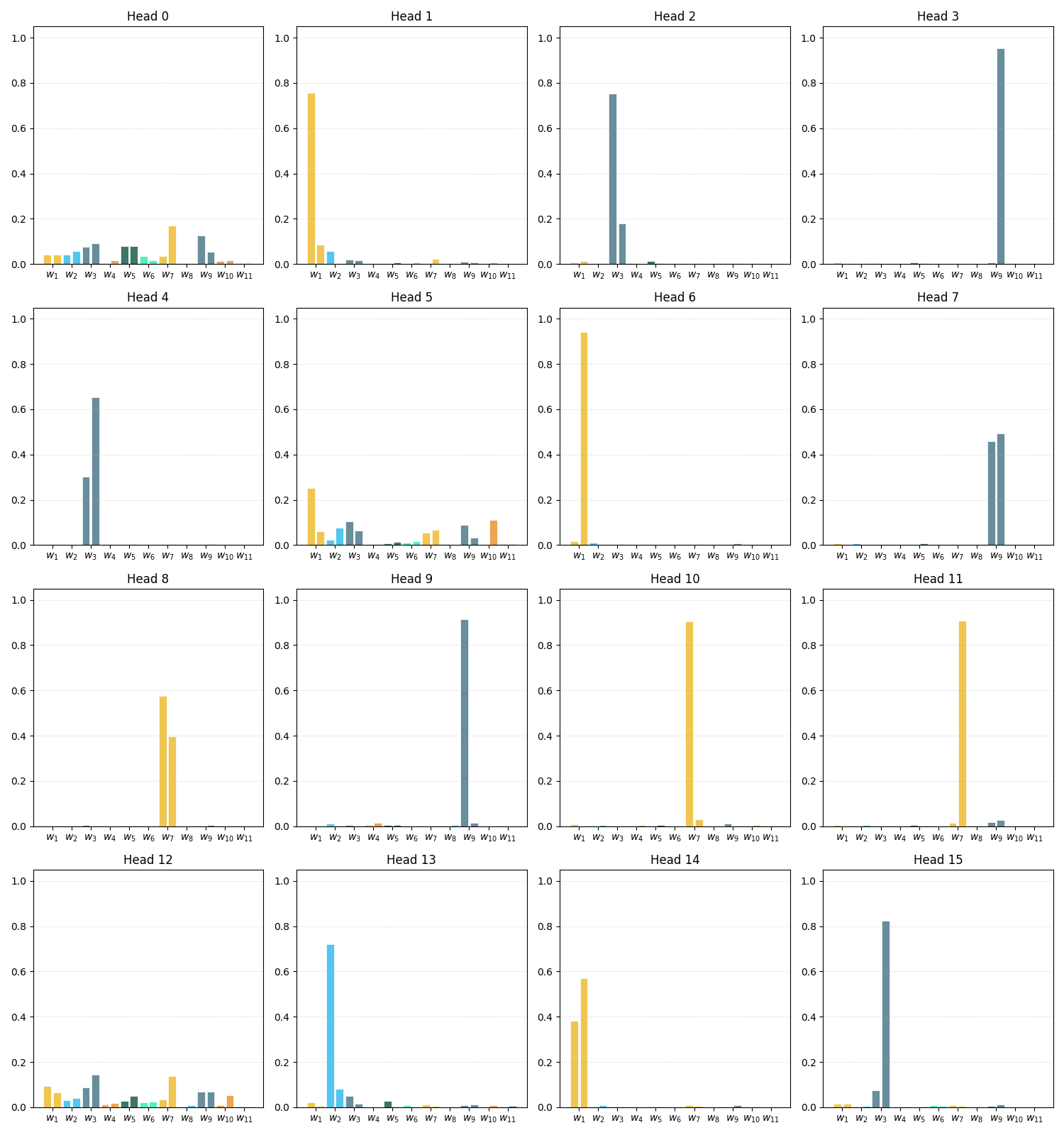}
        \caption{FVE barchart}
    \end{subfigure}
    \begin{subfigure}[b]{0.55\textwidth}
        \centering
        \includegraphics[width=\linewidth]{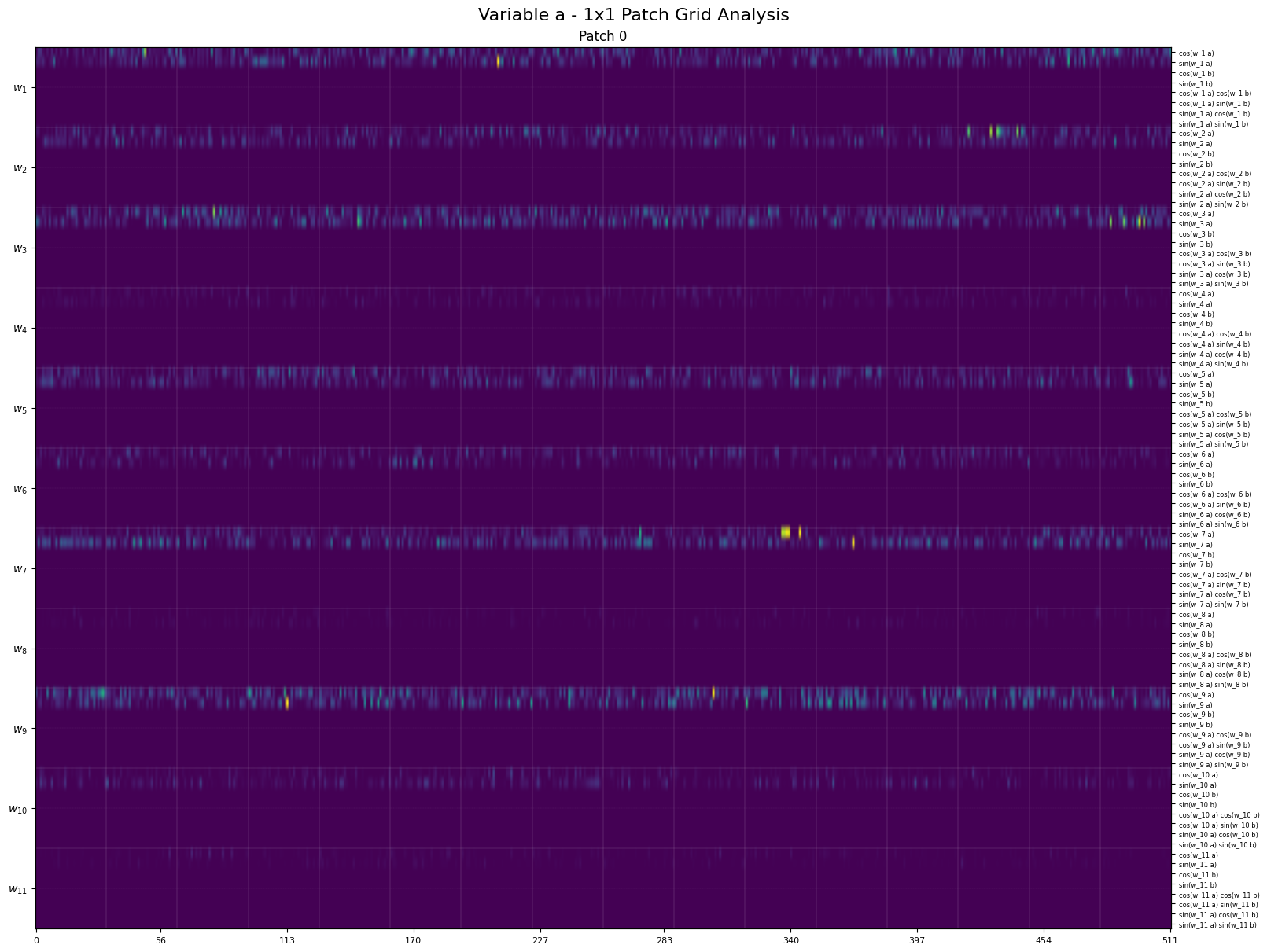}
        \caption{Heatmap of activations in neuron and head level}
    \end{subfigure}
    
\caption{
    \textbf{Attention Key Activation of the operand} $a$ 
}
    \label{fig:appendix_attention_key}
\end{figure}

\begin{figure}[H]
    \centering
    \begin{subfigure}[b]{0.35\textwidth}
        \centering
        \includegraphics[width=\linewidth]{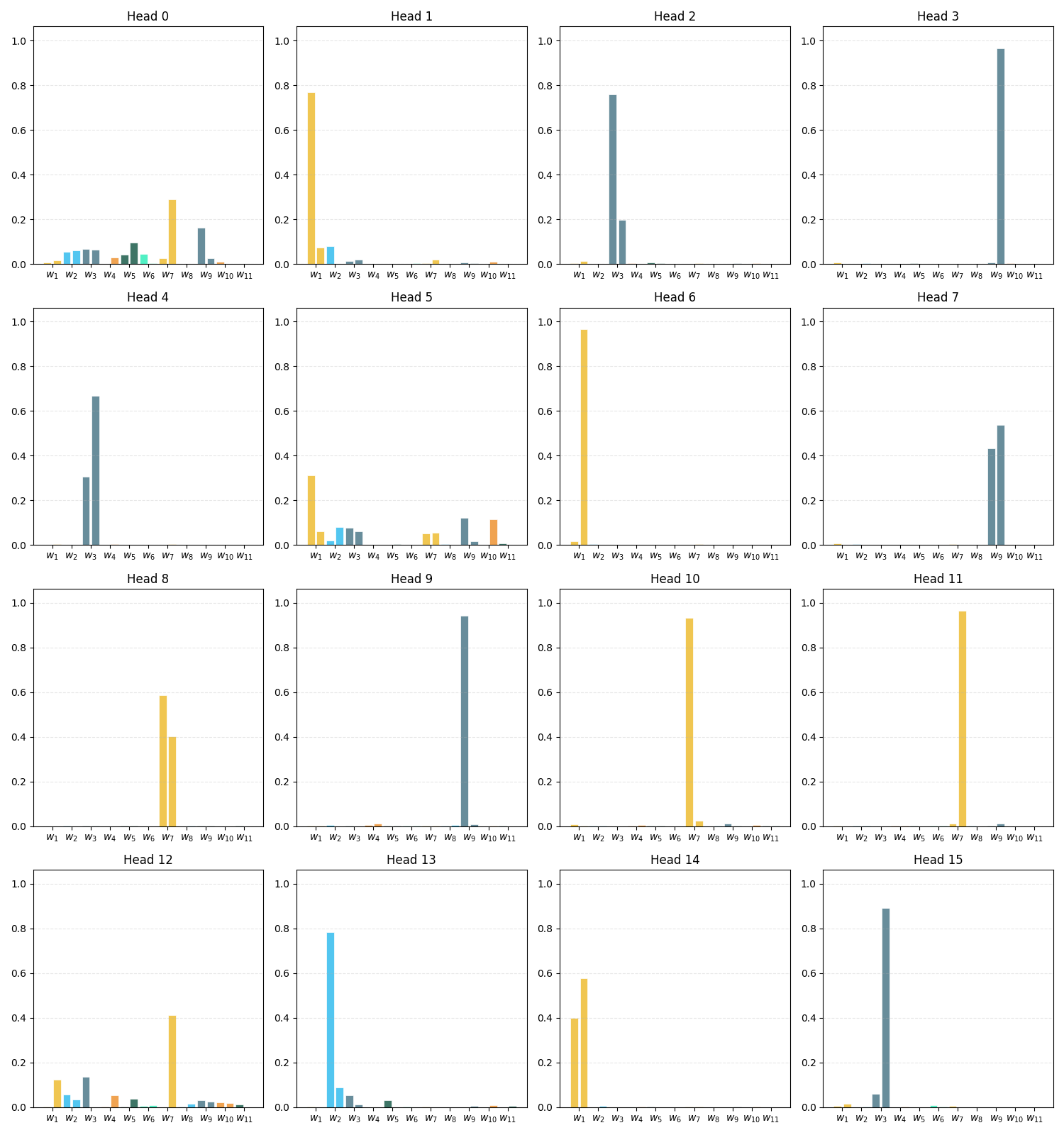}
        \caption{FVE barchart}
    \end{subfigure}
    \begin{subfigure}[b]{0.55\textwidth}
        \centering
        \includegraphics[width=\linewidth]{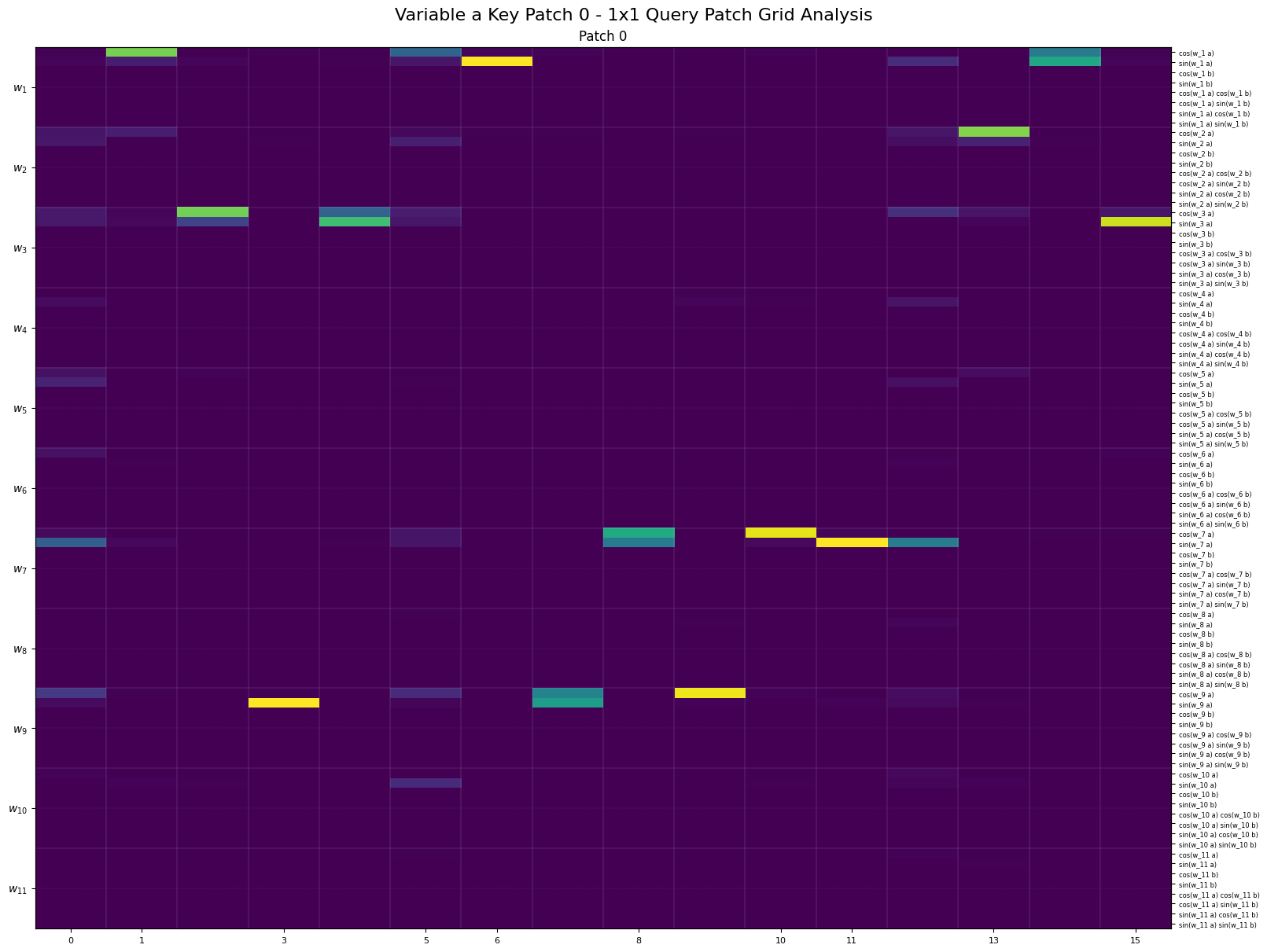}
        \caption{Heatmap of activations in head level}
    \end{subfigure}
    
\caption{
    \textbf{Attention Score Activation of the operand} $a$ 
}
    \label{fig:appendix_attention_score}
\end{figure}

\begin{figure}[H]
    \centering
    \begin{subfigure}[b]{0.35\textwidth}
        \centering
        \includegraphics[width=\linewidth]{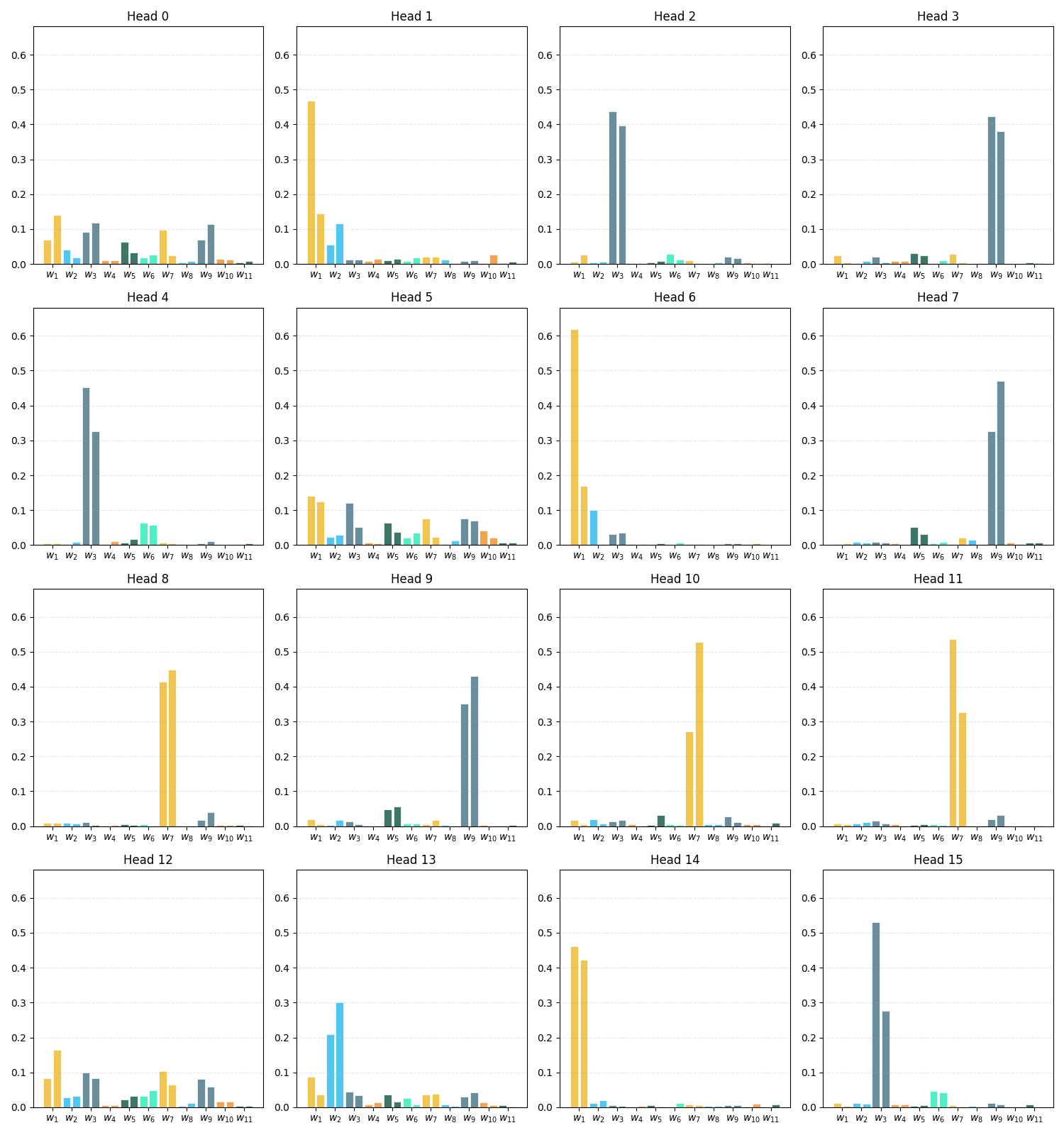}
        \caption{FVE barchart}
    \end{subfigure}
    \begin{subfigure}[b]{0.55\textwidth}
        \centering
        \includegraphics[width=\linewidth]{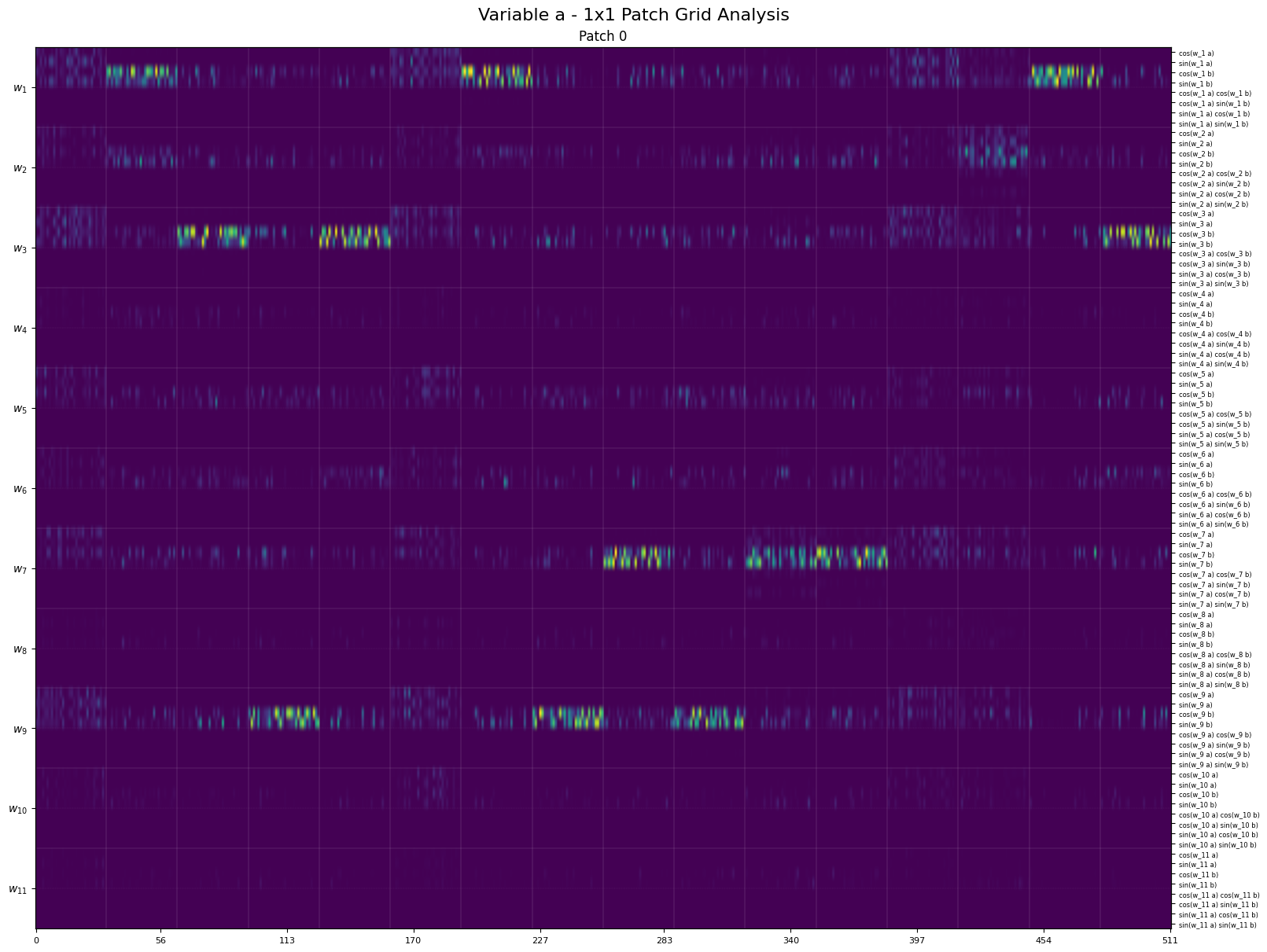}
        \caption{Heatmap of activations in neuron and head level}
    \end{subfigure}
    
\caption{
    \textbf{Attention $\mA\vv$ Activation of the operand} $a$ 
}
    \label{fig:appendix_attention_Av_operand_a}
\end{figure}

The self-attention mechanism combines activations associated with individual operands, $\mA\vv[a]$ and $\mA\vv[b]$. At the result position $c$, we observe both 1D and 2D Fourier components, as illustrated in the following graphs. Thus, this layer contains a mixture of operand-specific and interaction terms rather than a fully isolated 2D representation.

\begin{figure}[H]
    \centering
    \begin{subfigure}[b]{0.49\textwidth}
        \centering
        \includegraphics[width=\linewidth]{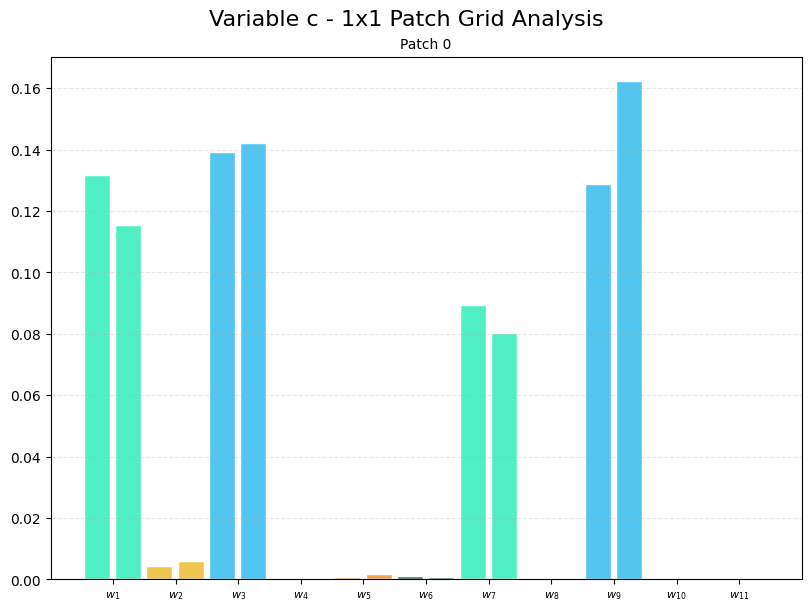}
        \caption{FVE barchart}
    \end{subfigure}
    \hfill 
    \begin{subfigure}[b]{0.49\textwidth}
        \centering
        \includegraphics[width=\linewidth]{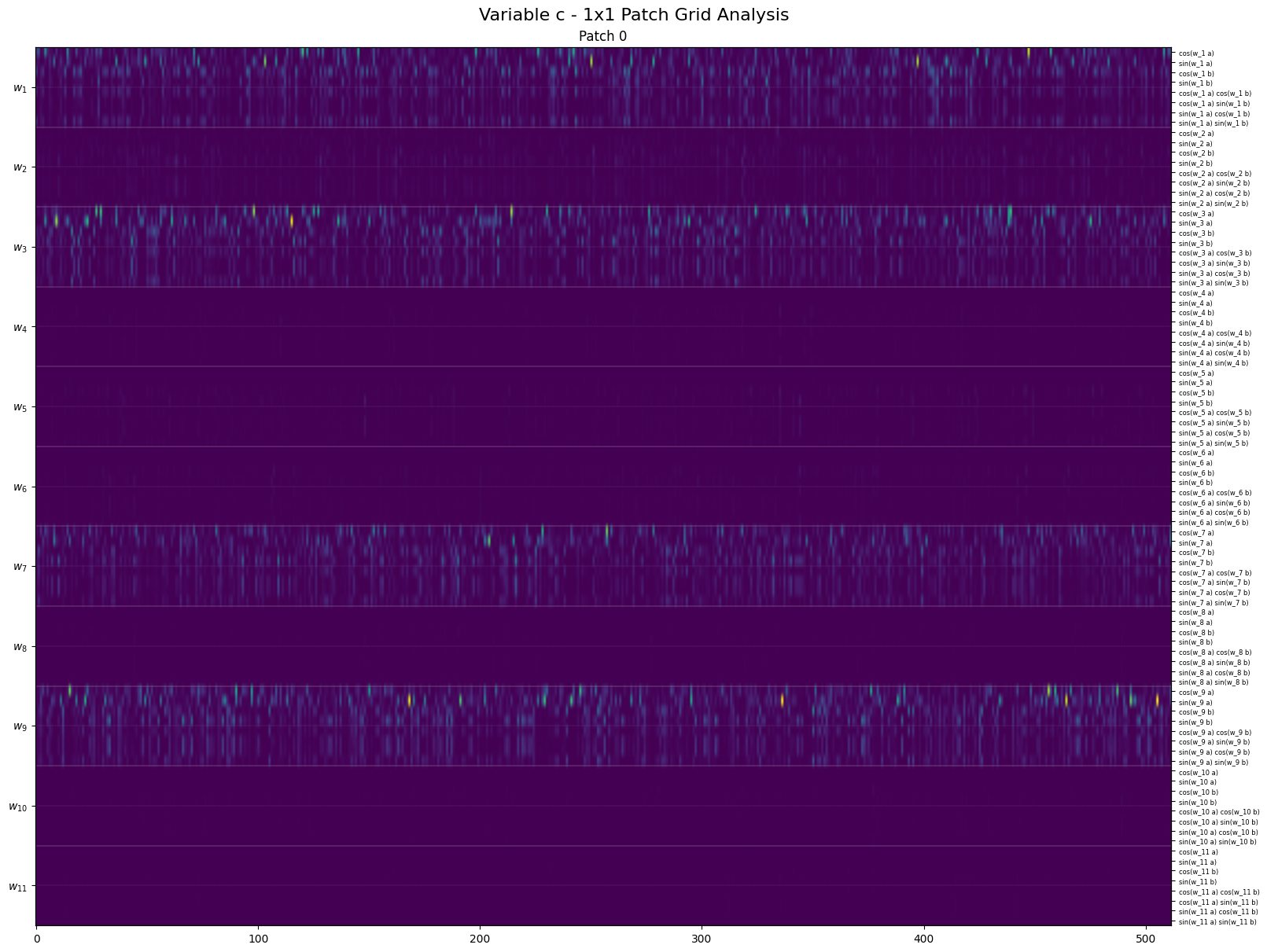}
        \caption{Heatmap of activations in neuron and head level}
    \end{subfigure}
    
\caption{
    \textbf{Attention Out Activation of the operation result} $c$ 
}
    \label{fig:appendix_attention_out_c}
\end{figure}

Finally, at the pre-Gelu activation stage of the FFN layer, the activation at position $c$ is clearly composed of 2D FVEs corresponding to the selective frequencies shared throughout the preceding SA block. This characteristic motivated our focus on this specific layer, as it structurally represents the emergence of arithmetic generalization.

\begin{figure}[H]
    \centering
    \begin{subfigure}[b]{0.49\textwidth}
        \centering
        \includegraphics[width=\linewidth]{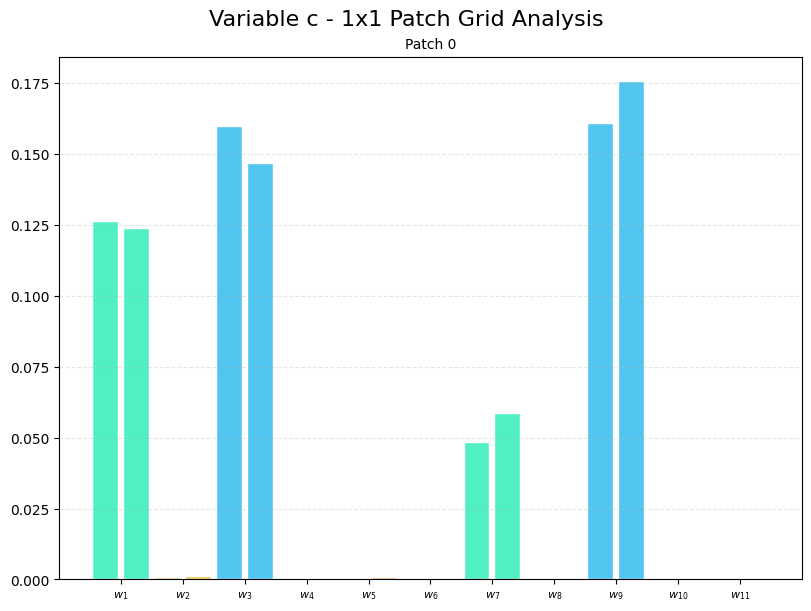}
        \caption{FVE barchart}
    \end{subfigure}
    \hfill 
    \begin{subfigure}[b]{0.49\textwidth}
        \centering
        \includegraphics[width=\linewidth]{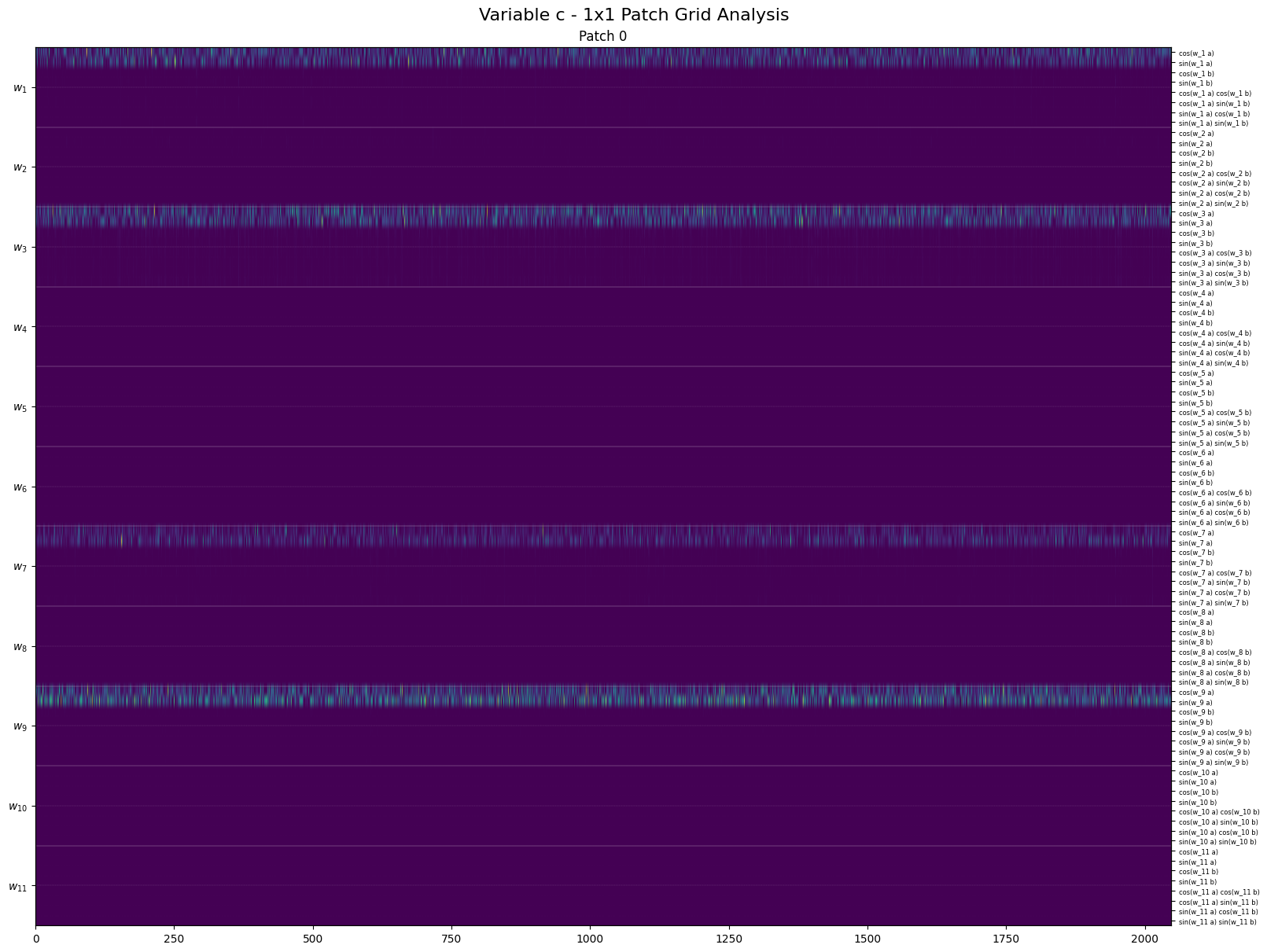}
        \caption{Heatmap of activations in neuron and head level}
    \end{subfigure}
    
\caption{
    \textbf{FFN Pre-GeLU Activation of the operation result} $c$ 
}
    \label{fig:appendix_ffn_pre_act_c}
\end{figure}

At this layer, we derive target-image readout directions and recover the underlying trigonometric identities, since the activation exhibits the strongest degree-two periodic structure. The following table presents the complete set of projected functions, including both the selective frequencies and the non-significant frequencies.

\begin{table}[H]
\centering
\small
\caption{
    \textbf{Fourier Analysis of FFN Pre-activations.}
    Projection of $\mathrm{FFNPreGELU}_c$ onto target-image readout directions yields scalar functions well explained by the trigonometric addition identities in Eq.~\ref{trigonometric_addition_identities}.
}
\label{tab:fourier_projection_full}
\resizebox{0.7\linewidth}{!}{
\begin{tabular}{ccc} 
\toprule
Projected channel & Fourier expansion of projected function & FVE  \\
\midrule
$\cos(w_{1} (a+b))$&$138910     \cos(w_1 a)  \cos(w_1 b)  -139849    \sin(w_1 a)  \sin(w_1 b)$ & $\textbf{0.95}$\\
$\sin(w_{1} (a+b))$&$137133     \cos(w_1 a)  \sin(w_1 b)  +136206     \sin(w_1 a)  \cos(w_1 b)$ & $\textbf{0.94}$\\
$\cos(w_{2} (a+b))$&$939        \cos(w_2 a)  \cos(w_2 b)  -426       \sin(w_2 a)  \sin(w_2 b)$ & $0.01$\\
$\sin(w_{2} (a+b))$&$1017       \cos(w_2 a)  \sin(w_2 b)  +1081       \sin(w_2 a)  \cos(w_2 b)$ & $0.01$\\
$\cos(w_{3} (a+b))$&$171404     \cos(w_3 a)  \cos(w_3 b)  -181490    \sin(w_3 a)  \sin(w_3 b)$ & $\textbf{0.95}$\\
$\sin(w_{3} (a+b))$&$163861     \cos(w_3 a)  \sin(w_3 b)  +160307     \sin(w_3 a)  \cos(w_3 b)$ & $\textbf{0.93}$\\
$\cos(w_{4} (a+b))$&$168        \cos(w_4 a)  \cos(w_4 b)  -168       \sin(w_4 a)  \sin(w_4 b)$ & $0.00$\\
$\sin(w_{4} (a+b))$&$74         \cos(w_4 a)  \sin(w_4 b)  +83         \sin(w_4 a)  \cos(w_4 b)$ & $0.00$\\
$\cos(w_{5} (a+b))$&$315        \cos(w_5 a)  \cos(w_5 b)  -410       \sin(w_5 a)  \sin(w_5 b)$ & $0.00$\\
$\sin(w_{5} (a+b))$&$403        \cos(w_5 a)  \sin(w_5 b)  +420        \sin(w_5 a)  \cos(w_5 b)$ & $0.00$\\
$\cos(w_{6} (a+b))$&$245        \cos(w_6 a)  \cos(w_6 b)  -117       \sin(w_6 a)  \sin(w_6 b)$ & $0.00$\\
$\sin(w_{6} (a+b))$&$259        \cos(w_6 a)  \sin(w_6 b)  +294        \sin(w_6 a)  \cos(w_6 b)$ & $0.00$\\
$\cos(w_{7} (a+b))$&$52852      \cos(w_7 a)  \cos(w_7 b)  -53542     \sin(w_7 a)  \sin(w_7 b)$ & $\textbf{0.87}$\\
$\sin(w_{7} (a+b))$&$64718      \cos(w_7 a)  \sin(w_7 b)  +64358      \sin(w_7 a)  \cos(w_7 b)$ & $\textbf{0.88}$\\
$\cos(w_{8} (a+b))$&$105        \cos(w_8 a)  \cos(w_8 b)  -65        \sin(w_8 a)  \sin(w_8 b)$ & $0.00$\\
$\sin(w_{8} (a+b))$&$87         \cos(w_8 a)  \sin(w_8 b)  +81         \sin(w_8 a)  \cos(w_8 b)$ & $0.00$\\
$\cos(w_{9} (a+b))$&$176727     \cos(w_9 a)  \cos(w_9 b)  -178583    \sin(w_9 a)  \sin(w_9 b)$ & $\textbf{0.95}$\\
$\sin(w_{9} (a+b))$&$192981     \cos(w_9 a)  \sin(w_9 b)  +195516     \sin(w_9 a)  \cos(w_9 b)$ & $\textbf{0.96}$\\
$\cos(w_{10}(a+b))$&$41      \cos(w_{10} a) \cos(w_{10} b) -86    \sin(w_{10} a) \sin(w_{10} b)$ & $0.00$\\
$\sin(w_{10}(a+b))$&$149     \cos(w_{10} a) \sin(w_{10} b) +141   \sin(w_{10} a) \cos(w_{10} b)$ & $0.00$\\
$\cos(w_{11}(a+b))$&$109     \cos(w_{11} a) \cos(w_{11} b) -103   \sin(w_{11} a) \sin(w_{11} b)$ & $0.00$\\
$\sin(w_{11}(a+b))$&$47      \cos(w_{11} a) \sin(w_{11} b) +46    \sin(w_{11} a) \cos(w_{11} b)$ & $0.00$\\
\bottomrule
\end{tabular}
}

\end{table}

Interestingly, the FFN layer's GeLU activation spreads signals that were previously concentrated on selective frequencies across a broader frequency spectrum. This non-linear transformation redistributes arithmetic-related information into a representation space used for final image synthesis.

\begin{figure}[H]
    \centering
    \begin{subfigure}[b]{0.49\textwidth}
        \centering
        \includegraphics[width=\linewidth]{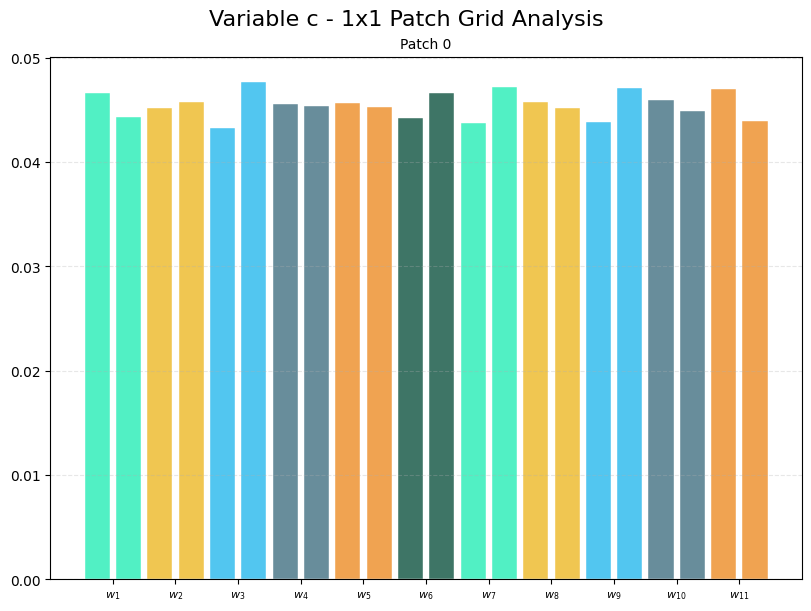}
        \caption{FVE barchart}
    \end{subfigure}
    \hfill 
    \begin{subfigure}[b]{0.49\textwidth}
        \centering
        \includegraphics[width=\linewidth]{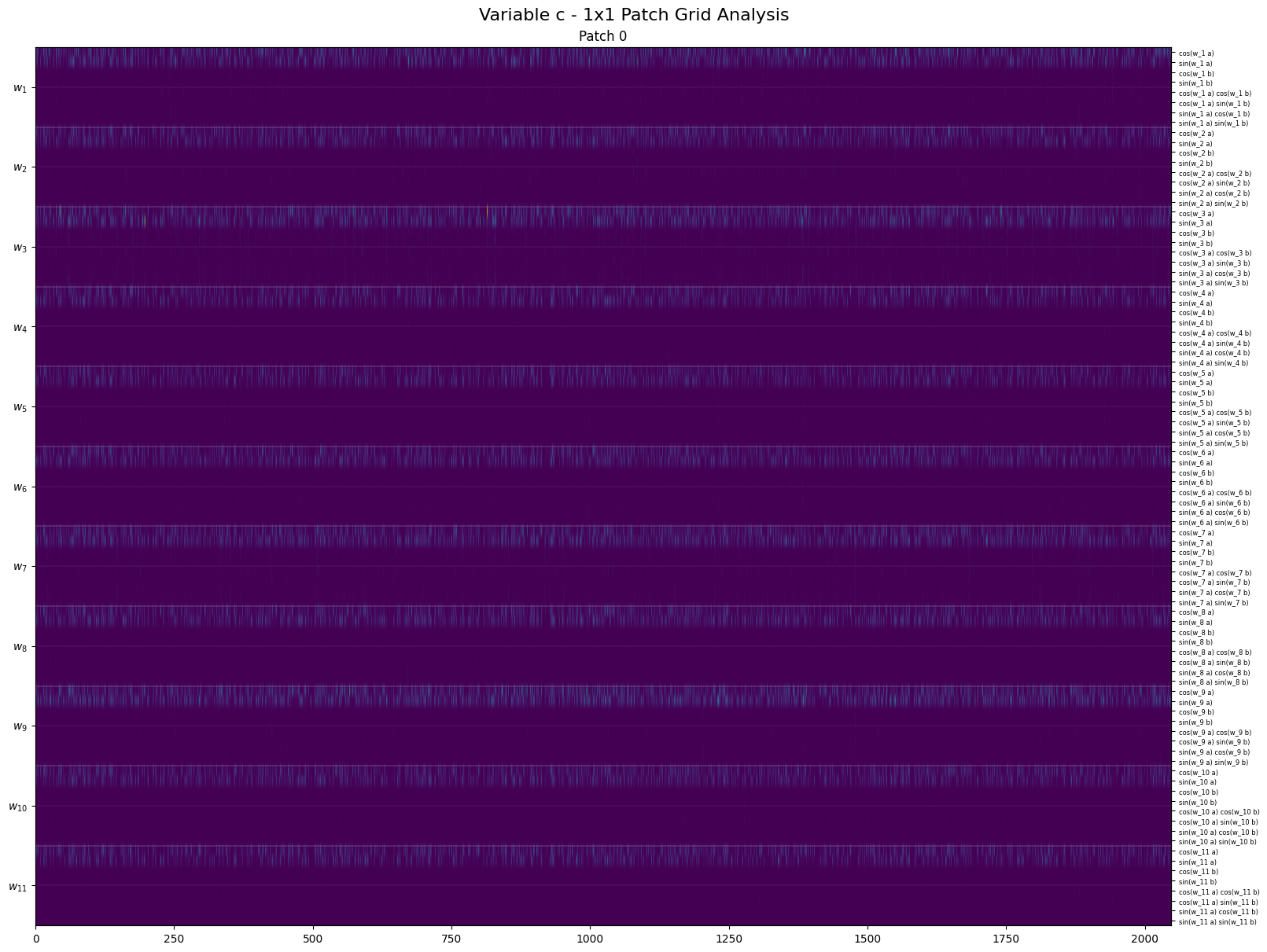}
        \caption{Heatmap of activations in neuron and head level}
    \end{subfigure}
    
\caption{
    \textbf{FFN Post-GeLU Activation of the operation result} $c$ 
}
    \label{fig:appendix_ffn_post_act_c}
\end{figure}

Finally, we present the output activation of the final MLP layer. This activation is unpatchified by the network and mapped back into the image modality. As illustrated, the distinct frequency signals—previously dominant in the internal layers—are no longer present, indicating that the representation has been fully transformed into the spatial domain for final image synthesis.

\begin{figure}[H]
    \centering
    \begin{subfigure}[b]{0.49\textwidth}
        \centering
        \includegraphics[width=\linewidth]{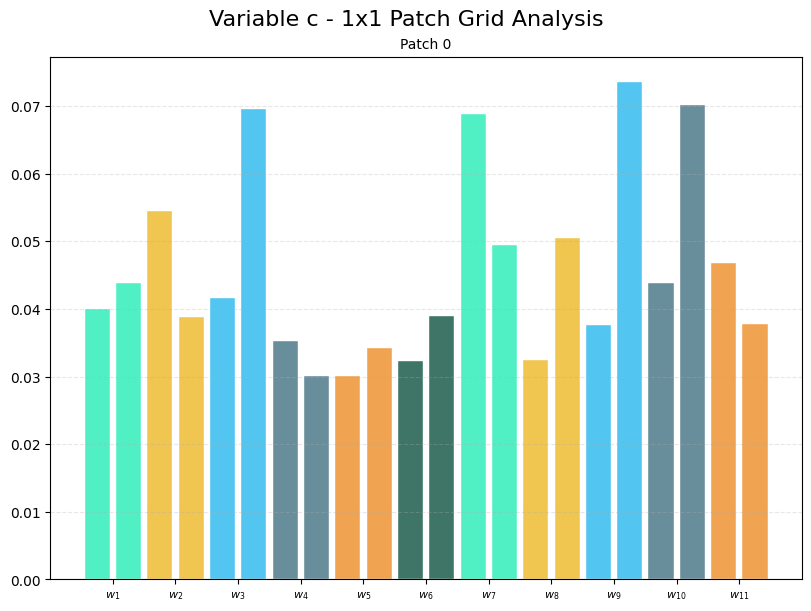}
        \caption{FVE barchart}
    \end{subfigure}
    \hfill 
    \begin{subfigure}[b]{0.49\textwidth}
        \centering
        \includegraphics[width=\linewidth]{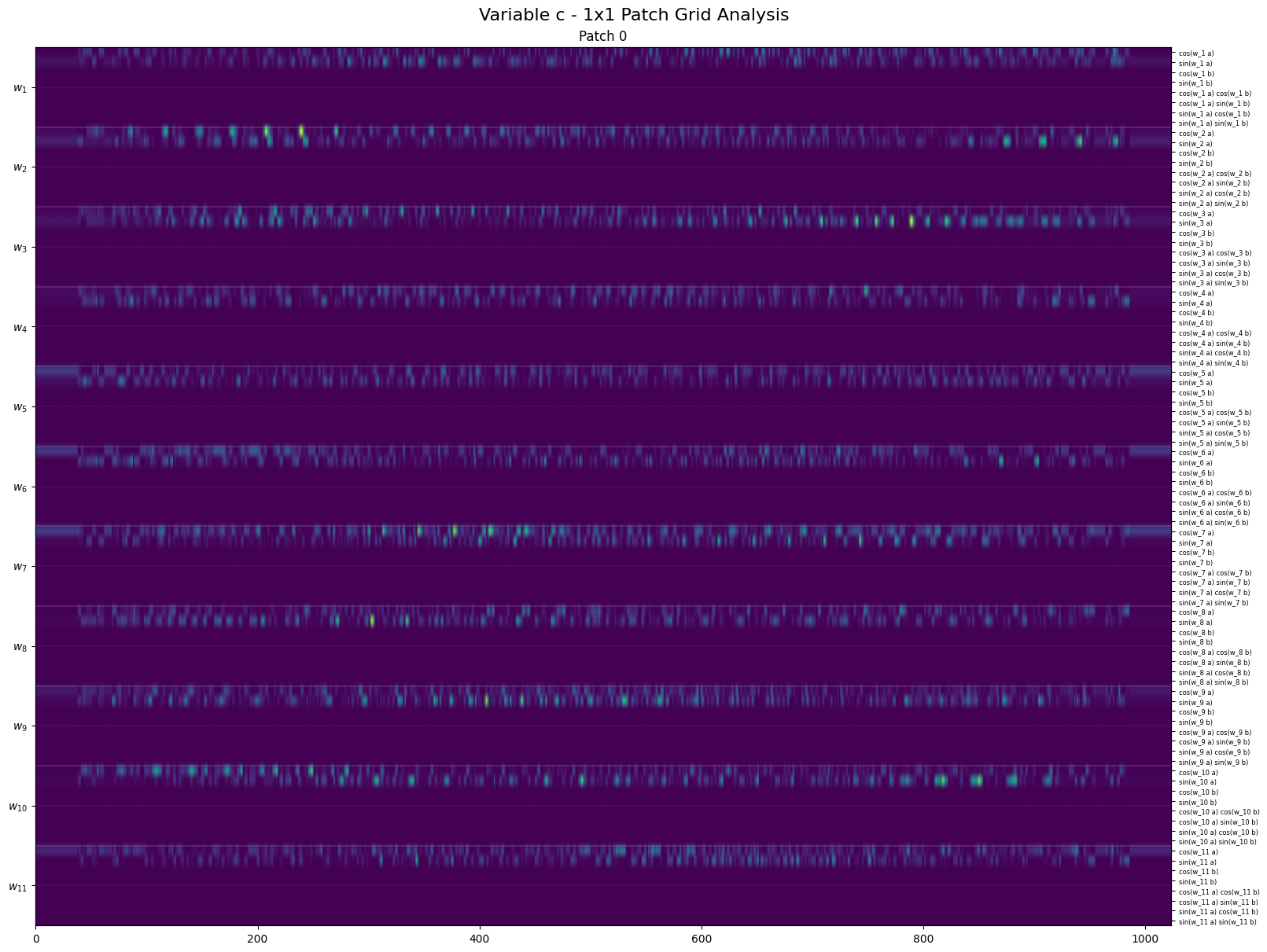}
        \caption{Heatmap of activations in neuron and head level}
    \end{subfigure}
    
\caption{
    \textbf{Final MLP Layer's output activation} on result position $c$ 
}
    \label{fig:appendix_final_out_c}
\end{figure}

\section{Multiple-Image Regime Detail}
\subsection{FFN Sandwich architecture}
\label{appendix:ffn_sandwich_justification}
To facilitate the mapping of high-dimensional visual inputs into a class-aligned representation space, we adopt an auxiliary FFN layer situated between the embedding layer and the Self-Attention (SA) block. PCA visualizations show that this "sandwich" architecture maps diverse pixel-level inputs into clearer label-specific clusters early in the model's processing pipeline.

\begin{figure}[H]
\centering
    \includegraphics[width=0.8\linewidth]{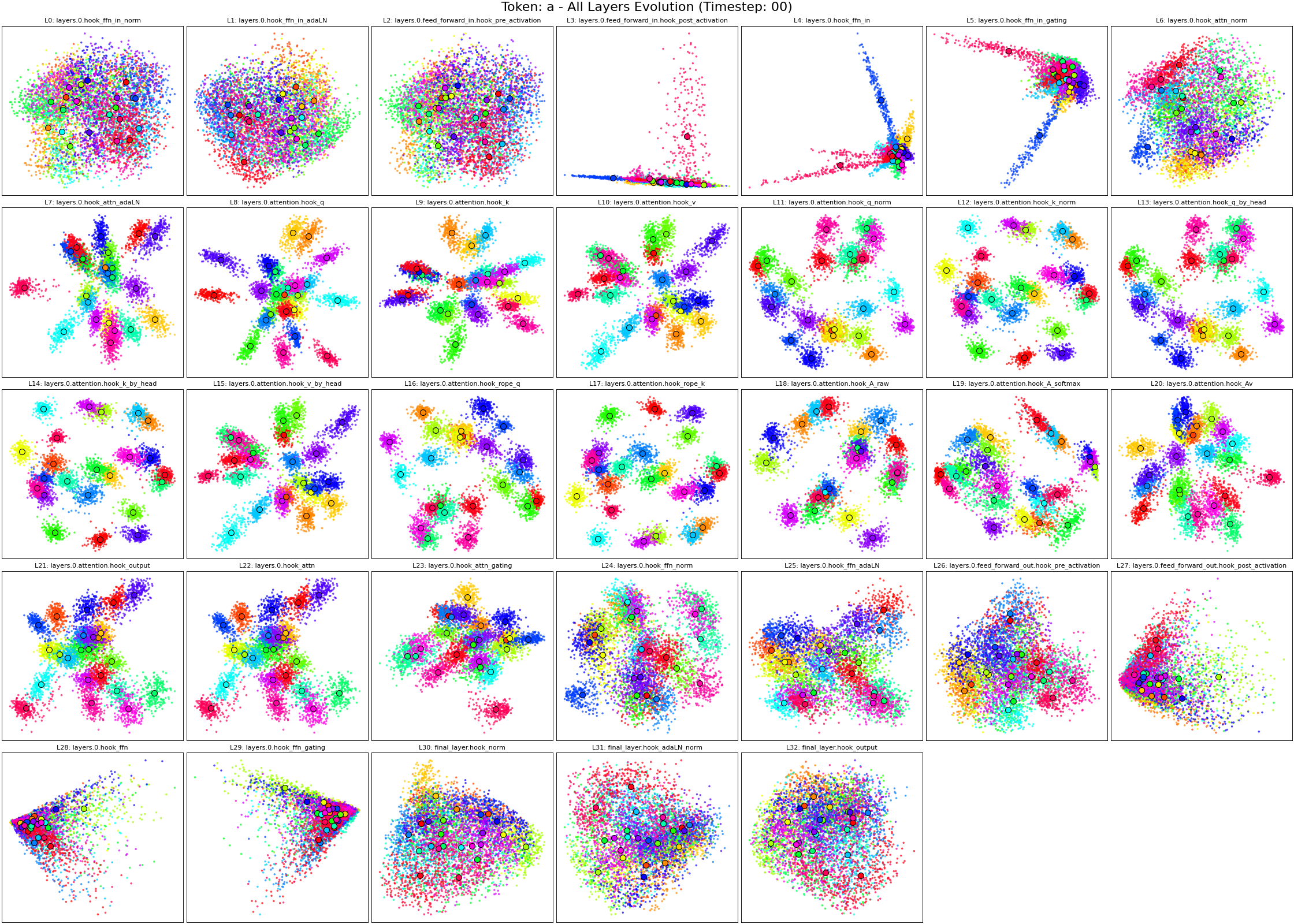}
    \caption{PCA at the operand $a$ position at timestep = 0}
    \label{fig:ffn_sandwich_pca1}
\end{figure}

Additionally, in Appendix~\ref{appendix:ablations_kuzushiji}, we provide an ablation study on the heterogeneous Kuzushiji-MNIST dataset \citep{DBLP:journals/corr/abs-1812-01718}. Because this dataset contains relatively more complex spatial structures compared to the original EMNIST dataset, the recurrence of selective Fourier structure and mode-shift behavior there shows that the FFN-sandwich architecture can produce class-aligned periodic representations beyond the baseline EMNIST setting.

\begin{figure}[H]
\centering
    \includegraphics[width=0.8\linewidth]{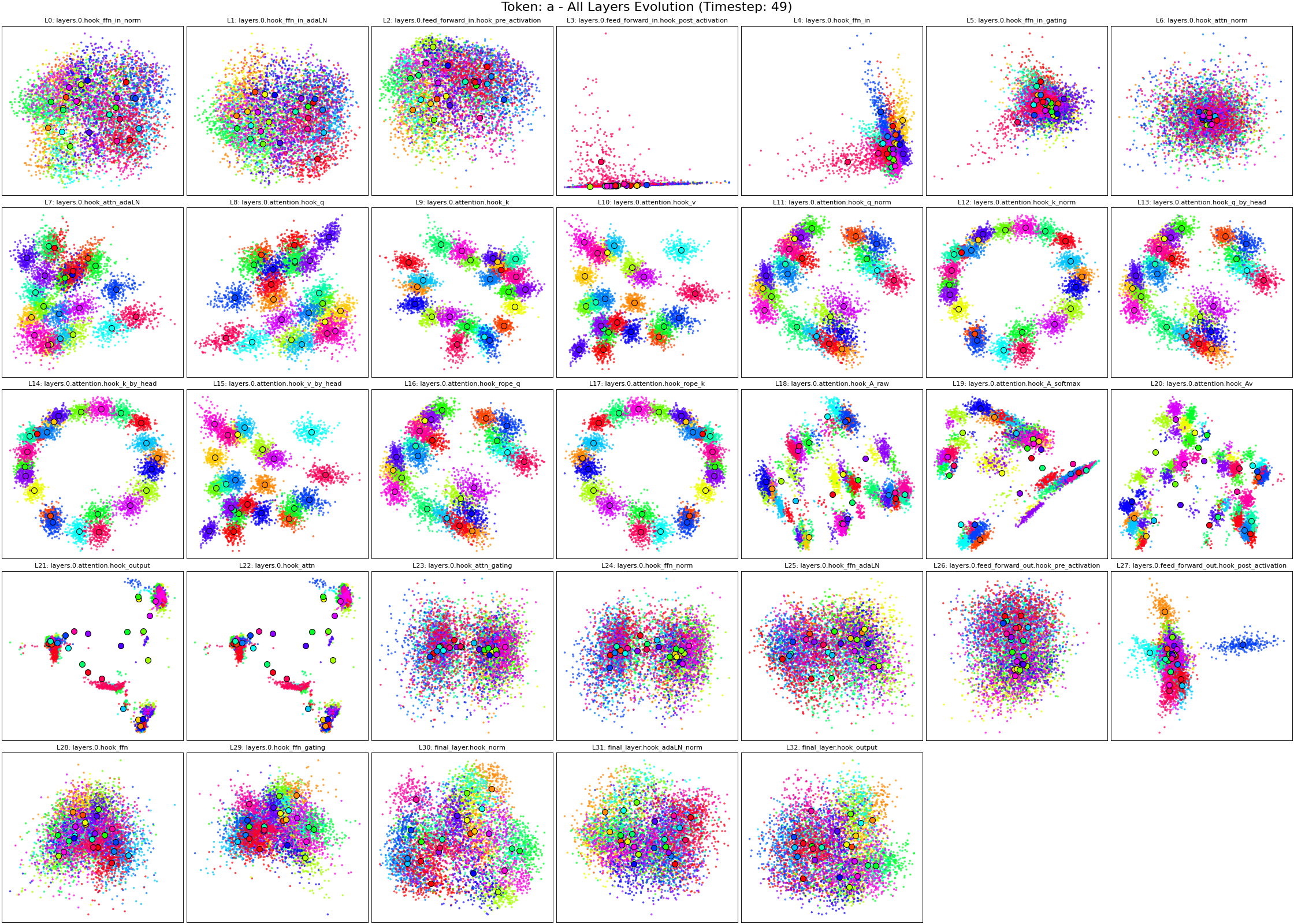}
    \caption{PCA at the operand $a$ position at timestep = 50}
    \label{fig:ffn_sandwich_pca2}
\end{figure}

Beyond the PCA evidence for class separation, these visualizations reveal the geometry underlying the periodic structures discussed in our Fourier analysis. As shown in Figure~\ref{fig:ffn_sandwich_pca1}, the representations at the position of operand $a$ exhibit distinct clustering by modular labels from the initial timestep ($t=0$), and these clusters persist throughout the denoising trajectory. When projected into the subsequent Attention Key space, these activations form a ring-structured geometry (Figure~\ref{fig:ffn_sandwich_pca2}). This circular structure matches the frequency-domain findings and shows that the model embeds modular classes into a periodic representation space used for arithmetic computation.

\subsection{Construction of the Evaluation Dataset}
\label{appendix:multiple_image_dataset_construction}
To evaluate the model's algorithmic generalization beyond repeated visual mappings, we constructed a large, non-redundant evaluation dataset for the $N=256$ regime. The construction process was designed to ensure arithmetic coverage and visual diversity:

    \begin{enumerate}
        \item \textbf{Visual Diversity:} For each label in the modular arithmetic group $\mathbb{Z}_P$, we utilized a pool of 256 unique, high-confidence EMNIST images.
        \item \textbf{Non-redundant Pairing:} To prevent the model from memorizing specific image pairs, we implemented a shuffling-and-sampling mechanism. For each possible arithmetic combination $(a, b)$, images were drawn from their respective pools without replacement until all 256 images per label were consumed.
        \item \textbf{Total Sample Volume:} This process resulted in a total of $11 \times P^2$ unique evaluation pairs (e.g. 5,819 pairs for the $P=23$ baseline case), where each pair represents a distinct visual instantiation of the underlying modular addition.
    \end{enumerate}

\subsection{Inference Performance}
\label{appendix:multiple_image_inference_results}
On this high-confidence evaluation dataset, the diffusion model achieved an inference accuracy of 95\%. This number is comparable to the ResNet-18 classifier's accuracy on the full EMNIST test set, but the diffusion evaluation itself uses filtered samples selected to reduce classifier ambiguity. The high accuracy on non-repeated visual combinations shows that the model has learned the abstract modular rule rather than only memorizing image pairs.

We visualize the sampling process across discrete time steps. Following convention, we denote $t=1$ as the initial noise state and $x_0$ as the final synthesized image. The interval $[1, 0]$ is discretized into 50 uniform steps to illustrate the transition. For clarity in the visualization below, we index these steps as $T=0$ (initial noise) to $T=50$ (final image). Prediction accuracy is evaluated using a ResNet classifier pre-trained on the EMNIST dataset. Red, yellow, and green bounding boxes indicate that the classifier identifies the generated image as incorrect, low-confidence, or correct, respectively.

\begin{figure}[H]
\centering
    \includegraphics[width=0.90\linewidth]{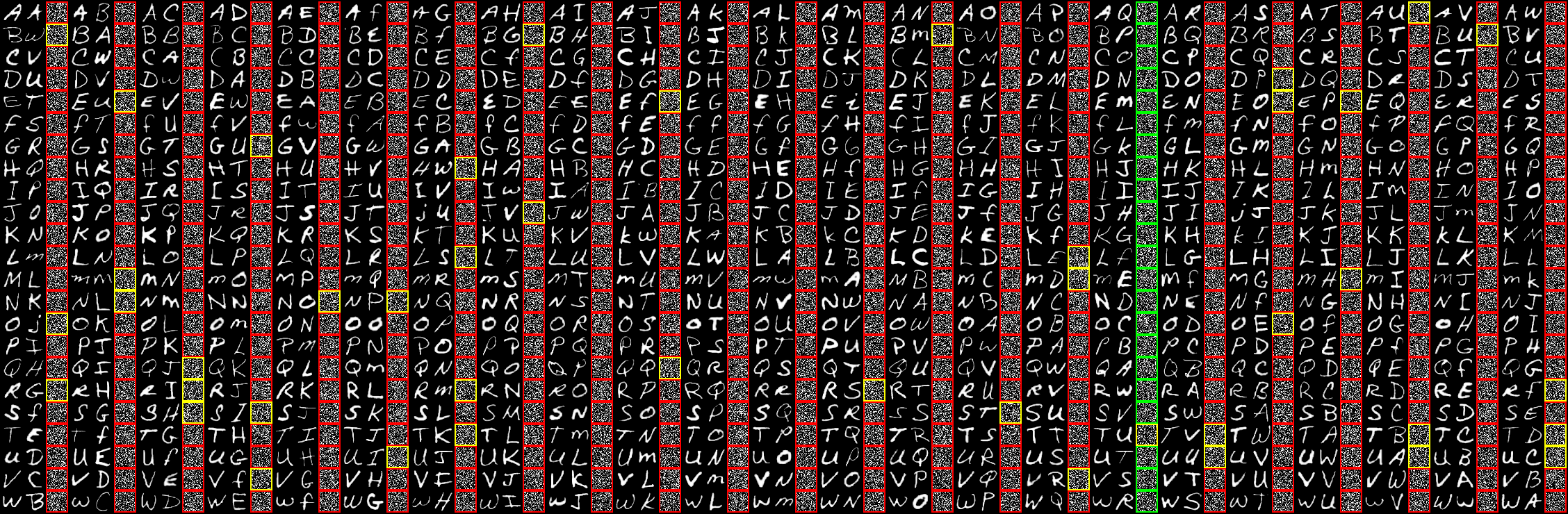}
    \caption{
    Timestep 0 with accuracy 0\%
    }
    \label{fig:appendix_sampling_step_000}
\end{figure}

\begin{figure}[H]
\centering
    \includegraphics[width=0.90\linewidth]{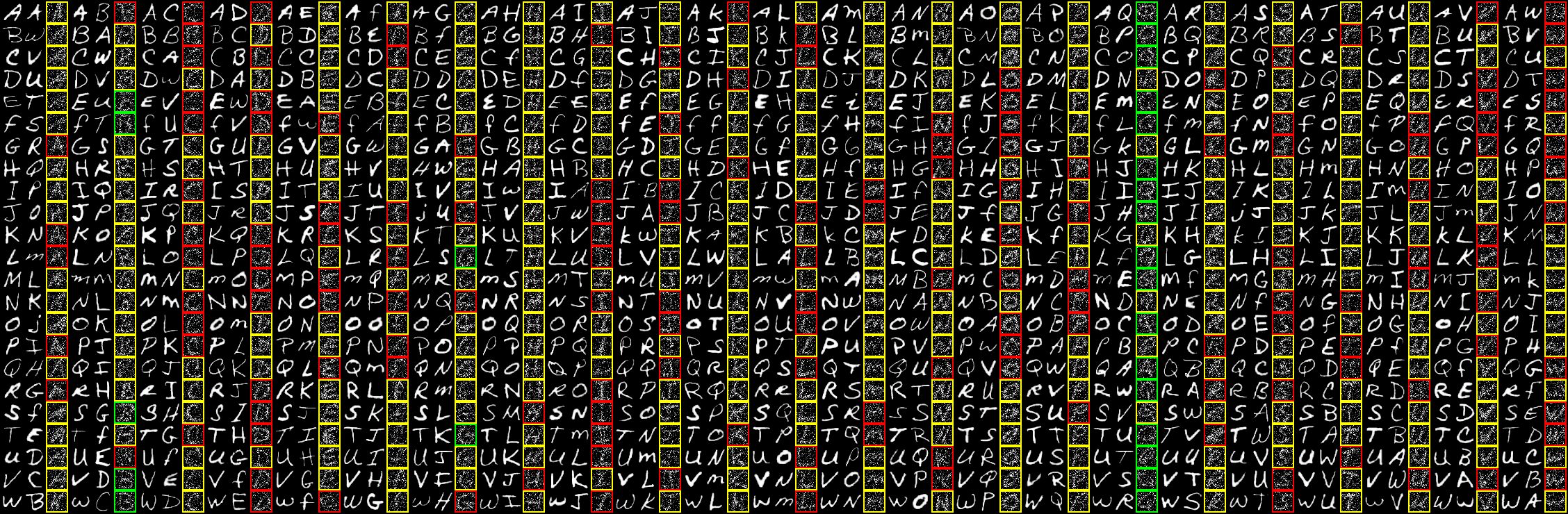}
    \caption{
    Timestep 21 with accuracy 7\%
    }
    \label{fig:appendix_sampling_step_021}
\end{figure}

\begin{figure}[H]
\centering
    \includegraphics[width=0.90\linewidth]{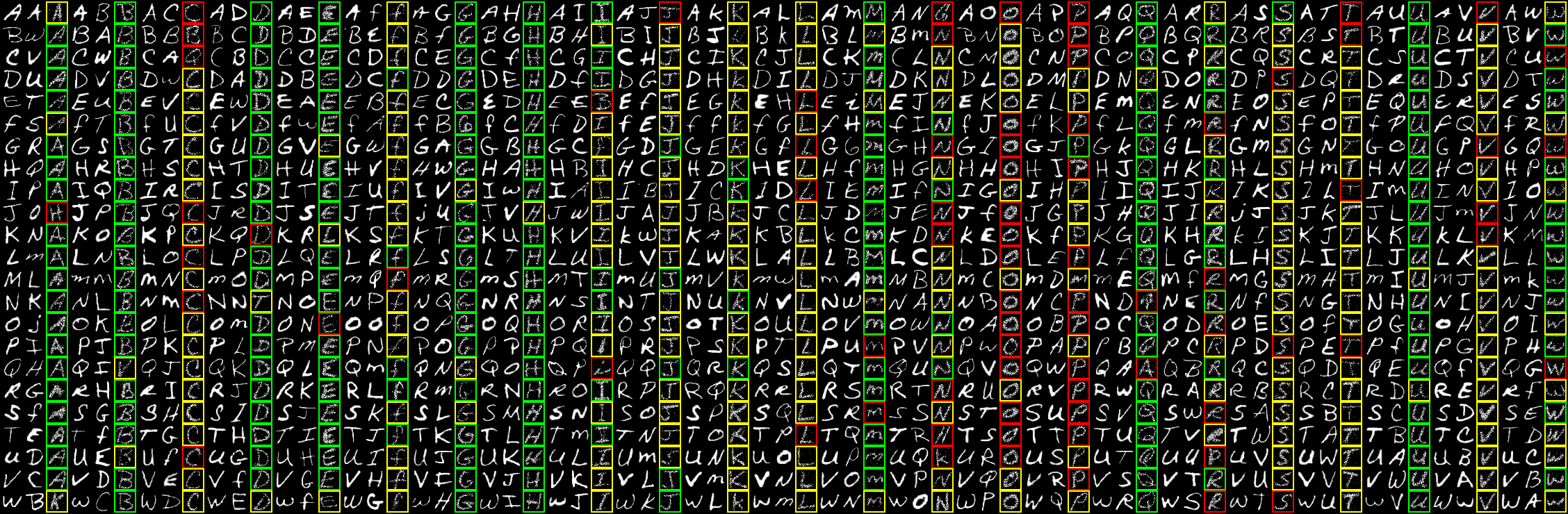}
    \caption{
    Timestep 32 with accuracy 54\%
    }
    \label{fig:appendix_sampling_step_032}
\end{figure}

\begin{figure}[H]
\centering
    \includegraphics[width=0.90\linewidth]{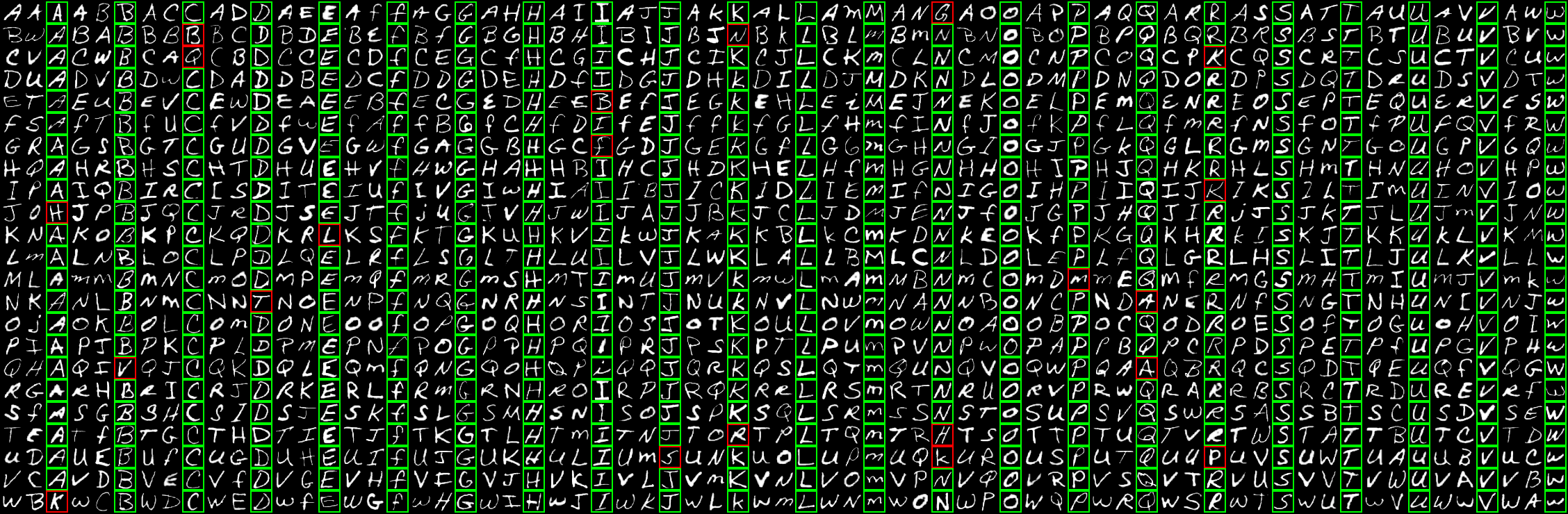}
    \caption{
    Timestep 50 with accuracy 96\%
    }
    \label{fig:appendix_sampling_step_050}
\end{figure}

\section{FVE Entropy Trajectory}
\label{appendix:fve_entropy_trajectory}
\begin{figure}[H]
    \centering
    \begin{subfigure}[b]{0.49\textwidth}
        \centering
        \includegraphics[width=\linewidth]{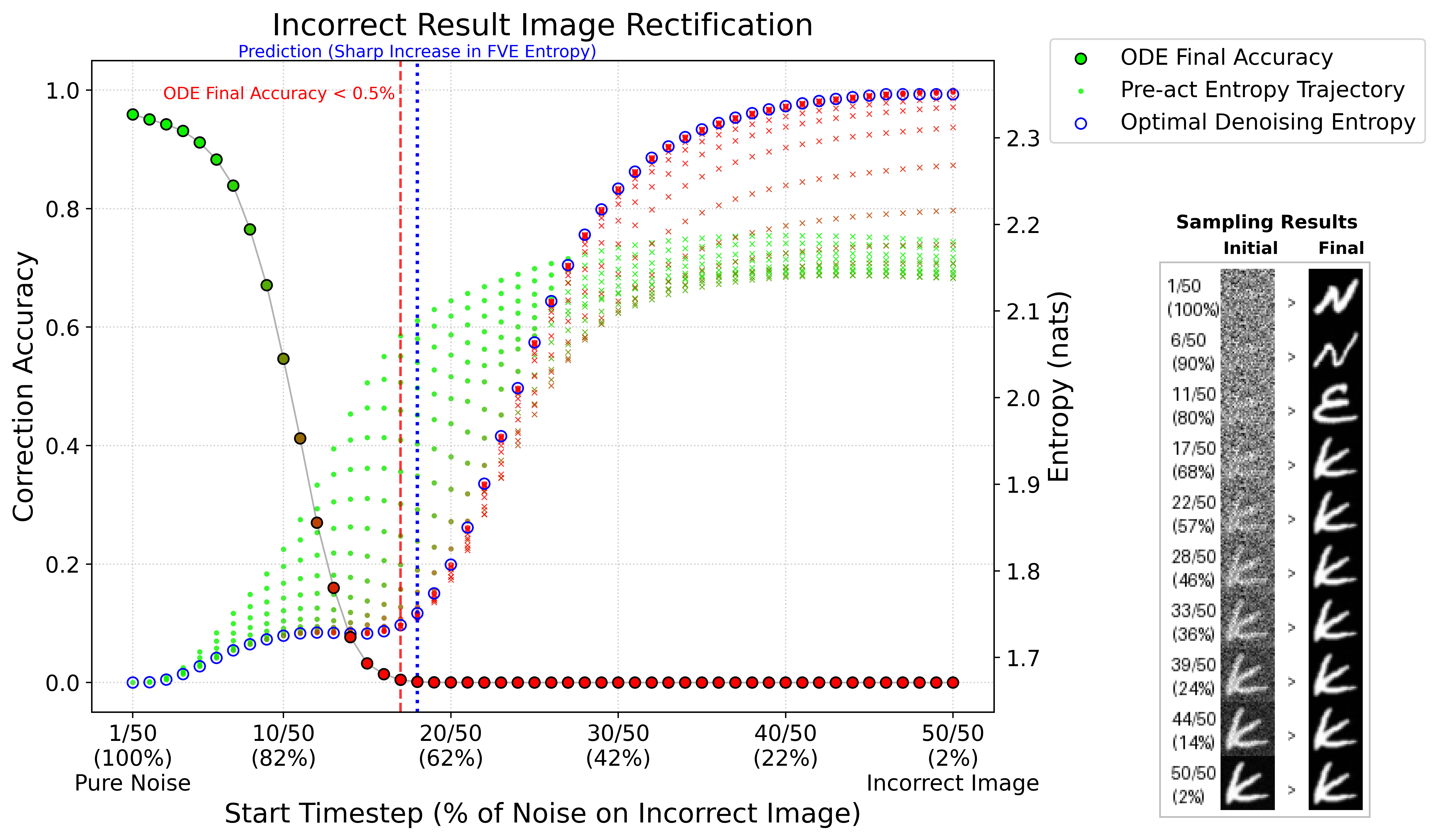}
        \caption*{Incorrect result image rectification}
    \end{subfigure}
    \hfill 
    \begin{subfigure}[b]{0.49\textwidth}
        \centering
        \includegraphics[width=\linewidth]{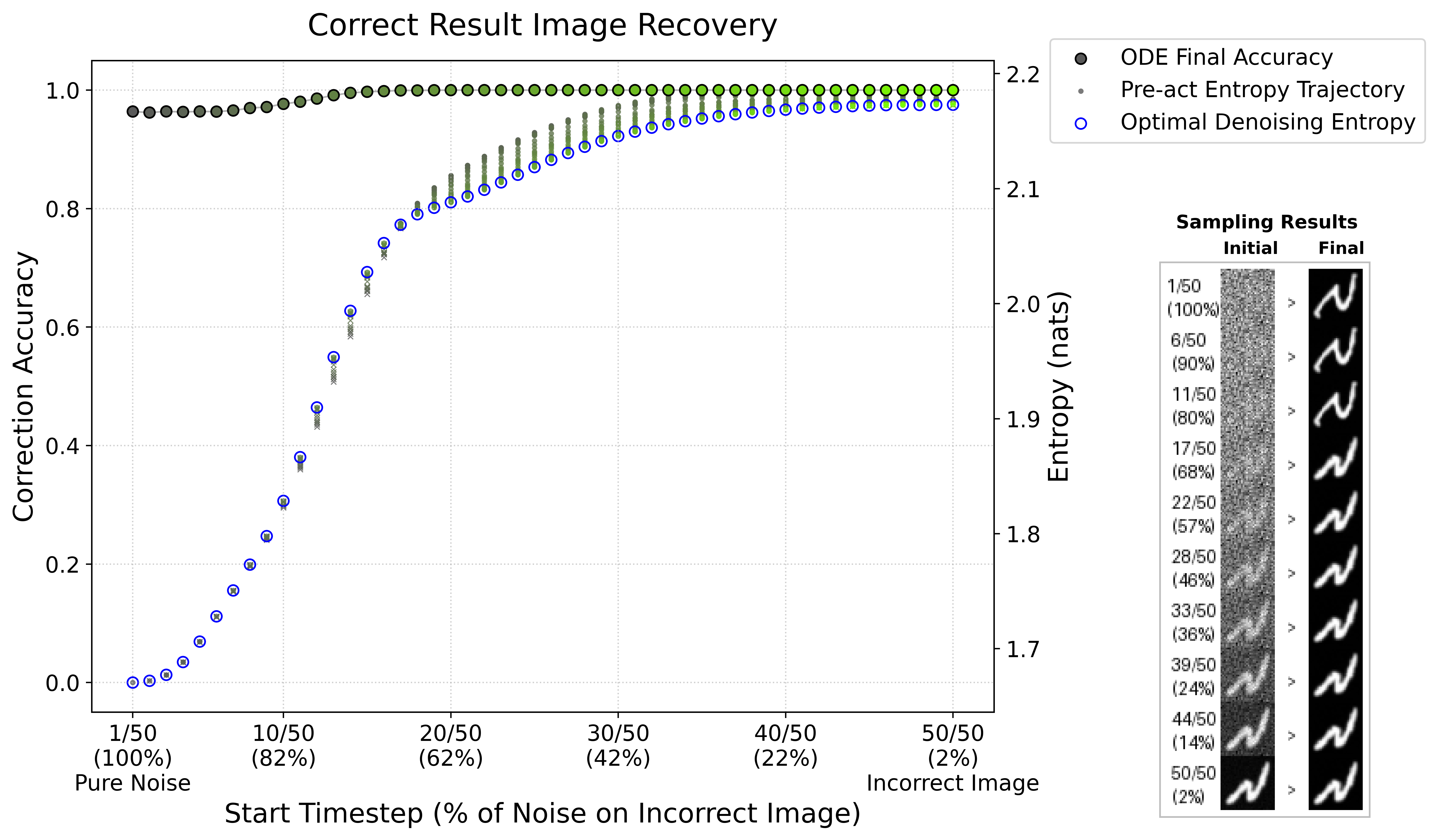}
        \caption*{Correct result image recovery}
    \end{subfigure}    
\caption{
    Intervention experiment dynamics for $P=23$ on the EMNIST dataset. \textbf{(Left)} The raw FVE entropy trajectory for the incorrect image rectification experiment is plotted with blue circle markers. Corresponding to the gradual collapse of the periodic structure, the FVE entropy increases as the sampling timestep approaches the final step. The prediction point (vertical blue line) is again derived from the maximum second derivative, which captures the sudden spike in the FVE entropy. \textbf{(Right)} When provided with the correct image of class $c$ perturbed with noise proportional to the ODE sampling start timestep $\hat{t}$, the model successfully recovers the perturbed images, achieving over 90\% accuracy for all $\hat{t}$. Nevertheless, the gradual collapse of the periodic structure---revealed by the increase in FVE entropy---emerges again, supporting the existence of a phase transition between arithmetic computation and visual denoising. 
}
    \label{fig:app_mode_shift_p_23_emnist}
\end{figure}

In Section~\ref{subsec:multi_step}, we provide the second derivative trajectory of the FVE entropy for the FFN pre-GeLU layer's activations in Figure~\ref{fig:2nd_derivative_and_prediction}. Figure~\ref{fig:app_mode_shift_p_23_emnist} (Left) shows the raw FVE entropy trajectory for the incorrect image rectification experiment. Figure~\ref{fig:app_mode_shift_p_23_emnist} (Right) illustrates a similar experiment where the perturbed correct image is provided to evaluate the model's recovery capability. In both cases, an increase in entropy is observed, indicating the collapse of the periodic structure within the model. Additionally, we provide the initial and final images for both the recovery and rectification tasks in the lower right corner of each figure.

\section{Alternative Statistic for Periodic Representations}
\label{appendix:alternative_statistic_periodic}
While we utilize FVE entropy to quantify the spectral concentration on a few frequencies, this Fourier-based analysis is inherently tied to tasks with explicit periodic structures, such as modular addition. To explore the generality of our findings beyond periodic tasks, we investigate the low-rank properties of the model's representations. Assuming that the dominance of a few significant frequencies reflects an underlying low-rank structure, we evaluate an alternative, domain-agnostic statistic: the \textit{stable rank} \citep{rudelson2006samplinglargematricesapproach}. 

For a matrix $\mathbf{A}$, the stable rank is defined as $\|\mathbf{A}\|_F^2 / \|\mathbf{A}\|_2^2$, where $\|\cdot\|_F$ and $\|\cdot\|_2$ denote the Frobenius and spectral norms, respectively. Unlike the Von Neumann entropy, which depends on the entire singular value spectrum, the stable rank places more weight on the most significant singular value, providing a more robust measure of numerical low-rankness. In our $P=23$ EMNIST experiments, the stable rank exhibits a trajectory remarkably similar to that of the FVE entropy. This alternative metric further supports the existence of a structural phase transition during the sampling process from a generalized linear-algebraic perspective.

\begin{figure}[H]
\centering
    \includegraphics[width=0.9\linewidth]{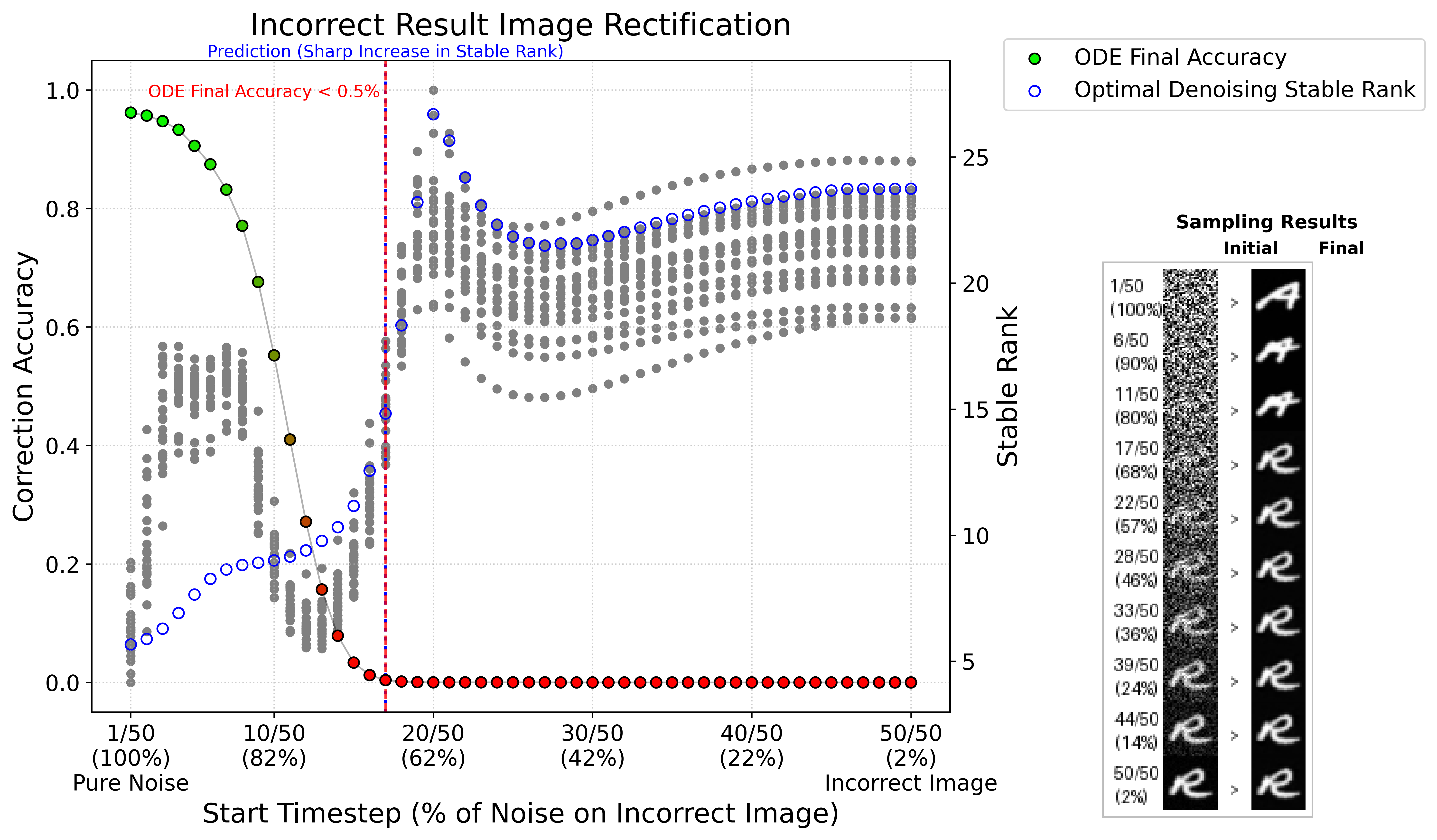}
    \caption{
    Analogous to the FVE entropy analysis in Figure~\ref{fig:app_mode_shift_p_23_emnist}, we plot the stable rank trajectory of the FFN pre-GeLU activations during the incorrect image rectification experiment. The blue dotted trajectory represents the average stable rank across all $P=23$ classes, while the gray dotted lines show the trajectories for individual classes ($0, 1, \ldots, 22$). Both illustrate a gradual increase in this statistic, indicating a breakdown of the low-rank structure at later sampling timesteps.
    }
    \label{fig:appendix_stable_rank_p_23_emnist}
\end{figure}

Figure~\ref{fig:appendix_stable_rank_p_23_emnist} displays a gradual increase in the stable rank of the representations, closely mirroring the FVE entropy dynamics. Beyond its domain-agnostic versatility, a notable advantage of the stable rank is that it can be computed independently for individual classes (Figure~\ref{fig:appendix_stable_rank_p_23_emnist}, gray dotted trajectories). In contrast, computing the FVE requires the full set of classes to construct the Fourier basis.

\section{Multiple Image Fourier Analysis Results}
\label{appendix:multiple_image_fourier_analysis}

We extend our Fourier analysis to the multiple-image (multiple-step) scenario, introducing the temporal dimension (timestep) into our investigation. Following the observations in the single-image case, we concentrate on the layers critical for arithmetic generalization: the SA block and the output FFN pre-GeLU activation of the FFN-sandwich block.

A notable departure from the single-image baseline is the emergence of five significant frequencies, compared to the four previously observed. In the multiple-image setting, these selective frequencies exhibit high significance from the initial timestep within both the attention score and output FFN pre-GeLU activations. Conversely, the attention value layer displays an opposing trend, where frequency significance evolves differently over time. We provide FVE bar charts and neuron-level heatmaps for these key layers, alongside the corresponding sampling trajectories.

\begin{figure}[H]
\centering
    \includegraphics[width=0.9\linewidth]{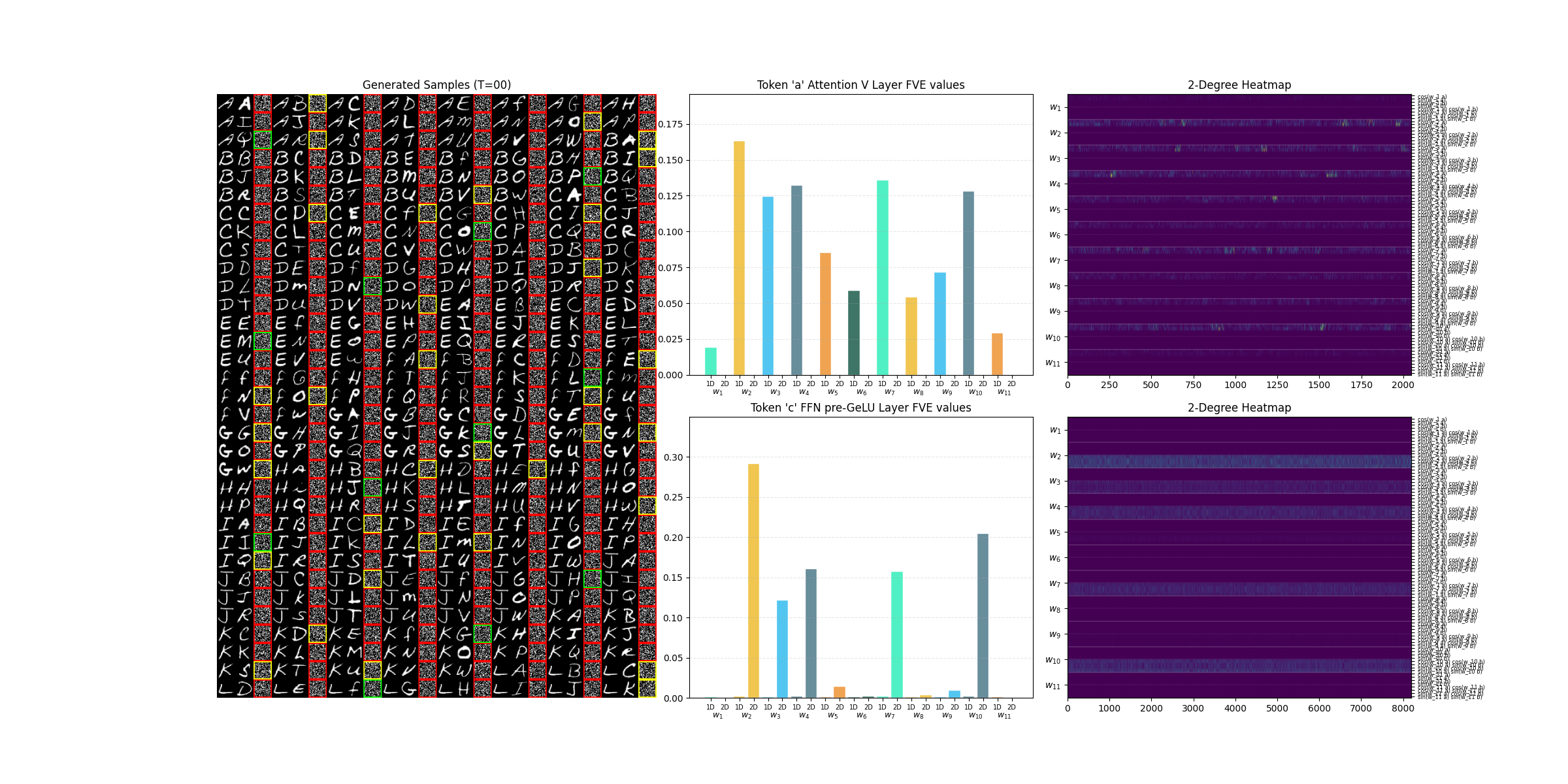}
    \caption{
    Timestep 0 with accuracy 0\%
    }
    \label{fig:appendix_fourier_timestep_000}
\end{figure}

As discussed in Section~\ref{subsec:multi_step}, the low entropy observed in the output FFN pre-GeLU activation signifies the model's ``arithmetic focusing mode.'' The visualizations below demonstrate this separation: while the sampled images at early timesteps remain perceptually blurry, the FFN activation's FVEs are already sharply concentrated on the five significant frequencies. This decoupling shows that the structural resolution of the arithmetic task precedes the visual refinement of the output modality.

\begin{figure}[H]
\centering
    \includegraphics[width=0.9\linewidth]{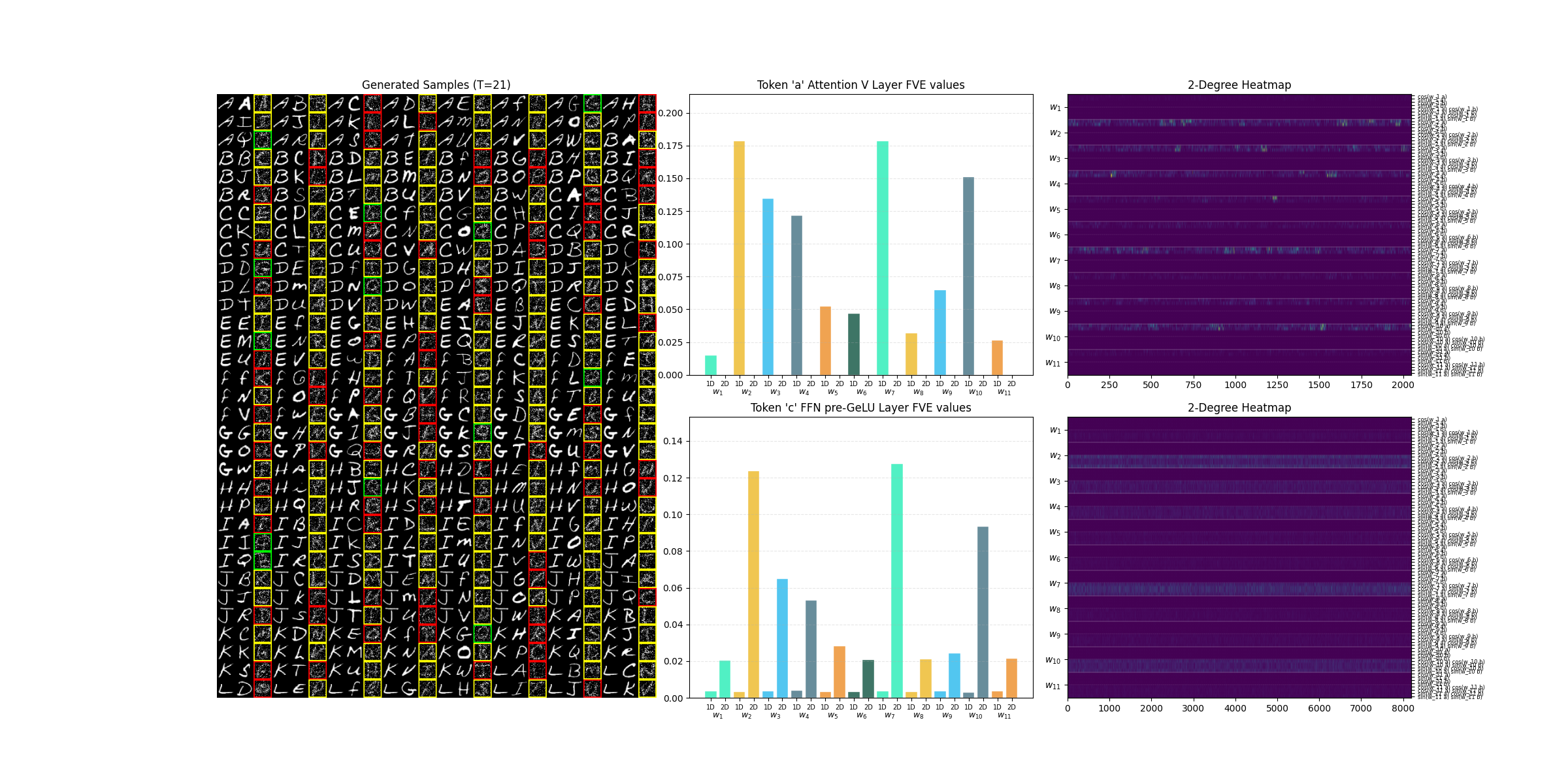}
    \caption{
    Timestep 21 with accuracy 0\%
    }
    \label{fig:appendix_fourier_timestep_021}
\end{figure}

As the generated digit becomes perceptually discernable in the synthesized image, the sharp FVE concentration begins to dissipate, as evidenced by the increasing entropy in the internal layers. This transition, illustrated in the visualizations below, signifies the model's shift from an abstract arithmetic operation to the spatial reconstruction of the visual output.

\begin{figure}[H]
\centering
    \includegraphics[width=0.9\linewidth]{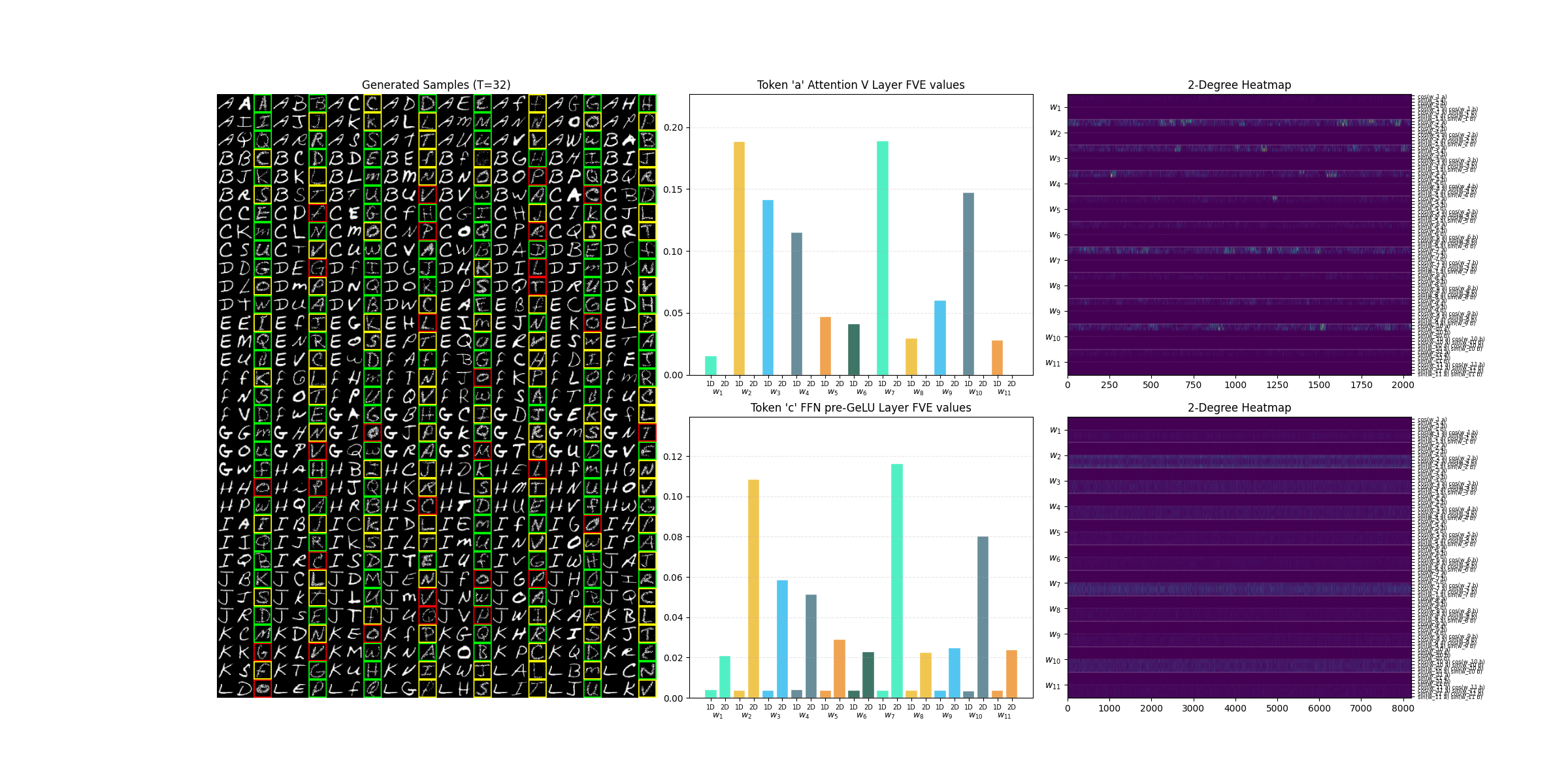}
    \caption{
    Timestep 32 with accuracy 54\%
    }
    \label{fig:appendix_fourier_timestep_032}
\end{figure}

\begin{figure}[H]
\centering
    \includegraphics[width=0.9\linewidth]{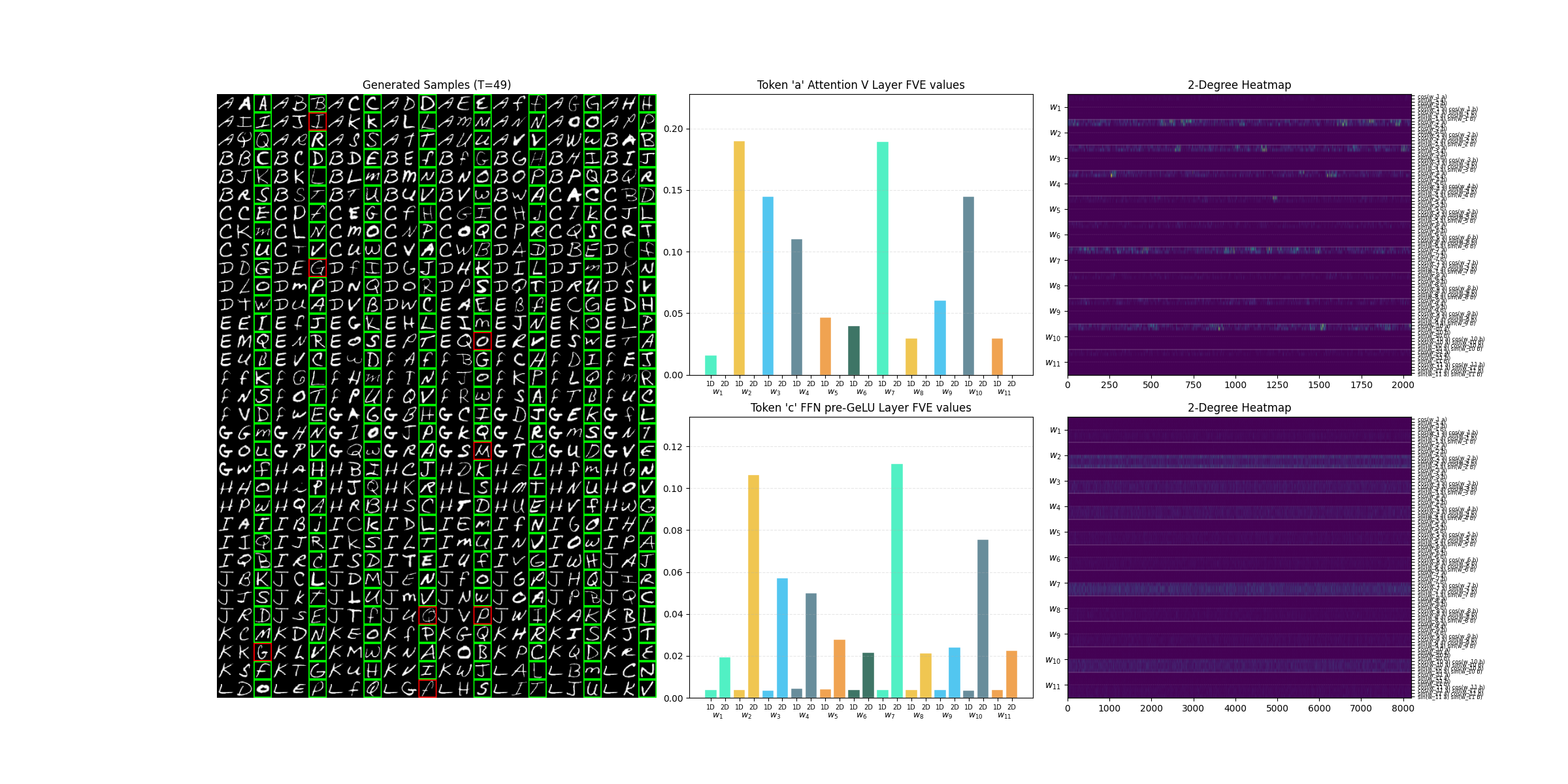}
    \caption{
    Timestep 50 with accuracy 96\%
    }
    \label{fig:appendix_fourier_timestep_049}
\end{figure}

\section{Ablation Study 1: Various Moduli $P$}
\label{appendix:ablations_p}

\begin{figure}[H]
\centering
    \includegraphics[width=0.82\linewidth]{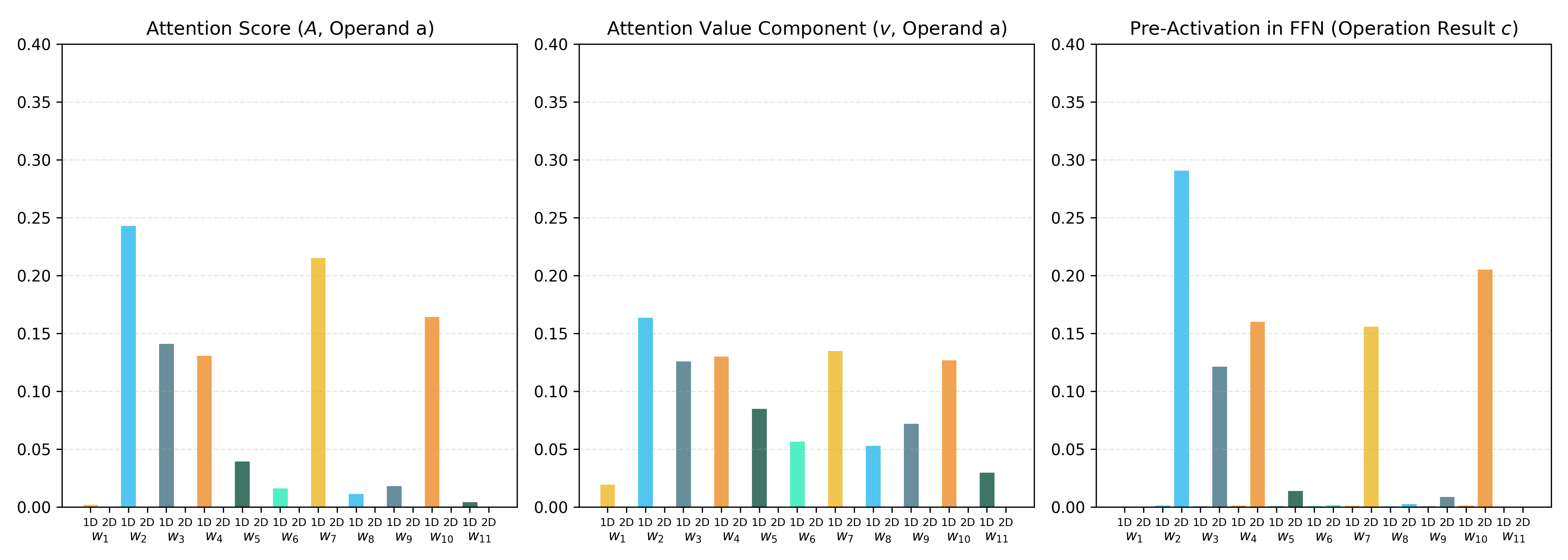}
    \caption{
    $P=23$. 
    }
    \label{fig:app_three_fve_p_23_emnist}
\end{figure}

The observed periodic structures persist beyond the baseline modulus. In ablations over several values of $P$, we find the same qualitative 1D and 2D periodicities in the $P=27$, $31$, and $35$ cases. Although the EMNIST dataset provides both uppercase and lowercase alphabets, we observed that the number of lowercase images is relatively small and that lowercase letters are sometimes mislabeled as uppercase. Thus, we use a mixed set of digit images and uppercase alphabet images to provide enough data for cases where $P>26$. The number of images per operation result class is fixed as $N=256$ in accordance with Section~\ref{subsec:multi_step}. Figures~\ref{fig:app_three_fve_p_23_emnist}$\sim$~\ref{fig:app_three_fve_p_35_emnist} show the 1D and 2D FVE values at three layers: Attention Score at $a$, Attention Value Component at $a$, and Pre-Activation in FFN at $c$. Sample steps are fixed at $t=0$, where the entropies of the FVEs remain low (i.e., high concentration on few frequencies).

\begin{figure}[H]
\centering
    \includegraphics[width=0.82\linewidth]{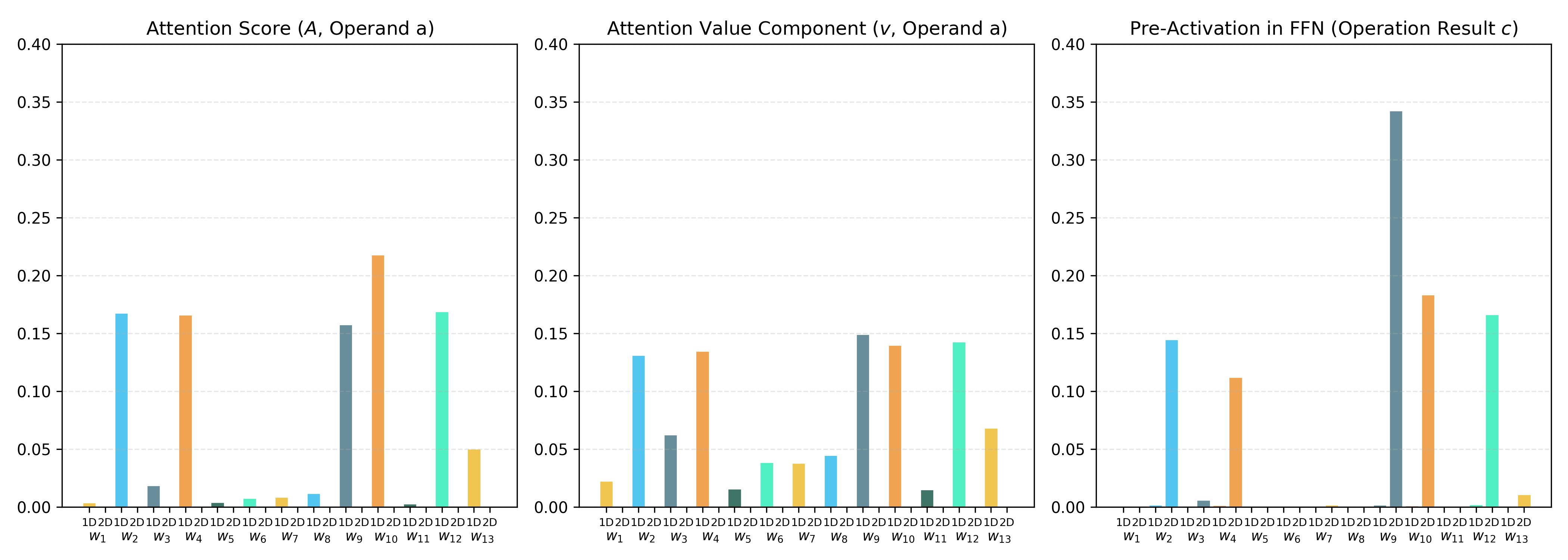}
    \caption{
    $P=27$
    }
    \label{fig:app_three_fve_p_27_emnist}
\end{figure}

\begin{figure}[H]
\centering
    \includegraphics[width=0.82\linewidth]{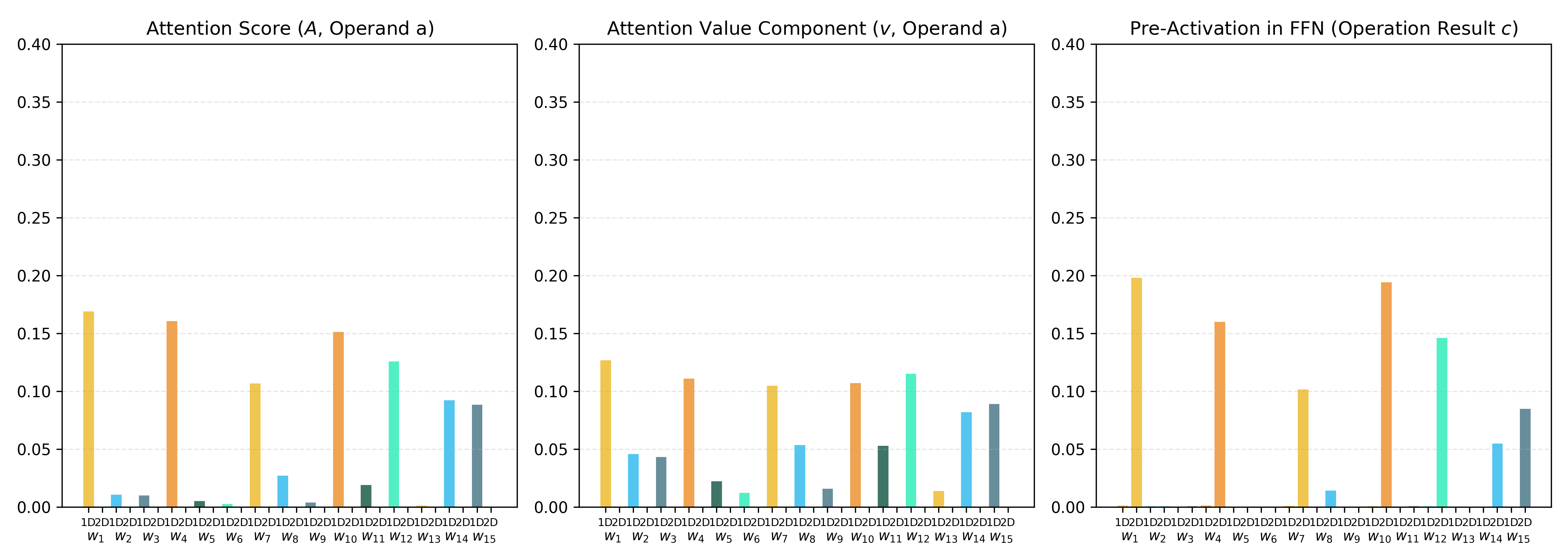}
    \caption{
    $P=31$
    }
    \label{fig:app_three_fve_p_31_emnist}
\end{figure}

\begin{figure}[H]
\centering
    \includegraphics[width=0.82\linewidth]{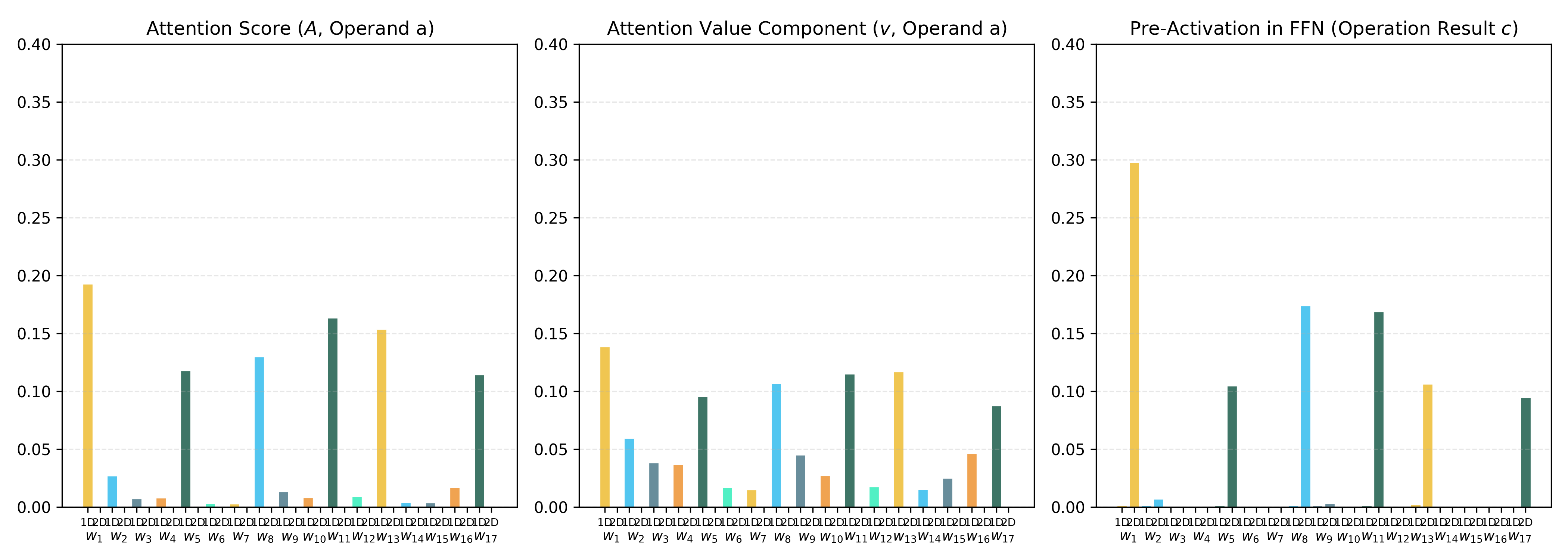}
    \caption{
    $P=35$
    }
    \label{fig:app_three_fve_p_35_emnist}
\end{figure}

\begin{figure}[H]
    \centering
    \begin{subfigure}[b]{0.45\textwidth}
        \centering
        \includegraphics[width=\linewidth]{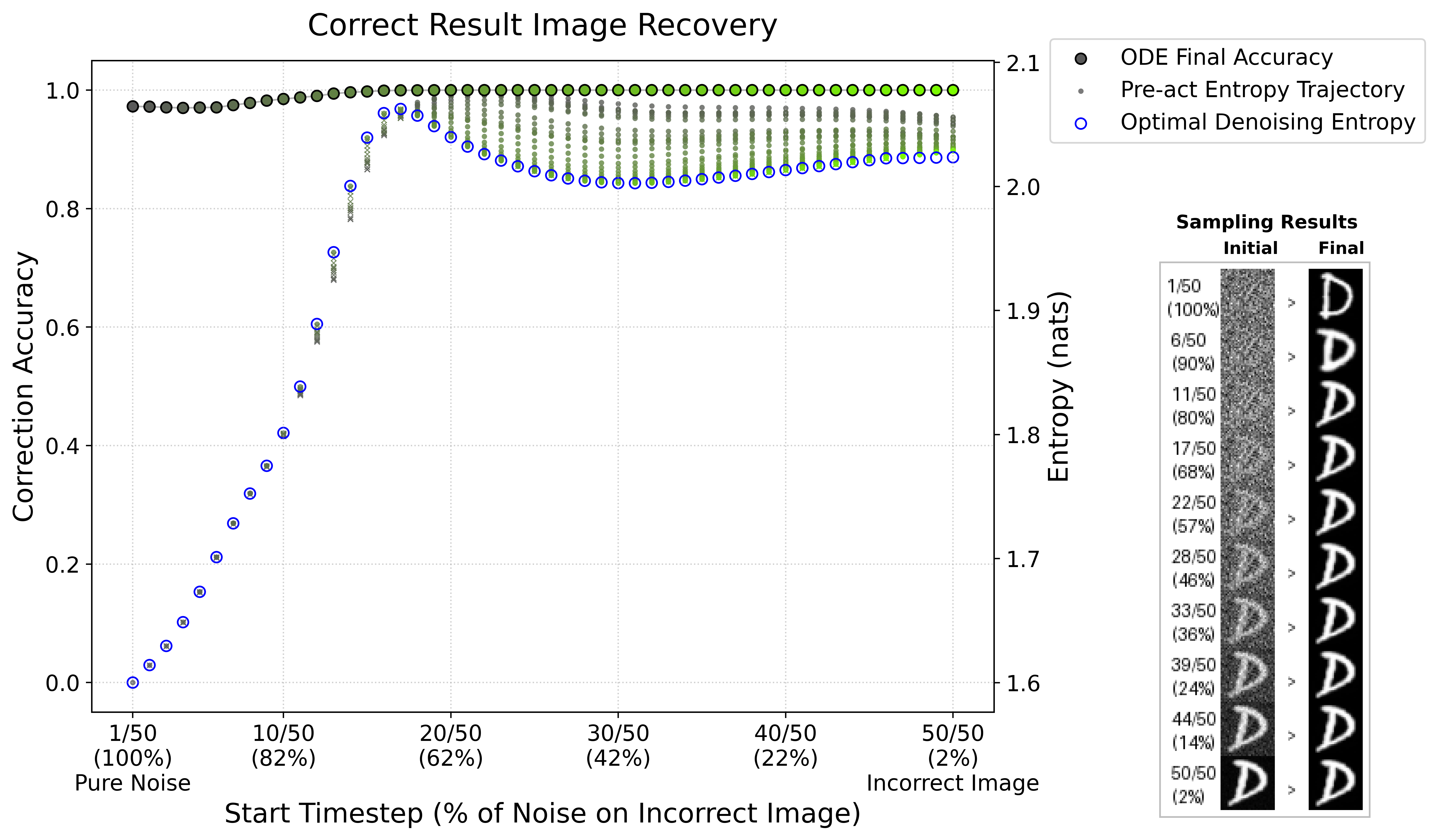}
        \caption{Correct result image recovery}
    \end{subfigure}
    \hfill 
    \begin{subfigure}[b]{0.45\textwidth}
        \centering
        \includegraphics[width=\linewidth]{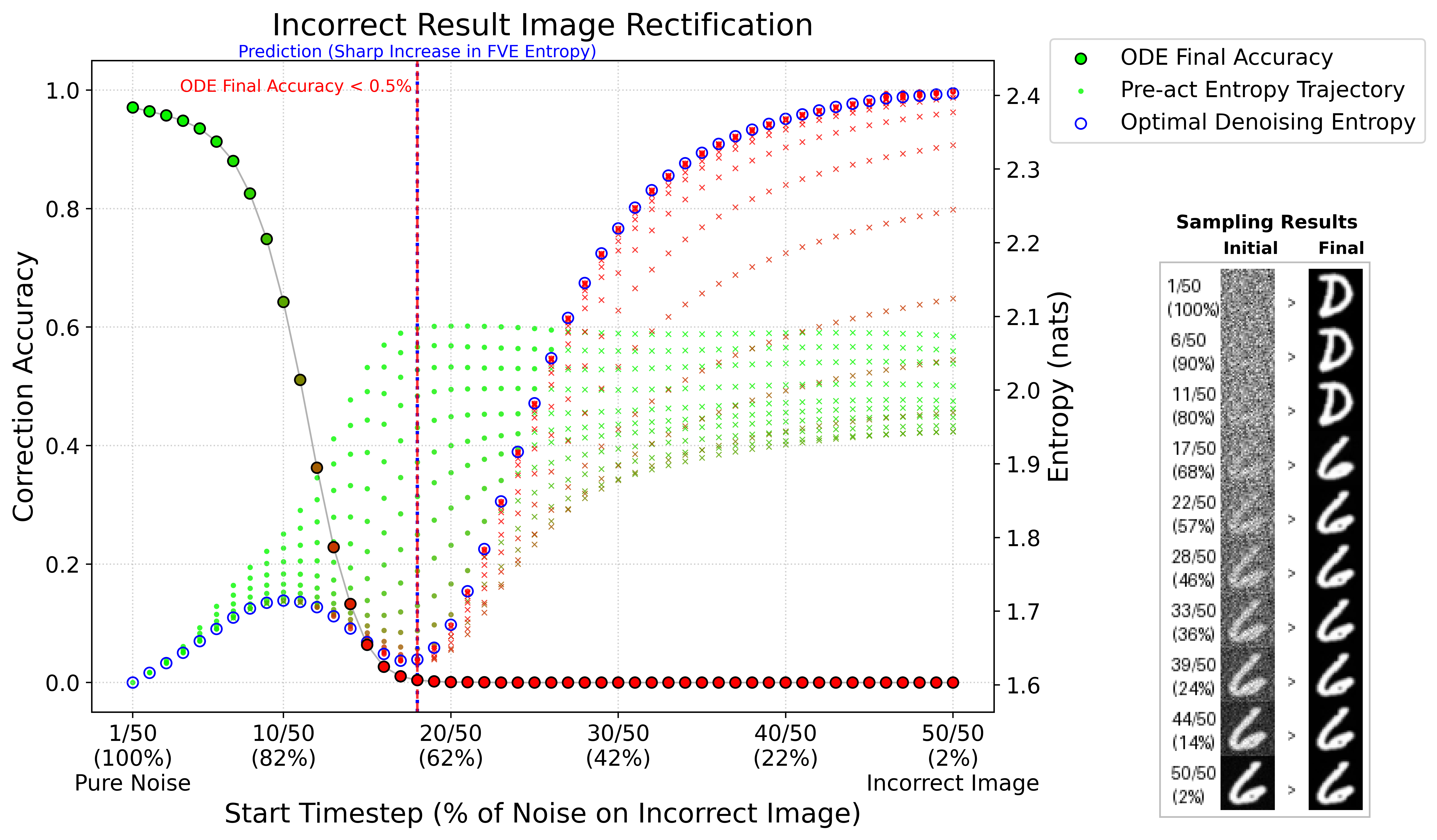}
        \caption{Incorrect result image rectification}
    \end{subfigure}
    
\caption{
    $P=27$
}
    \label{fig:app_mode_shift_p_27_emnist}
\end{figure}

\begin{figure}[H]
    \centering
    \begin{subfigure}[b]{0.45\textwidth}
        \centering
        \includegraphics[width=\linewidth]{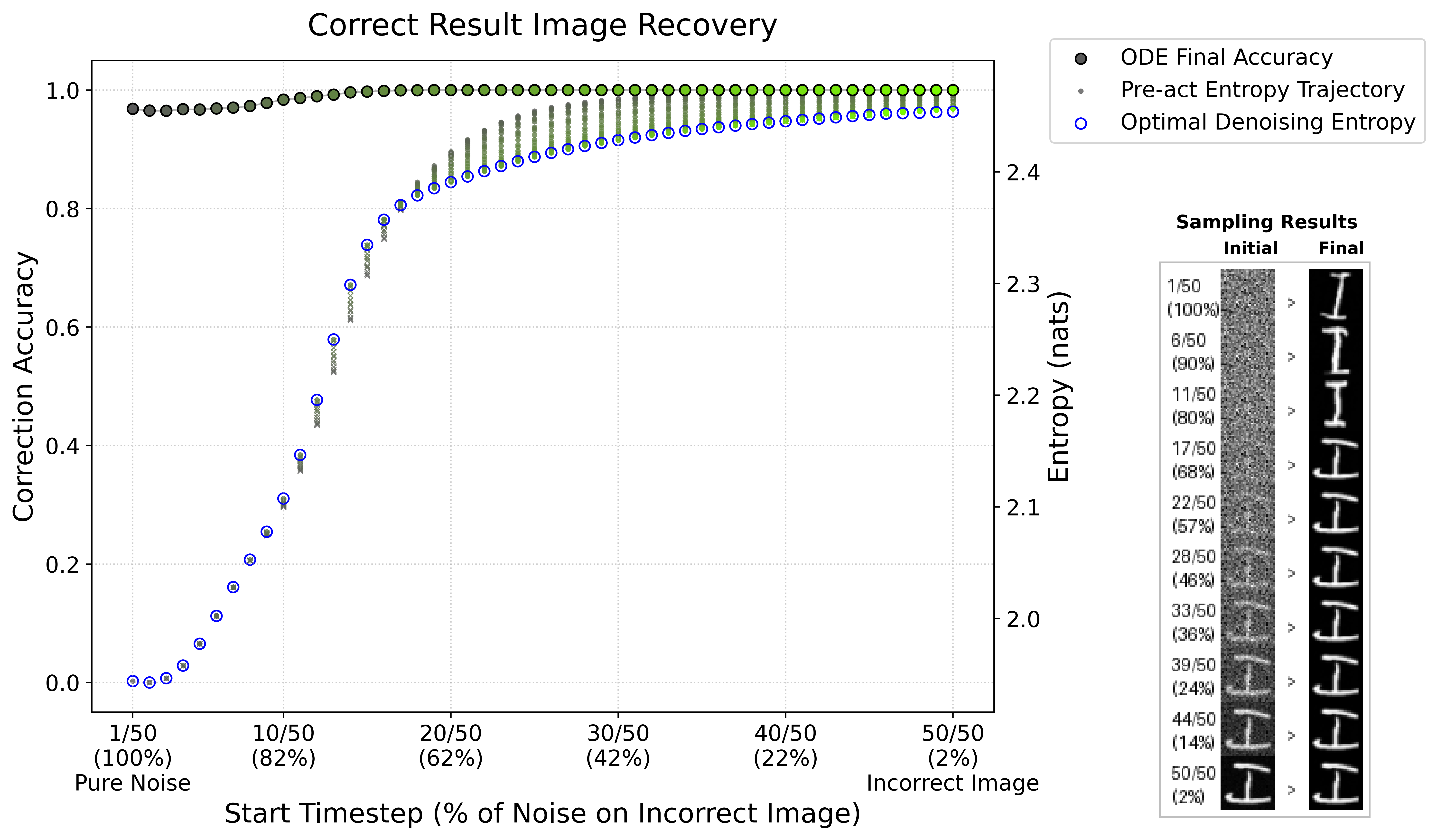}
        \caption{Correct result image recovery}
    \end{subfigure}
    \hfill 
    \begin{subfigure}[b]{0.45\textwidth}
        \centering
        \includegraphics[width=\linewidth]{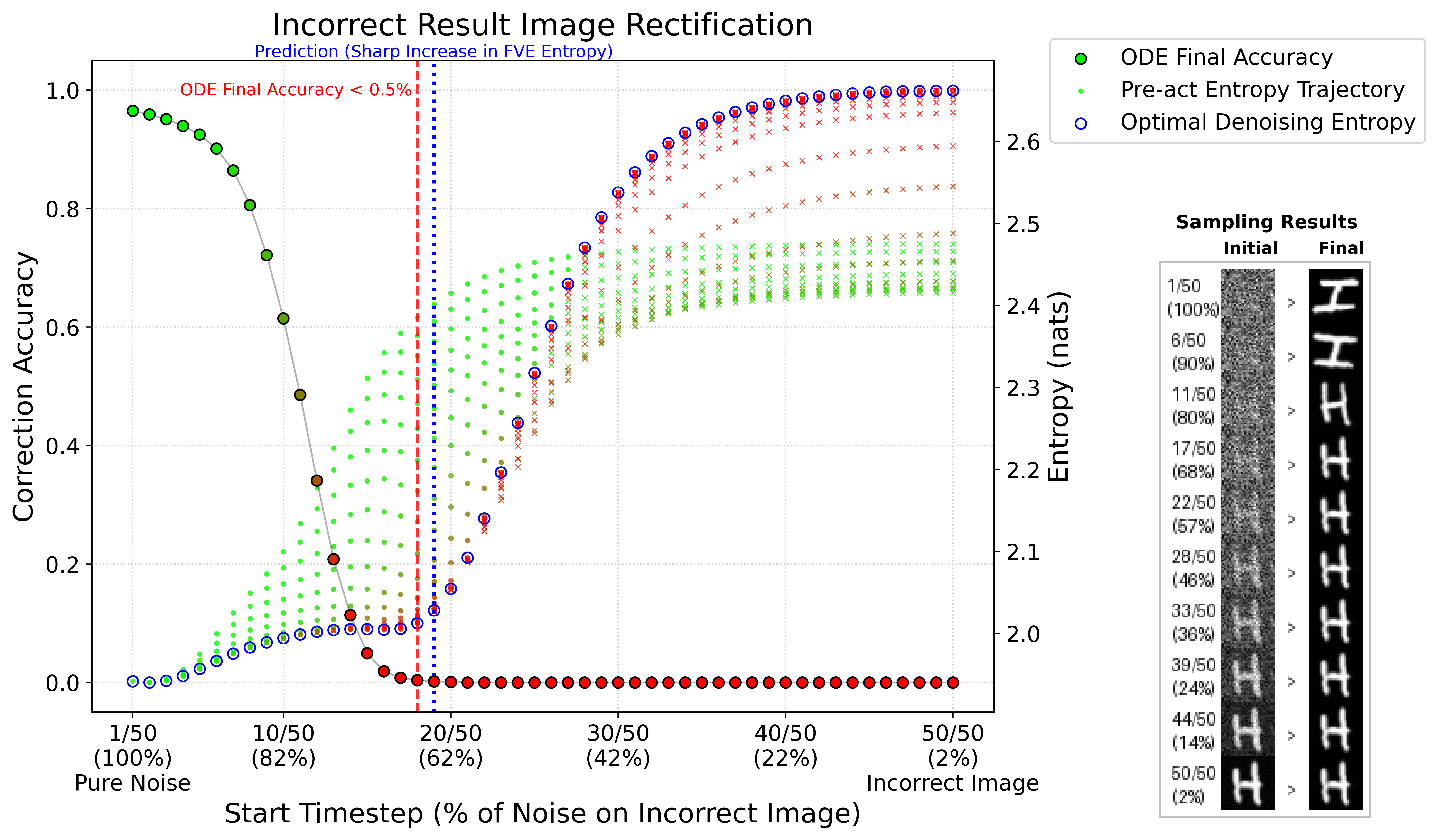}
        \caption{Incorrect result image rectification}
    \end{subfigure}
    
\caption{
    $P=31$
}
    \label{fig:app_mode_shift_p_31_emnist}
\end{figure}

\begin{figure}[H]
    \centering
    \begin{subfigure}[b]{0.49\textwidth}
        \centering
        \includegraphics[width=\linewidth]{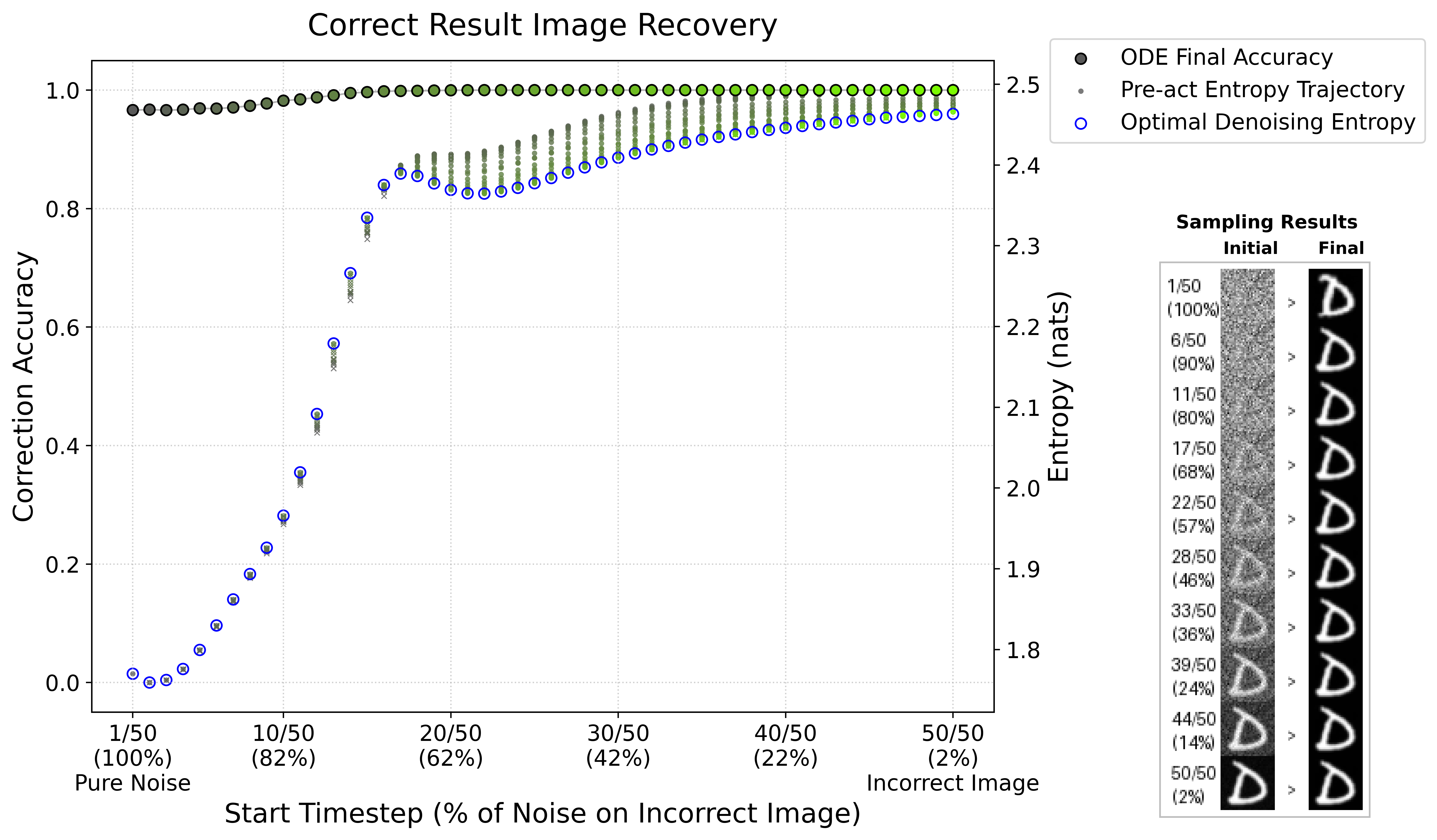}
        \caption{Correct result image recovery}
    \end{subfigure}
    \hfill 
    \begin{subfigure}[b]{0.49\textwidth}
        \centering
        \includegraphics[width=\linewidth]{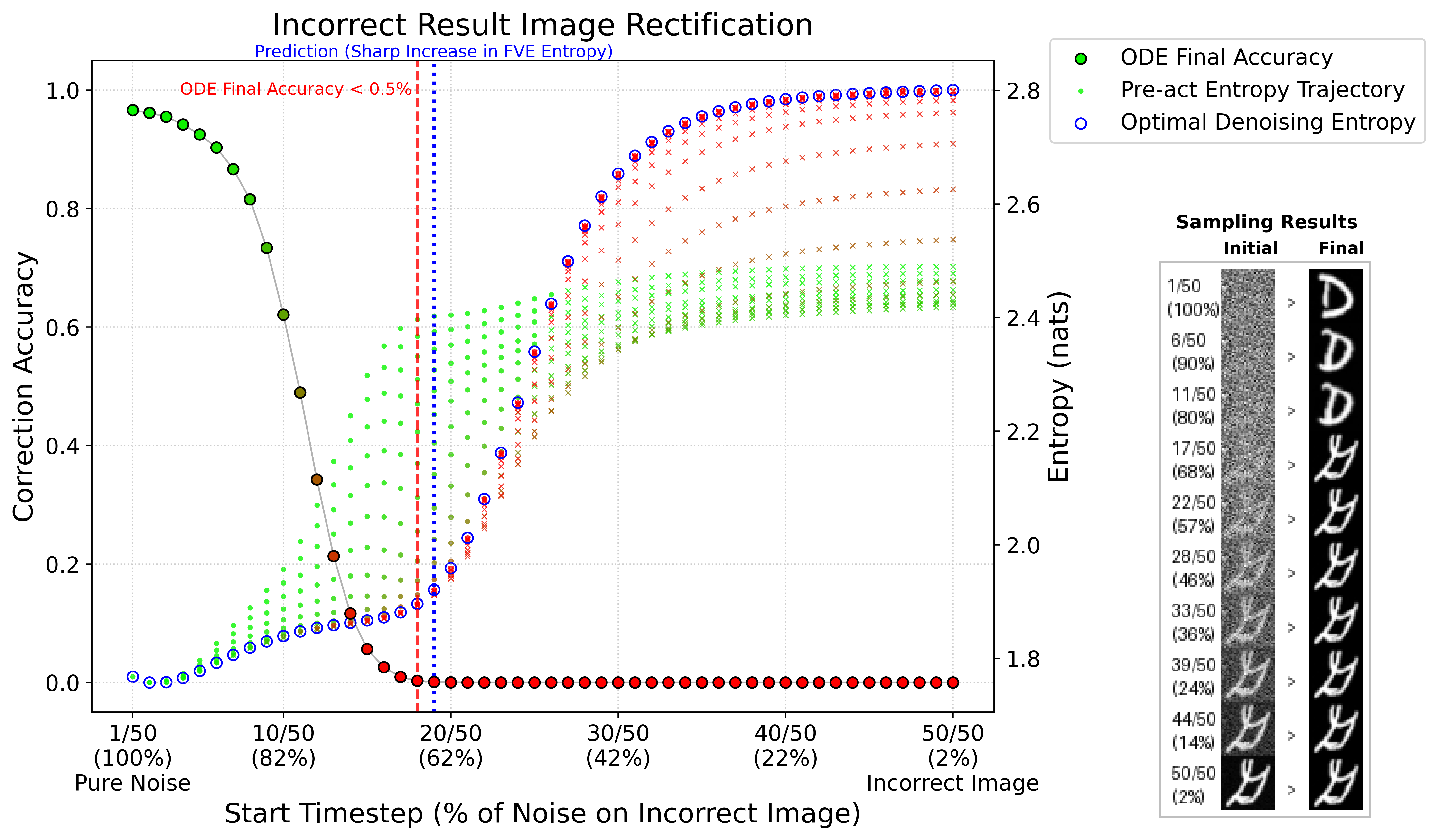}
        \caption{Incorrect result image rectification}
    \end{subfigure}
    
\caption{
    $P=35$
}
    \label{fig:app_mode_shift_p_35_emnist}
\end{figure}

Figures~\ref{fig:app_mode_shift_p_27_emnist}--\ref{fig:app_mode_shift_p_35_emnist} show that the mode-shift pattern during ODE sampling recurs across additional $P$ values. As in the baseline $P=23$ case, the models exhibit the same trajectory structure for both recovery from correct answers and rectification from initially incorrect images.

\section{Ablation Study 2: Kuzushiji-MNIST dataset}
\label{appendix:ablations_kuzushiji}

We further show that the emergence of periodic structures is not specific to EMNIST by applying the same Fourier analysis to models trained on the Kuzushiji-MNIST dataset \citep{DBLP:journals/corr/abs-1812-01718}. Because handwritten Japanese characters contain relatively more complex shapes than the English alphabet, these experiments demonstrate that the FFN-sandwich structure can form class-aligned periodic representations in a more visually heterogeneous setting. Using the richer class set provided by Kuzushiji-MNIST, we observe the same qualitative spectral and transition patterns for $P=39, 43, \text{ and } 47$, as shown below.

\begin{figure}[H]
\centering
    \includegraphics[width=0.85\linewidth]{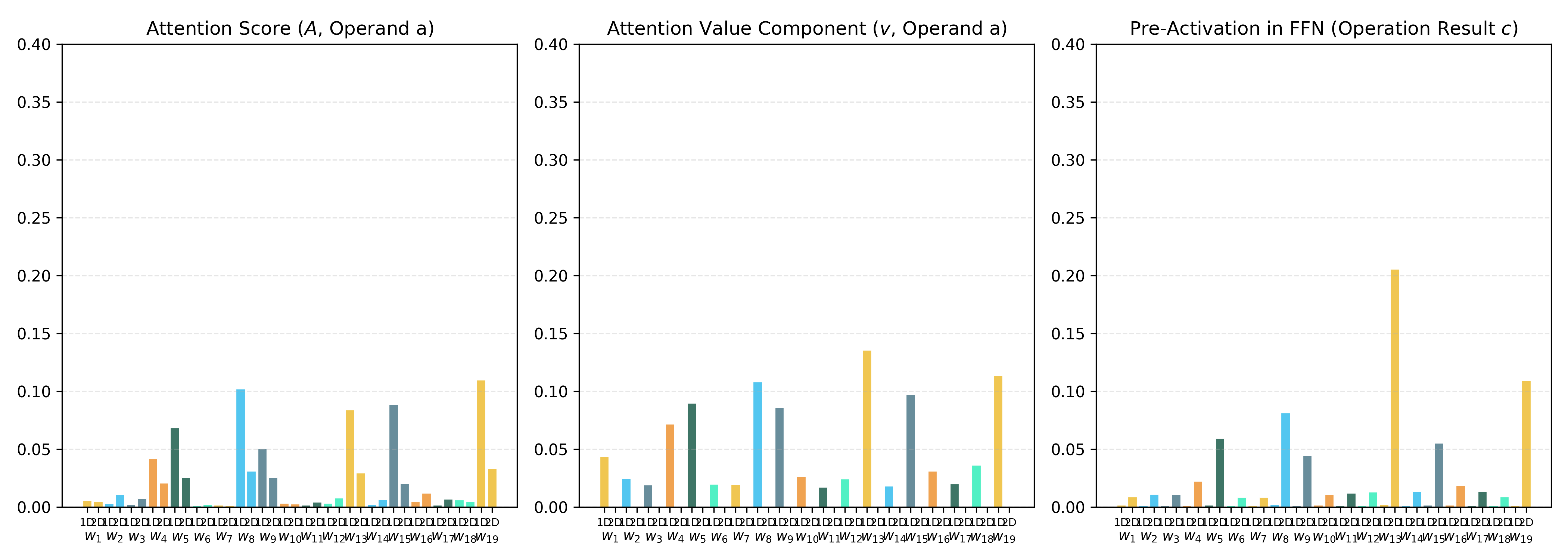}
    \caption{
    $P=39$ with Kuzushiji-MNIST dataset
    }
    \label{fig:app_three_fve_p_39_kmnist}
\end{figure}

\begin{figure}[H]
\centering
    \includegraphics[width=0.85\linewidth]{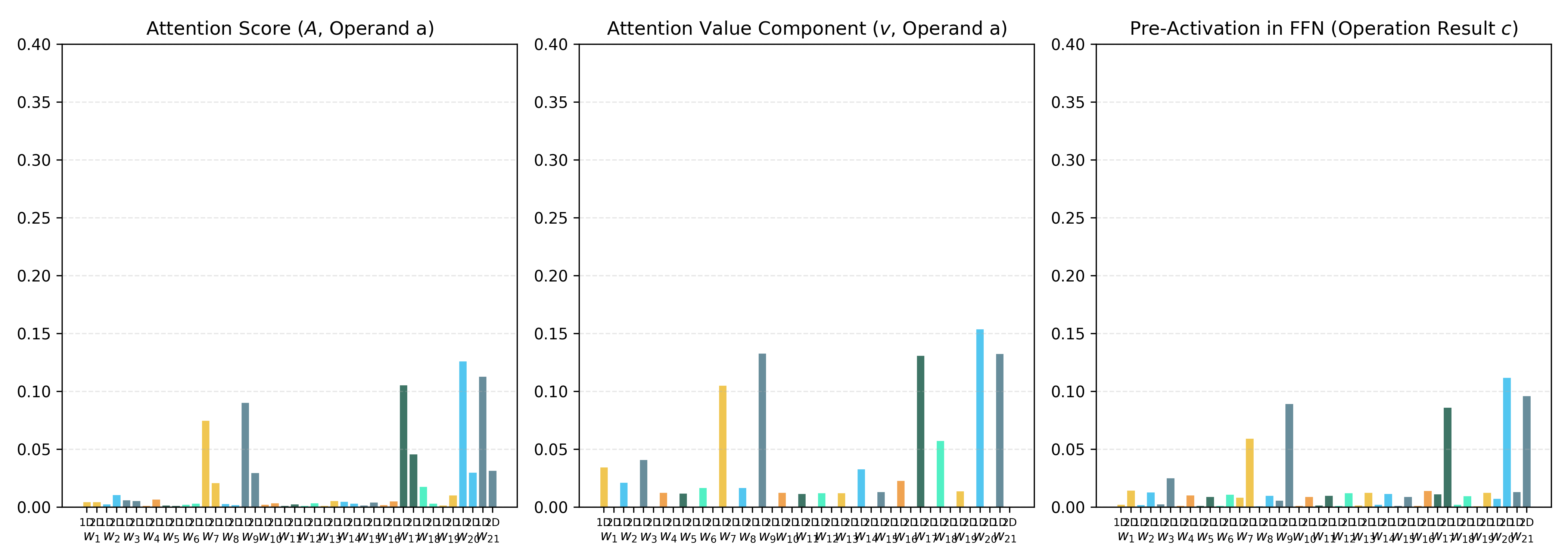}
    \caption{
    $P=43$ with Kuzushiji-MNIST dataset
    }
    \label{fig:app_three_fve_p_43_kmnist}
\end{figure}

\begin{figure}[H]
\centering
    \includegraphics[width=0.85\linewidth]{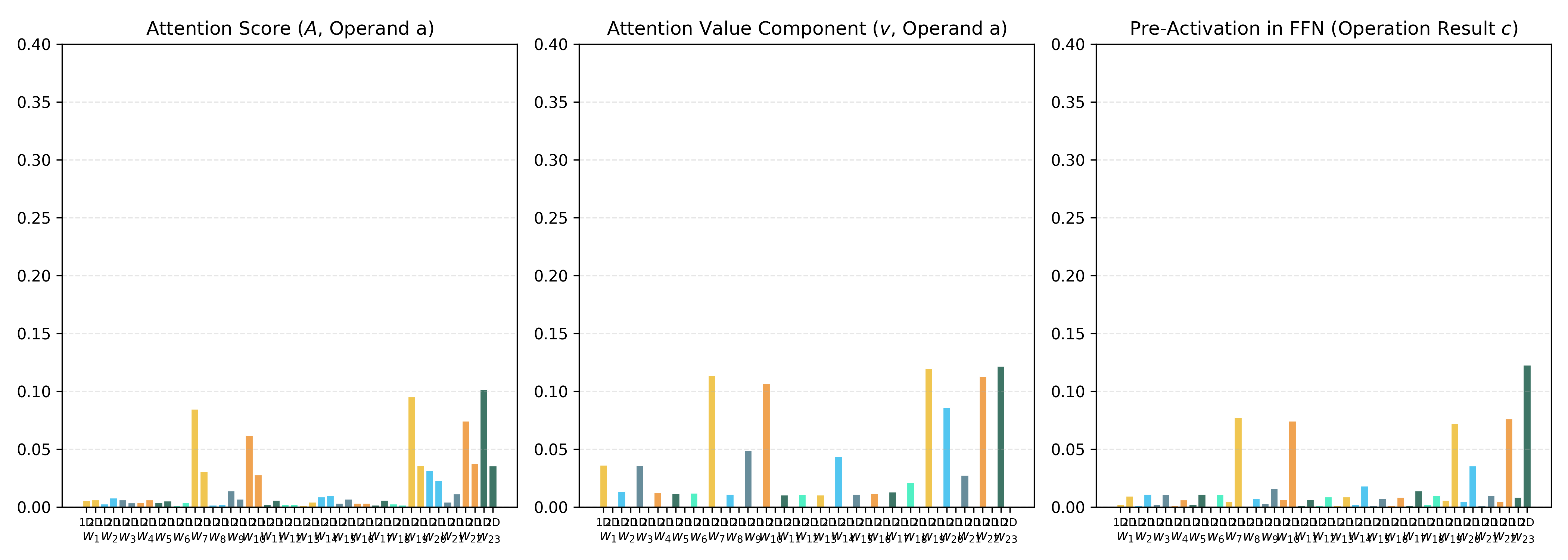}
    \caption{
    $P=47$ with Kuzushiji-MNIST dataset
    }
    \label{fig:app_three_fve_p_47_kmnist}
\end{figure}

\begin{figure}[H]
    \centering
    \begin{subfigure}[b]{0.49\textwidth}
        \centering
        \includegraphics[width=\linewidth]{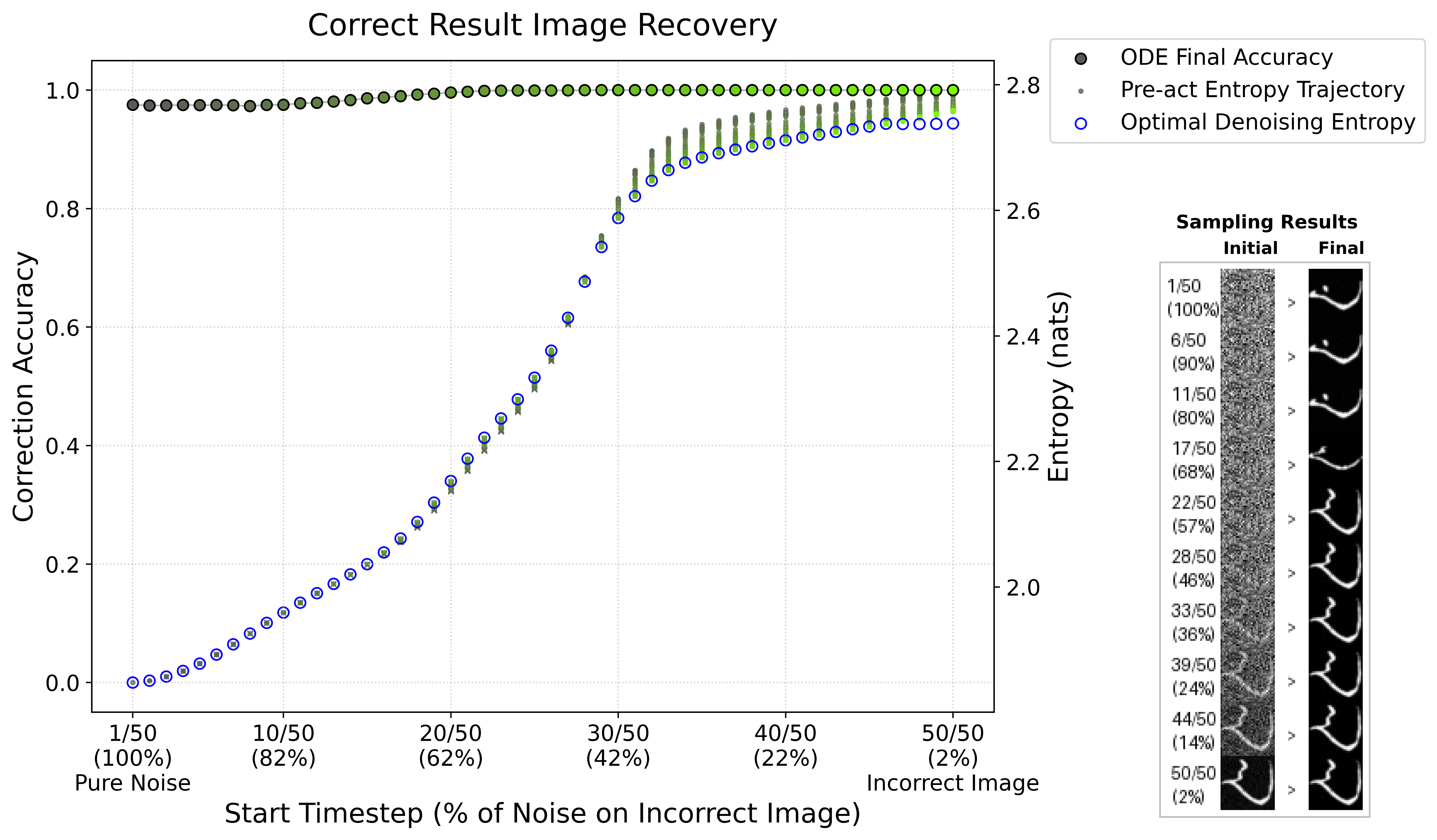}
        \caption{Correct result image recovery}
    \end{subfigure}
    \hfill 
    \begin{subfigure}[b]{0.49\textwidth}
        \centering
        \includegraphics[width=\linewidth]{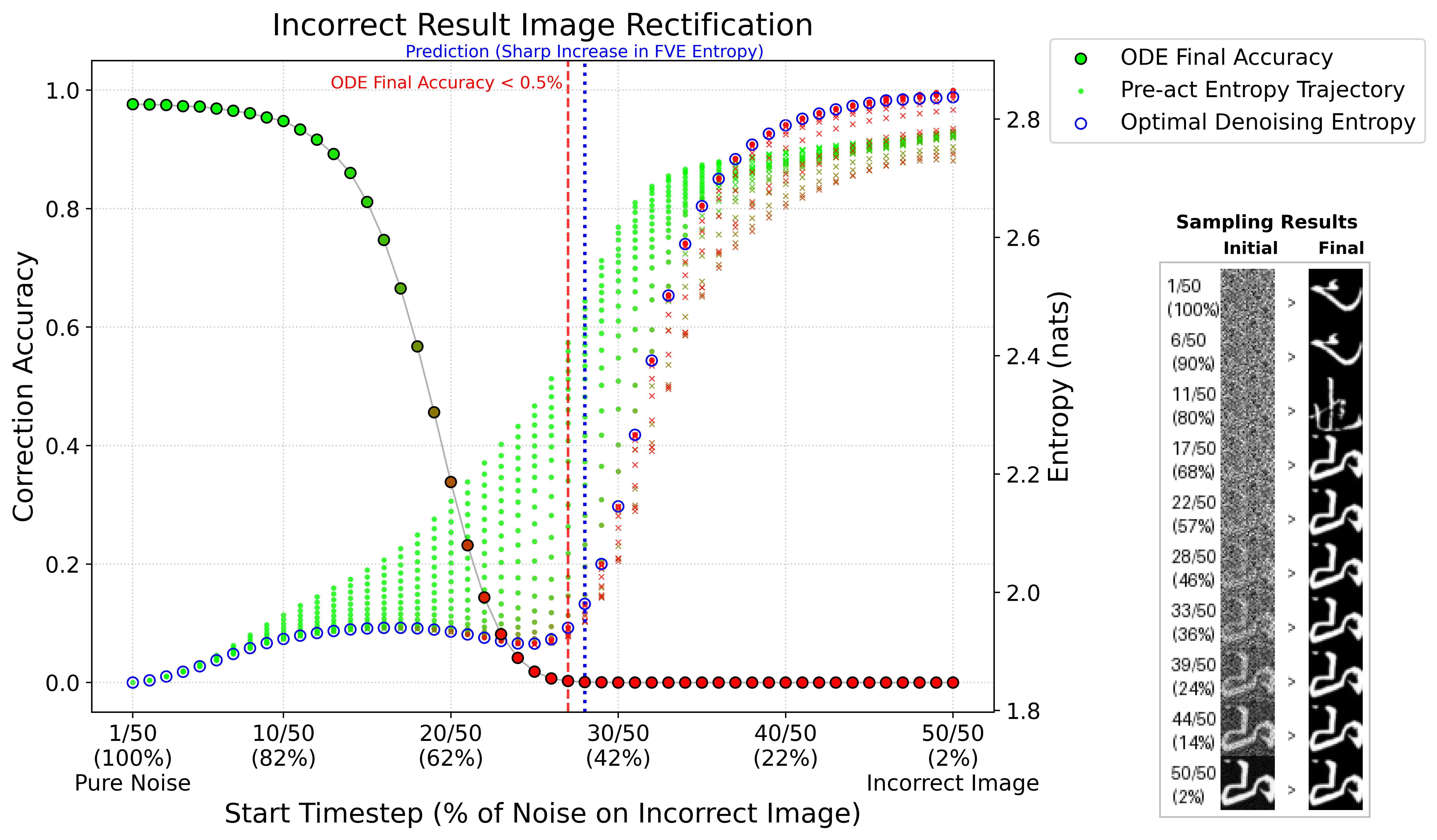}
        \caption{Incorrect result image rectification}
    \end{subfigure}
    
\caption{
    $P=39$ on Kuzushiji-MNIST dataset.
}
    \label{fig:app_mode_shift_p_39_kmnist}
\end{figure}

Figures~\ref{fig:app_mode_shift_p_39_kmnist}$\sim$~\ref{fig:app_mode_shift_p_47_kmnist} show that the transition between arithmetic-dominated and visual-refinement-dominated behavior also recurs along the ODE sampling path for models trained on Kuzushiji-MNIST.

\begin{figure}[H]
    \centering
    \begin{subfigure}[b]{0.49\textwidth}
        \centering
        \includegraphics[width=\linewidth]{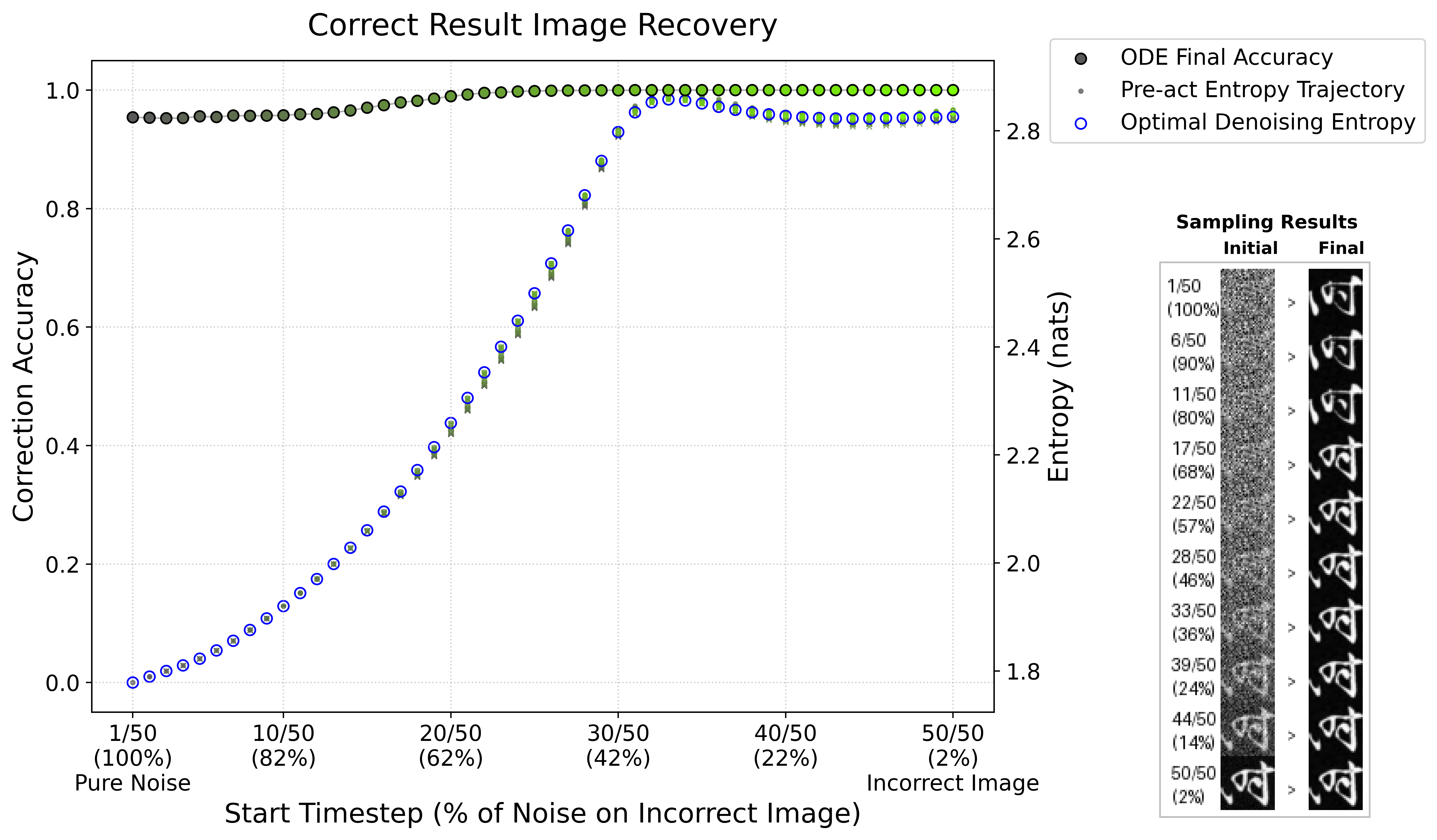}
        \caption{Correct result image recovery}
    \end{subfigure}
    \hfill 
    \begin{subfigure}[b]{0.49\textwidth}
        \centering
        \includegraphics[width=\linewidth]{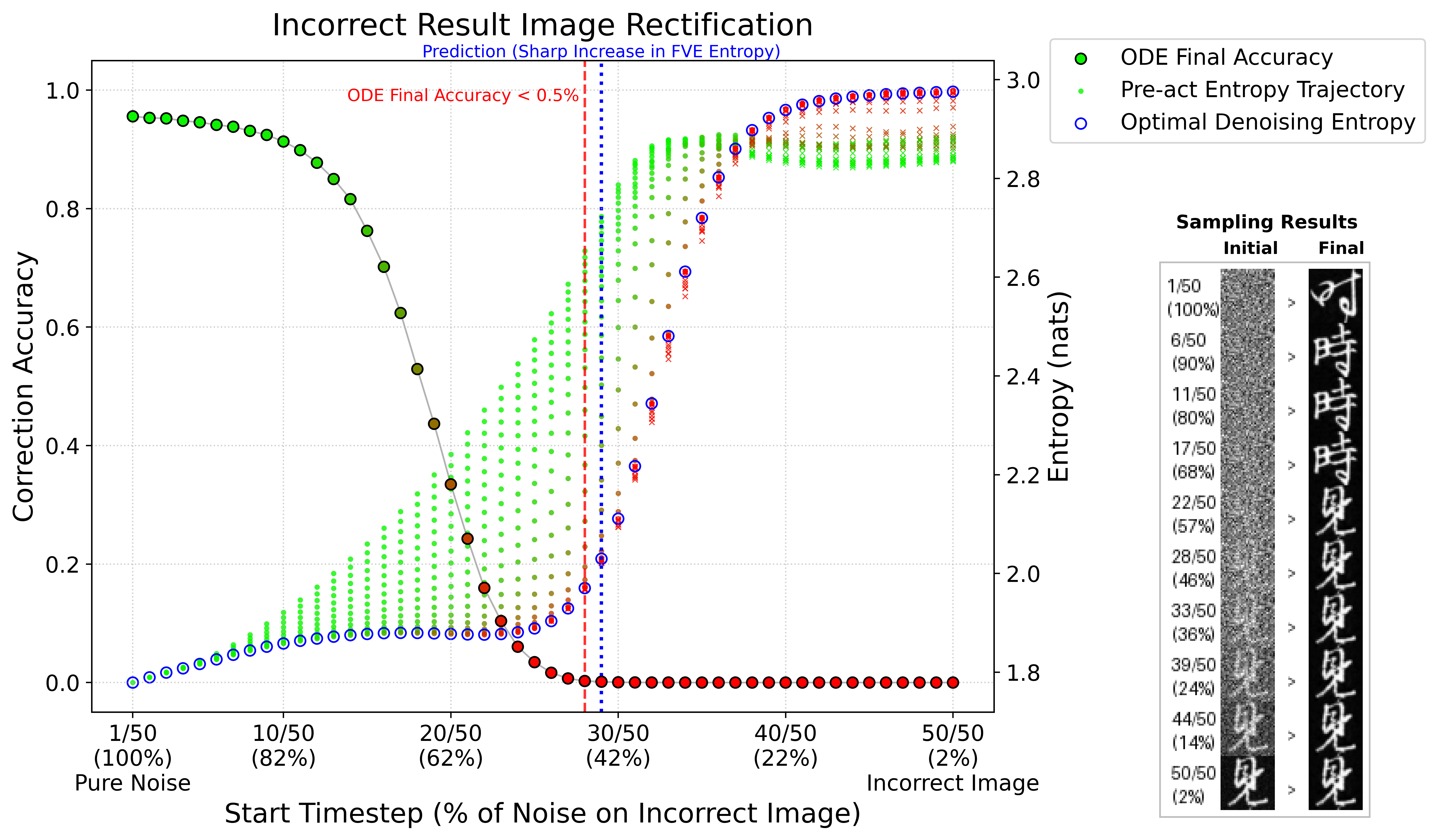}
        \caption{Incorrect result image rectification}
    \end{subfigure}
    
\caption{
    $P=43$ on Kuzushiji-MNIST dataset.
}
    \label{fig:app_mode_shift_p_43_kmnist}
\end{figure}

\begin{figure}[H]
    \centering
    \begin{subfigure}[b]{0.45\textwidth}
        \centering
        \includegraphics[width=\linewidth]{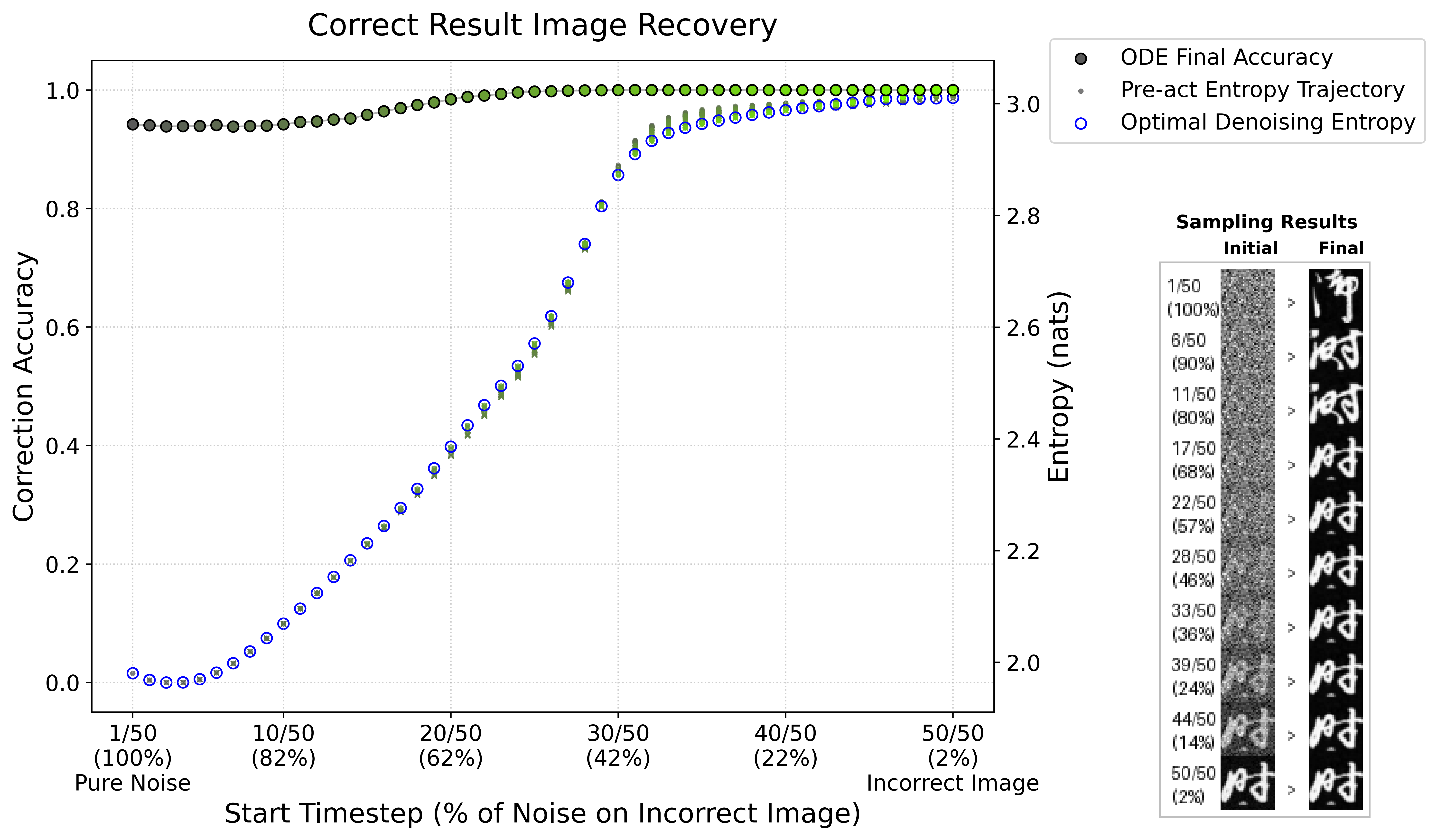}
        \caption{Correct result image recovery}
    \end{subfigure}
    \hfill 
    \begin{subfigure}[b]{0.45\textwidth}
        \centering
        \includegraphics[width=\linewidth]{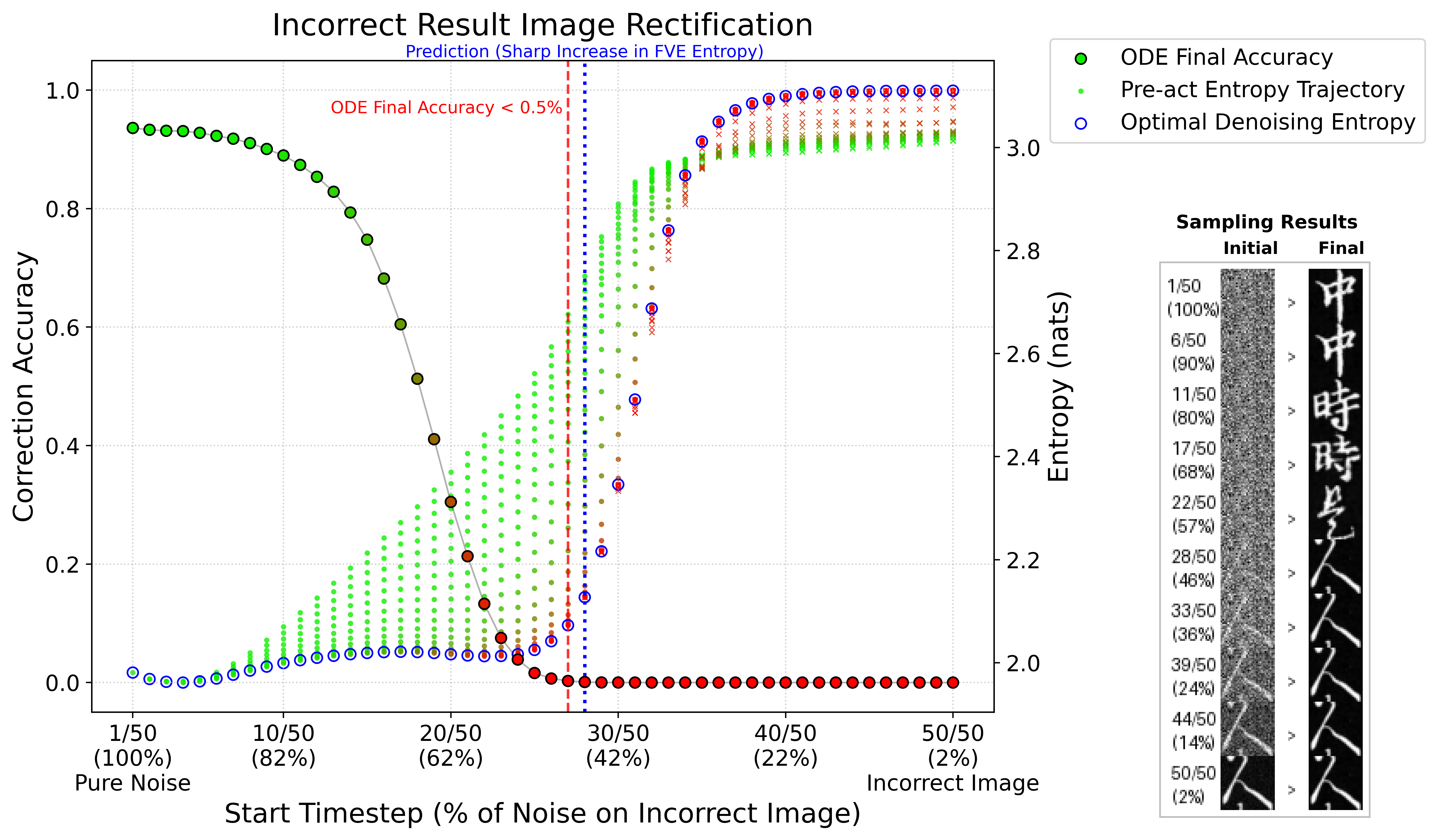}
        \caption{Incorrect result image rectification}
    \end{subfigure}
    
\caption{
    $P=47$ on Kuzushiji-MNIST dataset.
}
    \label{fig:app_mode_shift_p_47_kmnist}
\end{figure}

\section{Ablation Study 4: Phase Transition Across Various $P$ Values and Datasets}
\label{appendix:phase_transition_ablation}

Across the tested $P$ values and datasets, a sudden increase in FVE entropy---reflecting reduced concentration on selective frequencies---closely coincides with the timestep at which the final ODE sampling accuracy drops to near zero ($< 0.5\%$). This critical timestep, $t^{\ast}$, marks the point where the model shifts from an arithmetic-dominated regime to a visual-refinement-dominated regime. Operationally, we estimate $t^{\ast}$ by finding the maximum of the second derivative of the FVE: $t^{\ast} = \operatorname{argmax}_t \frac{\partial^2 \mathrm{FVE}}{\partial t^2}$.

\begin{figure}[!htbp]
\centering
    \includegraphics[width=0.70\linewidth]{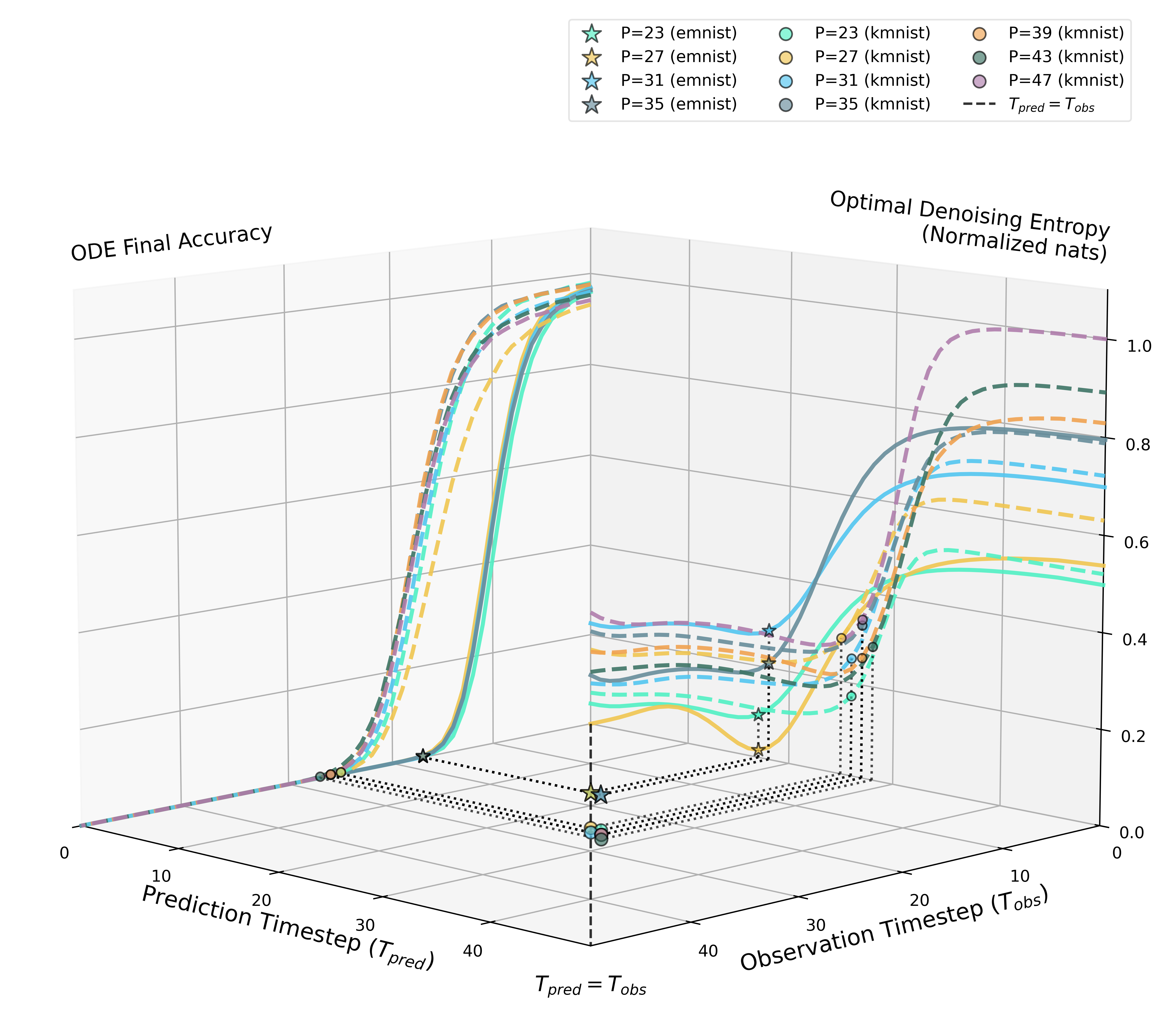}
    \caption{
    \textbf{Alignment of Predicted and Observed Timesteps for the Phase Transition.}
    This 3D visualization illustrates the incorrect image rectification dynamics detailed in Section~\ref{subsec:multi_step}. The left vertical plane displays ODE final accuracies across varying initial noise levels for models trained on different $P$ values and datasets (EMNIST and KMNIST). The marked points denote the $0.5\%$ accuracy threshold, where models largely fail to rectify incorrect inputs. The right vertical plane plots the Fourier Variance Explained (FVE) entropy along the optimal denoising trajectory, with markers indicating the sharp onset of entropy escalation ($\arg\max_t \frac{\partial^2 \text{FVE}}{(\partial t)^2}$). The bottom plane projects these two critical timesteps, revealing a strong correlation along the $T_{\text{pred}} = T_{\text{obs}}$ diagonal. This alignment establishes FVE entropy as an internal predictor of the transition from arithmetic-dominated behavior to visual-refinement-dominated behavior in our experiments. Refer to Appendix~\ref{appendix:ablations_p} for dataset ablations.
    }
    \label{fig:phase_transition_alignment}
\end{figure}

\section{Ablation Study 5: Grokking in Multi-layer Model}
\label{appendix:ablations_multi_layer}

Our architectural choice of a single-layer DiT in the main text was intentional; isolating a single self-attention block makes the periodic structure easier to analyze via Fourier methods and localizes where operand interactions occur. However, the grokking behavior itself is not restricted to the custom FFN-sandwich architecture in our experiments.

\begin{figure}[H]
    \centering
    \begin{subfigure}[b]{0.49\textwidth}
        \centering
        \includegraphics[width=\linewidth]{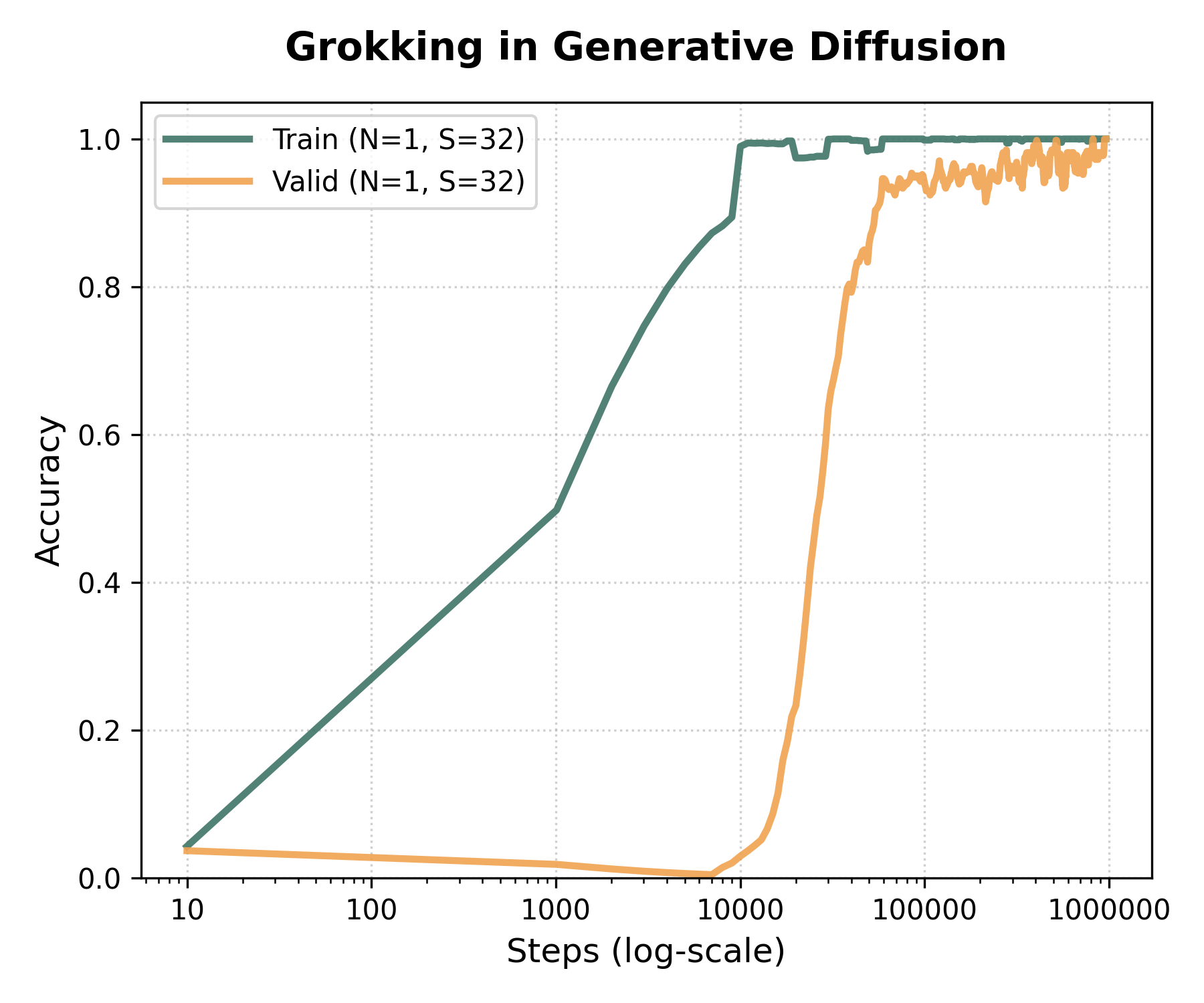}
        \caption{$N=1$ Single-Image-Regime Case}
        \label{fig:ablation_multi_layer_depth_2_left}
    \end{subfigure}
    \hfill 
    \begin{subfigure}[b]{0.49\textwidth}
        \centering
        \includegraphics[width=\linewidth]{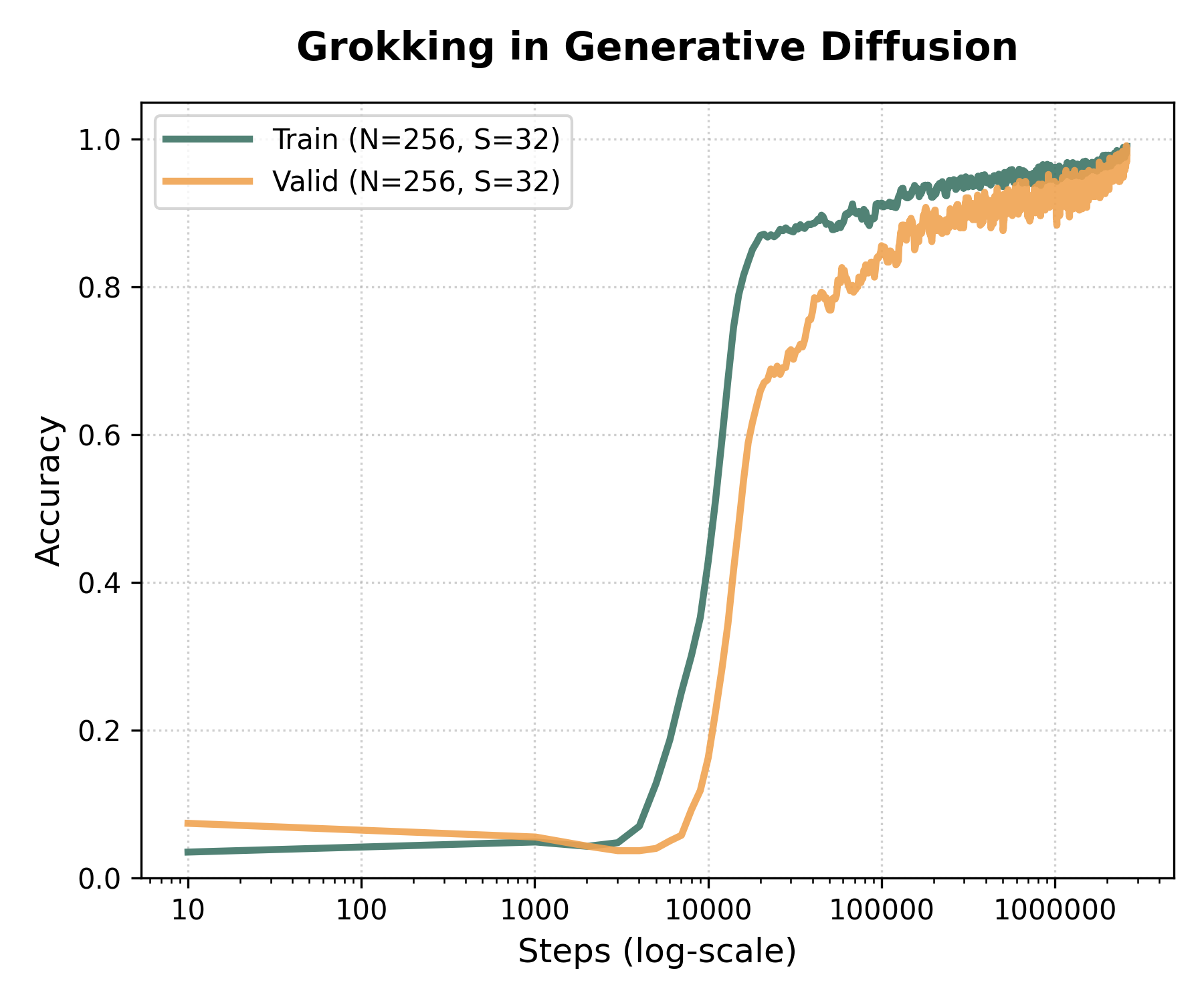}
        \caption{$N=256$ Diverse-Image-Regime Case}
        \label{fig:ablation_multi_layer_depth_2_right}
    \end{subfigure}
    
\caption{
    \textbf{Grokking phenomena observed on a standard depth-2 architecture.}  
    Validation accuracy trajectories show that successful generalization (grokking) occurs in both (a) the $N=1$ single-image regime and (b) the $N=256$ diverse-image regime, providing evidence that the phenomenon is not restricted to the single-layer FFN-sandwich model.
}
    \label{fig:ablation_multi_layer_depth_2}
\end{figure}

We confirm this with an ablation study using a standard multi-layer diffusion model with a depth of 2 (i.e., Embedding $\rightarrow$ SA $\rightarrow$ FFN $\rightarrow$ SA $\rightarrow$ FFN $\rightarrow$ Output). As shown in Figure~\ref{fig:ablation_multi_layer_depth_2}, grokking emerges in both the $N=1$ and the diverse-image $N=256$ regimes. This matches our class-separation account: in a depth-2 standard model, the first FFN layer precedes the second self-attention block and provides the non-linear capacity needed to form class-aligned representations, serving a role analogous to the pre-SA FFN in our sandwich architecture. The depth-2 model achieves near-perfect generalization in the $N=256$ case (Figure~\ref{fig:ablation_multi_layer_depth_2_right}), improving over the single-layer FFN-sandwich architecture, which plateaued at approximately 94\% accuracy (Figure~\ref{fig:grokking_train_result_n_256}).

\section{Phase Transition in a Standard DiT}
\label{appendix:standard_dit}

To examine whether the phase transition observed in our experiments depends on the preliminary feed-forward network (FFN) placed before self-attention, we evaluate a standard Diffusion Transformer (DiT) without this additional FFN on a two-dimensional modular addition task. We construct a dataset with $P=23$ classes represented by equidistant points on a circle of radius $R\in\{12,15,20\}$, with $N=4$ noisy samples per class. We follow the data-splitting strategy used in the main experiments and track the evolution of the internal representation using FVE entropy during rectification.

We estimate the transition point from the timestep at which the second derivative of the FVE entropy $H(t)$ reaches its maximum:
\begin{equation}
    \hat{t} = \operatorname*{arg\,max}_{t}\frac{d^2 H(t)}{dt^2}.
\end{equation}
Table~\ref{tab:standard_dit} compares this prediction with the timestep at which ODE-based rectification fails. FVE increases throughout the sampling trajectory in all three settings. The predicted and observed timesteps differ by one step for $R=12$ and $R=15$, whereas the discrepancy is larger for $R=20$. These results indicate that the observed transition is not exclusive to the FFN-sandwich architecture.

\begin{table}[H]
\centering
\caption{Phase-transition analysis with a standard DiT without a preliminary FFN. The reported timesteps are indices along a 50-step sampling trajectory.}
\label{tab:standard_dit}
\small
\begin{tabular}{cccc}
\toprule
\textbf{Radius ($R$)} & \textbf{ODE failure ($t/50$)} & \textbf{Prediction ($t/50$)} & \textbf{Increasing FVE} \\
\midrule
12 & 32 & 33 & Yes \\
15 & 29 & 30 & Yes \\
20 & 31 & 41 & Yes \\
\bottomrule
\end{tabular}
\end{table}

\section{Robustness to Noise Schedules and Prediction Targets}
\label{appendix:robustness}

We investigate the robustness of the phase transition under different noise schedules and prediction targets on the two-dimensional modular addition task. We use the same dataset configuration as in Appendix~\ref{appendix:standard_dit}, fixing the radius to $R=15$. We consider linear, variance-preserving (VP), and variance-exploding (VE) schedules, together with data ($x$), velocity ($v$), and noise ($\epsilon$) prediction targets.

As shown in Table~\ref{tab:robustness}, the FVE trajectory generally increases, and the phase transition remains observable across the tested configurations. For $x$-prediction, the maximum second derivative of FVE remains relatively close to the ODE failure timestep. However, for $v$- and $\epsilon$-prediction, the estimated transition occurs near the beginning of the trajectory and does not accurately localize ODE failure. Thus, while the transition phenomenon is observed across these settings, the reliability of the second-derivative criterion depends on the prediction target.

\begin{table}[H]
\centering
\caption{Robustness of the phase transition to noise schedules and prediction targets. Timesteps are reported as indices along a 50-step trajectory.}
\label{tab:robustness}
\small
\setlength{\tabcolsep}{4pt}
\begin{tabular}{llccl}
\toprule
\textbf{Target} & \textbf{Schedule} & \shortstack{\textbf{ODE failure}\\\textbf{($t/50$)}} & \shortstack{\textbf{Prediction}\\\textbf{($t/50$)}} & \textbf{FVE trend} \\
\midrule
$x$          & VP     & 25 & 27 & Increasing \\
$x$          & VE     & 38 & 38 & $\sim$20\% decrease initially \\
$x$          & Linear & 33 & 39 & Increasing \\
$v$          & Linear & 36 &  1 & Increasing \\
$\epsilon$   & Linear & 34 &  3 & Increasing \\
\bottomrule
\end{tabular}
\end{table}

\section{Effect of the Number of Denoising Steps}
\label{appendix:denoising_steps}

We examine how the number of denoising steps affects the rectification trajectory in the modular addition task using EMNIST with $N=256$ images per symbol. We evaluate $T\in\{1,2,3,4,5,10,25,50\}$ denoising steps and assess both the visual quality of the generated symbols and the behavior of the FVE entropy metric.

As summarized in Table~\ref{tab:denoising_steps}, outputs obtained with $T\leq2$ are visually indistinguishable, while $T=3$ or $4$ produces recognizable symbols but insufficient refinement for a reliable assessment of the entropy-based transition. For all tested settings with $T\geq5$, the FVE increases over the trajectory, and the peak of its second derivative aligns with the ODE failure timestep, consistently with the original $T=50$ setting.

\begin{table}[H]
\centering
\caption{Effect of the number of denoising steps on visual refinement and FVE-based transition detection.}
\label{tab:denoising_steps}
\small
\begin{tabular}{lll}
\toprule
\textbf{Denoising steps ($T$)} & \textbf{Visual output} & \textbf{FVE second derivative} \\
\midrule
$T\leq2$ & Indistinguishable & N/A \\
$T=3,4$ & Recognizable symbols & Difficult to evaluate \\
$T\geq5$ & Clear images & Aligns with ODE failure \\
\bottomrule
\end{tabular}
\end{table}

\section{Feature-Volume Analysis Beyond Modular Addition}
\label{appendix:feature_volume}

To investigate whether the relationship between internal feature geometry and rectification extends beyond modular arithmetic, we analyze class-wise feature separation in both modular addition models and a pretrained class-conditional CIFAR-10 EDM U-Net.

\paragraph{Feature-volume metric.}
For each class $c$ and sampling timestep $t$, we average features obtained when $c$ is used as the condition, yielding one feature centroid per class. We center these centroids and construct their Gram matrix. Using its non-zero eigenvalues, we compute a determinant-based volume-equivalent measure of class separation, denoted by $V(t)$. A small volume indicates that class centroids are poorly separated. Since this quantity varies in the opposite direction from the FVE entropy metric, we use its inverse and predict the transition point using the maximum second derivative:
\begin{equation}
    \hat{t} = \operatorname*{arg\,max}_{t}\frac{d^2}{dt^2}\left(\frac{1}{V(t)}\right).
\end{equation}
We quantify the prediction error as
\begin{equation}
    \Delta t = \left|\hat{t}-t_{\mathrm{ODE}}\right|,
\end{equation}
where $t_{\mathrm{ODE}}$ is the first timestep at which final rectification accuracy falls below $0.5\%$.

\paragraph{Modular addition.}
For each result class, we average the features across all $p$ operand decompositions and all $p-1$ fixed wrong-result images. Table~\ref{tab:volume_modular} reports prediction errors for eleven models trained on EMNIST and KMNIST with different moduli. Across these models, the median prediction error is $0.04$, and the mean and standard deviation are $0.035$ and $0.032$, respectively. Four models have zero prediction error.

\begin{table}[H]
\centering
\caption{Transition prediction errors using the inverse feature-volume metric for modular addition models.}
\label{tab:volume_modular}
\small
\begin{tabular}{lcc}
\toprule
\textbf{Dataset} & \textbf{Modulus ($p$)} & $\boldsymbol{\Delta t}$ \\
\midrule
EMNIST & 23 & 0.06 \\
EMNIST & 27 & 0.06 \\
EMNIST & 31 & 0.08 \\
EMNIST & 35 & 0.08 \\
\midrule
KMNIST & 23 & 0.04 \\
KMNIST & 27 & 0.02 \\
KMNIST & 31 & 0.00 \\
KMNIST & 35 & 0.00 \\
KMNIST & 39 & 0.00 \\
KMNIST & 43 & 0.00 \\
KMNIST & 47 & 0.04 \\
\bottomrule
\end{tabular}
\end{table}

\paragraph{CIFAR-10.}
We further examine a pretrained class-conditional CIFAR-10 EDM with a standard U-Net architecture. For each target class $c$, we average features from the 45,000 wrong-class training images to construct its centroid. We evaluate rectification accuracy using 270 held-out wrong-class interventions. At each U-Net resolution endpoint, we calculate the volume metric across spatial patches and average the resulting volumes.

Table~\ref{tab:volume_cifar10} reports prediction errors for all seven encoder, bottleneck, and decoder representations. The median error is $0.04$, and the mean and standard deviation are $0.043$ and $0.018$, respectively; the maximum error is $0.06$. These observations provide preliminary evidence that the relationship between internal class separation and the loss of rectification ability can also be detected in class-conditional image generation, beyond the modular addition setting.

\begin{table}[H]
\centering
\caption{Transition prediction errors from the inverse feature-volume metric at different resolution endpoints of a pretrained CIFAR-10 EDM U-Net.}
\label{tab:volume_cifar10}
\small
\begin{tabular}{lc}
\toprule
\textbf{U-Net representation} & $\boldsymbol{\Delta t}$ \\
\midrule
Encoder $32\times32$ & 0.02 \\
Encoder $16\times16$ & 0.06 \\
Encoder $8\times8$ & 0.04 \\
Bottleneck $8\times8$ & 0.04 \\
Decoder $8\times8$ & 0.06 \\
Decoder $16\times16$ & 0.06 \\
Decoder $32\times32$ & 0.02 \\
\bottomrule
\end{tabular}
\end{table}


\end{document}